\documentclass{CUP-JNL-DTM}

\usepackage{graphicx}
\usepackage{booktabs}
\usepackage{amssymb}
\usepackage{makecell}
\usepackage{multirow}
\usepackage{subcaption}
\usepackage{amsmath}
\usepackage{mathtools}
\usepackage{tabularx}
\usepackage{algpseudocode}
\usepackage{algorithm}
\usepackage[subtle]{savetrees}
\usepackage[T1]{fontenc}
\usepackage{xcolor}
\usepackage{hyperref}
\usepackage[textwidth=2.5cm,textsize=footnotesize,colorinlistoftodos]{todonotes}
\usepackage{adjustbox}
\usepackage{changepage}
\usepackage[numbers,sort&compress]{natbib}
\usepackage{lscape}

\usepackage{geometry}
\usepackage{lscape}

\usepackage{macros}

\begin{document}

\begin{Frontmatter}

\title[Article Title]{NLP Verification: Towards a General Methodology for Certifying Robustness}

\author[1]{Marco Casadio}
\author[1]{Tanvi Dinkar}
\author[1,2]{Ekaterina Komendantskaya}
\author[3]{Luca Arnaboldi}%
\author[4]{Matthew L. Daggitt}
\author[5]{Omri Isac}
\author[5]{Guy Katz}
\author[1]{Verena Rieser}
\author[1]{Oliver Lemon}

\authormark{Casadio M. \textit{et al}.}

\address[1]{\orgname{Heriot-Watt University}, \orgaddress{\city{Edinburgh}, \country{UK}} \email{{mc248,t.dinkar,v.t.rieser,o.lemon}@hw.ac.uk}}
\address[2]{\orgname{University of Southampton}, \orgaddress{\city{Southampton}, \country{UK}}\email{e.komendantskaya@soton.ac.uk}}
\address[3]{\orgname{University of Birmingham}, \orgaddress{\city{Birmingham}, \country{UK}} \email{l.arnaboldi@bham.ac.uk}}
\address[4]{\orgname{University of Western Australia}, \orgaddress{\city{Perth}, \country{Australia}} \email{matthew.daggitt@uwa.edu.au}}
\address[5]{\orgname{The Hebrew University of Jerusalem}, \orgaddress{\city{Jerusalem}, \country{Israel}} \email{{omri.isac,g.katz}@mail.huji.ac.il}}

\authormark{Casadio M. et al.}

\keywords{Neural Networks, Verification, Natural Language Processing, Robustness, Adversarial Training, Machine Learning}

\abstract{
Machine Learning (ML) has 
exhibited substantial success in the field of Natural Language Processing (NLP).
For example large language models (LLMs) have empirically proven to be capable of producing text of high complexity and cohesion.
However, at the same time, they are prone to inaccuracies and hallucinations.
As these systems are increasingly integrated into real-world applications, ensuring their safety and reliability becomes a primary concern. There are safety critical contexts where such models must be robust to variability or attack, and give guarantees over their output.
Computer Vision had pioneered the use of formal verification of neural networks for such scenarios and developed common verification standards and pipelines, leveraging precise formal reasoning about geometric properties of data manifolds. In contrast, NLP verification methods have only recently appeared in the literature. While presenting sophisticated algorithms in their own right, these papers have not yet crystallised into a common methodology. They are often light on the pragmatical issues of NLP verification, and the area remains fragmented.

In this paper, we attempt to distil and evaluate general components of an NLP verification pipeline, that emerges from the progress in the field to date. Our contributions are two-fold. 
Firstly, we propose a general methodology to analyse the effect of the \emph{embedding gap} -- a problem that refers to the discrepancy between verification of geometric subspaces, and the semantic meaning of sentences which the geometric subspaces are supposed to represent.
We propose a number of practical NLP methods that can help to quantify the effects of the embedding gap.
Secondly, we give a general method for training and verification of neural networks that leverages a more precise geometric estimation of semantic similarity of sentences in the embedding space and helps to overcome the effects of the embedding gap in practice.}

\end{Frontmatter}

 \section{Introduction}
\label{sec:section1}

Deep neural networks (DNNs) have demonstrated remarkable success at addressing challenging problems in various areas, such as Computer Vision (CV)~\cite{ren2016faster} and Natural Language Processing (NLP)~\cite{sutskever2014sequence,advancesNLP}.
However, as DNN-based systems are increasingly deployed in safety-critical applications~\cite{bender2021dangers,weidinger2021ethical,bergman2022guiding,dinan-etal-2022-safetykit,RisksofFoundationModels,E2ECAI}, ensuring their safety and security becomes paramount. Current NLP systems cannot \emph{guarantee} either truthfulness, accuracy, faithfulness, or groundedness of outputs given an input query, which can lead to different levels of harm.

One such example in the
NLP domain is the requirement of a chatbot to correctly disclose non-human identity, \emph{when prompted by the user to do so}. Recently there have been several pieces of legislation proposed that will enshrine this requirement in law~\cite{EUlaw,CAlaw}. 
In order to be compliant with these new laws, in theory the underlying DNN of the chatbot (or the sub-system responsible for identifying these queries) must be \emph{100\% accurate} in its recognition of such a query. However, a central theme of generative linguistics going back to von Humboldt, is that language is `an infinite use of finite means', i.e there exists many ways to say the same thing. In reality the questions can come in a near infinite number of different forms, all with similar semantic meanings. For example: \emph{``Are you a Robot?''}, \emph{``Am I speaking with a person?''}, \emph{``Am i texting to a real human?''}, \emph{``Aren't you a chatbot?''}. 
Failure to recognise the user's intent and thus failure to answer the question correctly could potentially have legal implications for designers of these systems~\cite{EUlaw,CAlaw}. 

Similarly, as such systems become widespread in their use, it may be desirable to have guarantees on queries concerning safety critical domains, for example when the user asks for medical advice. Research has shown that users tend to attribute undue expertise to NLP systems~\cite{abercrombie2022risk,dinan-etal-2022-safetykit} potentially causing real world harm~\cite{bickmore2018patient} (e.g. `Is it safe to take these painkillers with a glass of wine?').
However, a question remains on how to ensure that NLP systems can give formally guaranteed outputs, particularly for scenarios that require maximum control over the output.

One possible solution has been to apply formal verification techniques to deep neural networks (DNN), which aims at ensuring that, for every possible input, the output generated by the network satisfies the desired properties.
One example has already been given above, i.e. guaranteeing that a system will accurately disclose its non-human identity. 
This example is an instance of the more general problem of DNN \emph{robustness verification}, where the aim is to guarantee that every point in a given region of the embedding space is classified correctly.
Concretely, given a network  $\dnn{}: \; \real^\indim \rightarrow \real^\outdim$, one first defines \emph{subspaces}  $\subspace_1, \ldots, \subspace_\nsubspaces$ of the \emph{vector space} $\real^\indim$. For example, one can define ``$\epscubes$'' or ``$\epsballs$''\footnote{The terminology will be made precise in Example~\ref{ex:cube}.}
around all input vectors given by the dataset in question (in which case the number of $\subspace_1, \ldots, \subspace_\nsubspaces$ will correspond to the number of samples in the given dataset). Then, using a separate \emph{verification algorithm} $\verificationalgorithm$, we verify whether $\dnn{}$ is \emph{robust} for each $\subspace_i$, i.e. whether $\dnn{}$ assigns the same class for all vectors contained in $\subspace_i$.
Note that each $\subspace_i$ is itself infinite (i.e. continuous), and thus $\verificationalgorithm$ is usually based on equational reasoning, abstract interpretation or bound propagation (see related work in Section~\ref{sec:section2}). 
The subset of $\subspace_1, \ldots , \subspace_l$ for which $\dnn{}$ is proven robust, forms the set of  \emph{verified subspaces} of the given vector space (for $\dnn{}$). The percentage of verified subspaces 
is called the \emph{verification success rate} (or \emph{verifiability}). Given $\subspace_1, \ldots, \subspace_\nsubspaces$, we say a DNN \emph{$\dnn{1}$ is more verifiable than $\dnn{2}$} if $\dnn{1}$ has higher  \emph{verification success rate} on $\subspace_1, \ldots, \subspace_\nsubspaces$.
Despite not providing a formal guarantee about the entire embedding space, this result is useful as it provides guarantees about the behaviour of the network over a large set of unseen inputs.

Existing verification approaches primarily focus on computer vision (CV) tasks, where images are seen as vectors in a continuous space and every point in the space corresponds to a valid image. In contrast, sentences in NLP form a discrete domain\footnote{In this paper, we work with textual representations of sentences. Raw audio input can be seen as continuous, but this is out of scope of this paper.}, making it challenging to apply traditional verification techniques effectively. 
In particular, taking an NLP dataset $\dataset$  to be a set of sentences $\sentence_1, \ldots , \sentence_\nsentences$ written in natural language, an embedding $\embeddingfunc$ is a function that maps a sentence to a vector in $\real^\indim$. The resulting vector space is called \emph{the embedding space}. Due to discrete nature of the set $\dataset$, the reverse of the embedding function $\embeddingfunc^{-1}: \real^\indim \rightarrow \dataset$ is undefined for some elements of  $\real^\indim$. This problem is known as the \emph{``problem of the embedding gap''}. Sometimes, one uses the term to more generally refer to any discrepancies that   $\embeddingfunc$ introduces, for example, when it maps dissimilar sentences close
in $\real^\indim$. We use the term in both mathematical and NLP sense.

Mathematically, the general (geometric) ``DNN robustness verification'' approach of defining and verifying subspaces of $\real^\indim$ should work, and some prior works exploit this fact. However, pragmatically, because of the embedding gap, usually only a tiny fraction of vectors contained in the verified subspaces map back to valid sentences.
When a verified subspace contains no or very few sentence embeddings, we say that verified subspace has \emph{low generalisability}. Low generalisability may render verification efforts ineffective for practical applications.

From the NLP perspective, there are other, more subtle, examples where the embedding gap can manifest. %
Consider an example of a subspace containing sentences that are semantically similar to the sentence: \emph{`i~really like too chat to a human. are you one?'}.
Suppose we succeed in verifying a DNN to be robust on this subspace.
This provides a guarantee that the DNN will always identify sentences in this subspace as questions about human/robot identity.
But suppose the embedding function $\embeddingfunc$
wrongly embeds sentences belonging to an opposite class into this subspace. For example, the LLM Vicuna~\cite{vicuna2023} generates the following sentence as a rephrasing of the previous one: \emph{Do you take pleasure in having a conversation with someone?}. Suppose our verified subspace contained an embedding of this sentence too, and thus our verified DNN identifies this second sentence to belong to the same class as the first one. However, the second sentence is not a question about human/robot identity of the agent! When we can find such an example, we say that
the verified subspace is \emph{prone to embedding errors.}

Robustness verification in NLP is particularly susceptible to this problem, because we cannot cross the embedding gap in the opposite direction as the embedding function is not invertible. This means it is difficult for humans to understand what sort of sentences are captured by a given subspace.

\subsection*{Contributions} 
Our main aim is to provide a \textbf{general and principled verification methodology} that bridges the embedding gap when possible; and gives precise metrics to evaluate and report its effects in any case.
The contributions split into two main groups, depending on whether the embedding gap is approached from mathematical or NLP perspective. 

\begin{figure}[t]
\centering
	\begin{subfigure}[b]{0.24\textwidth}
            \label{fig:ball-hrect-a}
		\includegraphics[width=\textwidth]{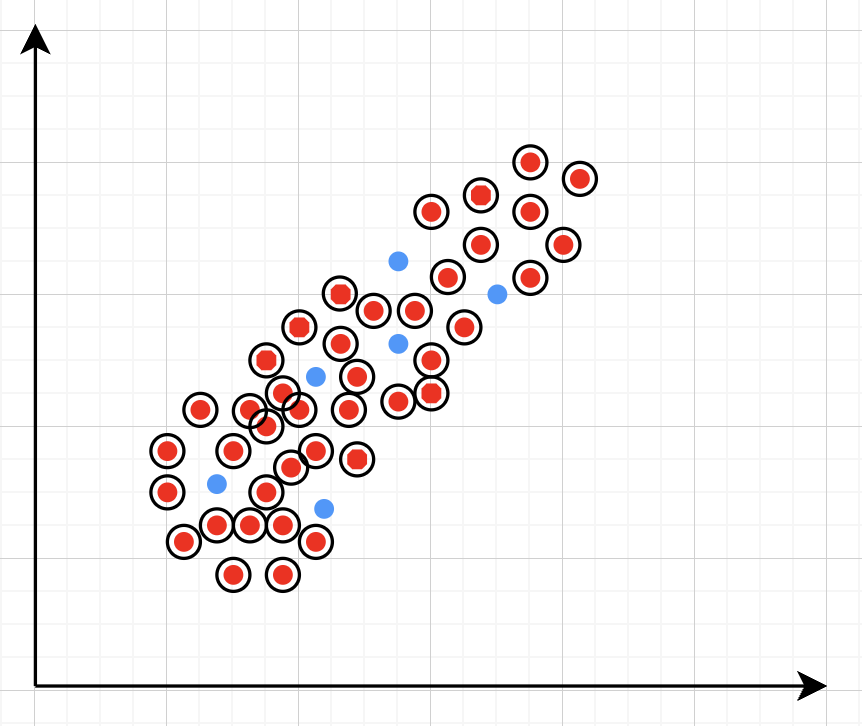}
            \caption{}
	\end{subfigure}
	\begin{subfigure}[b]{0.24\textwidth}
            \label{fig:ball-hrect-b}
		\includegraphics[width=\textwidth]{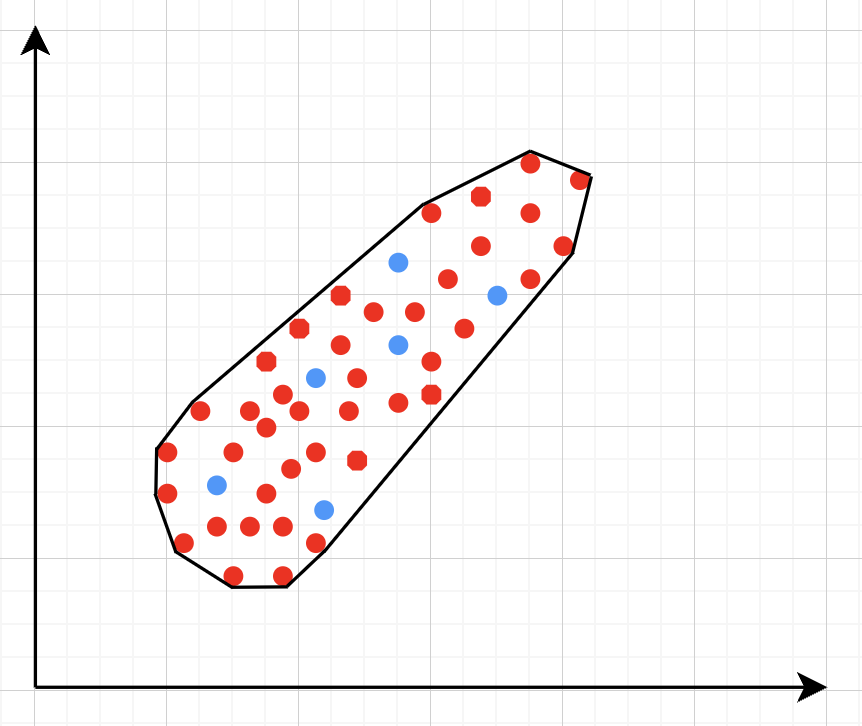}
            \caption{}
	\end{subfigure}
	\begin{subfigure}[b]{0.24\textwidth}
            \label{fig:ball-hrect-c}
		\includegraphics[width=\textwidth]{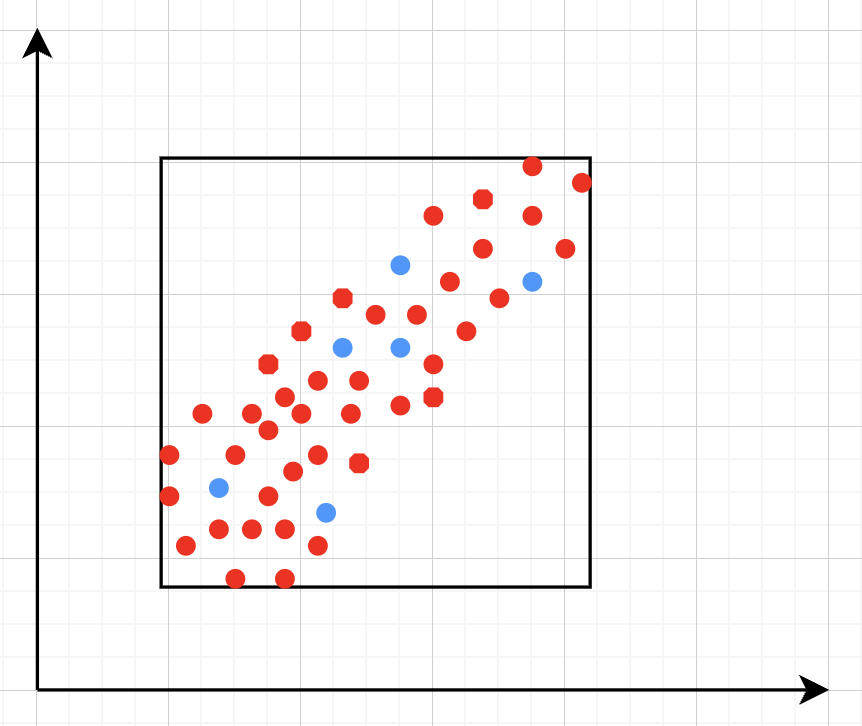}
            \caption{}
        \end{subfigure}
        \begin{subfigure}[b]{0.24\textwidth}
            \label{fig:ball-hrect-d}
		\includegraphics[width=\textwidth]{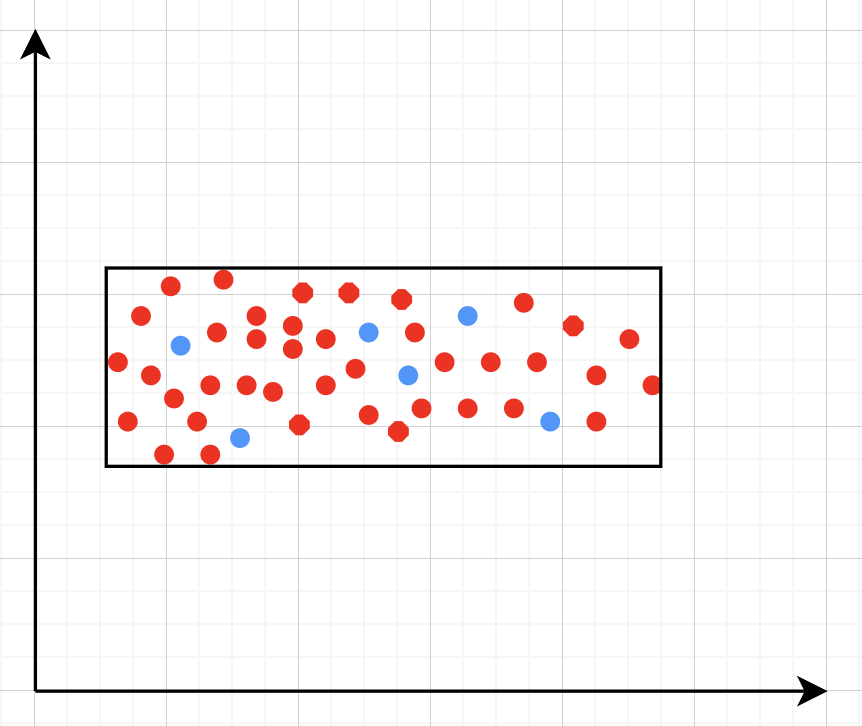}
            \caption{}
	\end{subfigure}
\caption{\small\emph{An example of verifiable but not generalisable $\epsballs$ (a), convex-hull around selected embedded points (b), hyper-rectangle around same points (c) and rotation of such hyper-rectangle (d) in 2-dimensions. The red dots represent sentences in the embedding space from the training set belonging to one class, while the turquoise dots are embedded sentences from the test set belonging to the same class.}}
\label{fig:ball-hrect}
\end{figure}

\paragraph{Contributions Part 1: Characterisation of Verifiable Subspaces and general} NLP Verification Pipeline. We start by showing, through a series of experiments, that purely geometric approaches to NLP verification (such as those based on the $\epsball$~\cite{shi2020robustness}) suffer from the \emph{verifiability-generalisability trade-off}: that is, when one metric improves, the other deteriorates. Figure~\ref{fig:ball-hrect} gives a good idea of the problem: the smaller the $\epsball$s are, the more verifiable they are, and less generalisable. To the best of our knowledge, this phenomenon has not been reported in the literature before (in the NLP context).
We propose a general method for measuring \textbf{generalisability of the verified subspaces}, based on algorithmic generation of semantic attacks on sentences included in the given verified semantic subspace.

An alternative method to the purely geometric approach
is to construct subspaces of the embedding space based on the \emph{semantic perturbations} of sentences (first attempts to do this appeared in~\cite{FoMLAS2023:ANTONIO_Towards_Systematic_Method,jia2019certified,huang2019achieving,zhang2021certified}). Concretely, the idea is to form each $\subspace_i$ by embedding a sentence $\sentence$ and $n$ semantic perturbations of $\sentence$ into the real vector space and enclosing them inside some geometric shape. Ideally, this shape should be the convex hull around the $n+1$ embedded sentences (see Figure~\ref{fig:ball-hrect}), however calculating convex hulls with sufficient precision is computationally infeasible for high number of dimensions. Thus, simpler shapes, such as \emph{hyper-cubes} and \emph{hyper-rectangles} are used in the literature.
We propose a novel refinement of these ideas, by including the method of a \emph{hyper-rectangle rotation} in order to increase the shape precision (see Figure~\ref{fig:ball-hrect}). We will call the resulting shapes \emph{semantic subspaces} (in contrast to those obtained purely geometrically).

A few questions have been left unanswered in the previous work~\cite{FoMLAS2023:ANTONIO_Towards_Systematic_Method,jia2019certified,huang2019achieving,zhang2021certified}. Firstly, because generalisability of the verified subspaces is not reported in the literature, we cannot know whether the prior semantically-informed approaches are better in that respect than purely geometric methods. If they are better in both verifiability and generalisability, it is unclear whether the improvement should be attributed to:

\begin{itemize}
    \item the fact that verified semantic~subspaces simply have an optimal volume (for the verifiability-generalisability trade-off), or 
    \item the improved precision of verified subspaces that comes from using the semantic knowledge. 
\end{itemize}

Through a series of experiments, we confirm that semantic subspaces are more verifiable and more generalisable than their geometric counterparts. Moreover, by comparing the volumes of the obtained verified semantic and geometric subspaces, we show that the improvement is partly due to finding an optimal size of subspaces (for the given embedding space), and partly due to improvement in shape precision.

The second group of unresolved questions concerns robust training regimes in NLP verification that is used as means of improving verifiability of subspaces in prior works~\cite{FoMLAS2023:ANTONIO_Towards_Systematic_Method,jia2019certified,huang2019achieving,zhang2021certified}. 
It was not clear what made robust training successful: 
\begin{itemize}
\item was it because additional examples generally improved the precision of the decision boundary (in which case dataset augmentation would have a similar effect);  
\item was it because adversarial examples specifically improved adversarial robustness (in which case simple $\epsball$ PGD attacks would have a similar effect);  or 
\item did the knowledge of semantic subspaces  play the key role? 
\end{itemize}

Through a series of experiments we show that the latter is the case.
In order to do this, we formulate a \emph{semantically robust training} 
method that uses projected gradient descent on semantic subspaces (rather than on $\epsballs$ as the famous PGD algorithm does~\cite{madry2018towards}).
We use different forms of semantic perturbations, at character, word and sentence levels (alongside the standard PGD training and data augmentation) to perform semantically robust training.
We conclude that \textbf{semantically robust training} generally wins over the standard robust training methods. Moreover, the more sophisticated semantic perturbations we use in semantically robust training, the more verifiable the neural network will be obtained as a result (at no cost to generalisability).
For example, using the strongest form of attack (the polyjuice attack~\cite{wu2021polyjuice}) in semantically robust training, we obtain DNNs that are more verifiable irrespective of the way the verified sub-spaces are formed.
  
As a result, we arrive at a fully parametric approach to NLP verification that disentangles the four components:
  \begin{itemize}
   \item choice of the semantic attack (on the NLP side), 
   \item semantic subspace formation in the embedding space (on the geometric side),
   \item semantically robust training (on the machine learning side),
   \item choice of the verification algorithm (on the verification side).
\end{itemize}
We argue that
this approach opens the way for more principled 
NLP verification methods that reduces the effects of the embedding gap; and generation of more transparent NLP verification benchmarks. 
We implement a tool ANTONIO that generates NLP verification benchmarks based on the above choices.
This paper is the first to use a complete SMT-based verifier (namely Marabou~\cite{marabou}) for NLP verification.

\paragraph{Contributions Part 2: NLP Verification Pipeline in Use: an NLP Perspective on the Embedding Gap.}

We test the theoretical results by suggesting an \textbf{NLP verification pipeline}, a general methodology that starts with NLP analysis of the dataset and obtaining semantically similar perturbations that together characterise the semantic meaning of a sentence; proceeds with embedding of the sentences into the real vector space and defining semantic subspaces around embeddings of semantically similar sentences; and culminates with using these subspaces for both training and verification.
This clear division into stages allows us to formulate practical NLP methods for
minimising the effects of the embedding gap. In particular, we show that 
the quality of the generated sentence perturbations maybe improved through the use of human evaluation, cosine similarity and ROUGE-N.  
We introduce the novel \textbf{embedding error} metric as an effective practical way to measure the quality of the embedding functions. Through a detailed case study, we show how geometric and NLP intuitions can be put at work towards obtaining DNNs that are more verifiable over better generalisable and less prone to embedding errors semantic subspaces. Perhaps more importantly, the proposed methodology opens the way for transparency in reporting NLP verification results, -- something that this domain will benefit from if it reaches the stage of practical deployment of NLP verification pipelines.  

\emph{Paper Outline.} From here, the paper proceeds as follows. Section~\ref{sec:section2} gives an extensive literature review encompassing DNN verification methods generally, and NLP verification methods in particular. The section culminates with distilling a common \emph{``NLP verification pipeline''} encompassing the existing literature. Based on the understanding of major components of the pipeline, the rest of the paper focuses on improving understanding or implementation of its components.   Section~\ref{sec:section3} formally defines the components of the pipeline in a general mathematical notation, which abstracts away from particular choices of sentence perturbation, sentence embedding, training and verification algorithms. The central notion the section introduces is that of \emph{geometric and semantic subspaces}. The next Section~\ref{sec:section4} makes full use of this general definition, and shows that semantic subspaces play a pivotal role in improving verification and training of DNNs in NLP. This section formally defines the \emph{generalisability metric} and considers the problem of \emph{generalisability-verifiability trade-off}. Through thorough empirical evaluation, it shows that a principled approach to defining semantic subspaces
can help to improve both generalisability and verifiability of DNNs, thus reducing the effects of the trade-off. The final Section~\ref{sec:section5} further tests the NLP verification pipelines using state-of-the-art NLP tools, and analyses the effects of the embedding gap from the NLP perspective, in particular it introduces a method of measuring the \emph{embedding error} and reporting this metric alongside verifiability and generalisability. Section~\ref{sec:section6} concludes the paper and discusses future work.

 \section{Related Work}
\label{sec:section2}

\subsection{DNN Verification}
Formal verification is an active field across several domains including hardware~\cite{kp01,patankar1999formal}, software~\cite{jourdan2015formally}, network protocols~\cite{metere2022automating} and many more~\cite{woodcock2009formal}. 
However, it was only recently that this
became applicable to the field of machine learning~\cite{katz2017reluplex}.
An input query to a verifier consists of a subspace within the embedding space and a target subspace of outputs, typically a target output class. 
The verifier then returns either \emph{true}, \emph{false} or \emph{unknown}.
\emph{True} indicates that there exists an input within the given input subspace whose output falls within the given output subspace, often accompanied by an example of such input. \emph{False} indicates that no such input exists.
Several verifiers are popular in DNN verification and competitions~\cite{bak2021second,baluta2021scalable,liu2021algorithms,singh2019beyond}.
We can divide them into 2 main categories: complete verifiers which return \emph{true/false} and incomplete verifiers which return \emph{true/unknown}.
While complete verifiers are always deterministic, incomplete verifiers may be probabilistic.
Unlike deterministic verification, probabilistic verification is not sound and a verifier may incorrectly output \emph{true} with a very low probability (typically 0.01\%).

\emph{Complete Verification based on Linear Programming \& Satisfiability Modulo Theories (SMT)} solving.
Generally, SMT solving is a group of methods for determining the satisfiability of logical formulas with respect to underlying mathematical theories such as real arithmetic, bit-vectors, or arrays~\cite{barrett2018satisfiability}. These methods extend traditional satisfiability (SAT) solving by incorporating domain-specific reasoning, making them particularly useful for verifying complex systems.
In the context of neural network verification, SMT solvers encode network behaviours and safety properties as logical constraints, enabling rigorous checks for violations of specifications~\cite{albarghouthi2021introduction}.
When the activation functions are piecewise linear (e.g. ReLU), the DNN can be encoded by conjunctions and disjunctions of linear inequalities and thus linear programming algorithms can be directly applied to solve the satisfiability problem.
A state-of-the-art tool is Marabou~\cite{marabou}, which answers queries about neural networks and their properties in the form of constraint satisfaction problems. Marabou takes the network as input and first applies multiple pre-processing steps to infer bounds for each node in the network.  It applies the algorithm ReLUplex~\cite{katz2017reluplex},  a combination of \emph{Simplex}~\cite{dantzig1963linear} search over linear constraints, modified to work for networks with piece-wise linear activation functions.
With time, Marabou grew into a complex prover with multiple heuristics supplementing the original ReLUplex algorithm~\cite{marabou}, for example it now includes mixed-integer linear programming (MILP)~\cite{winston2004operations} and abstract interpretation based algorithms which we survey below.
MILP-based approaches~\cite{cheng2017maximum,lomuscio2017approach,tjeng2019evaluating} encode the verification problem as a mixed-integer linear programming problem, in which the constraints are linear inequalities and the objective is represented by a linear function.
Thus, the DNN verification problem can be precisely encoded as a MILP problem.
For example, ERAN~\cite{NEURIPS2018_f2f44698}
combines abstract interpretation with the MILP solver GUROBI~\cite{gurobi2020gurobi}.
By the time Branch and Bound (BaB) methodologies are introduced later, it becomes evident that the verification community has effectively consolidated diverse approaches into a unified taxonomy. Modern verifiers, such as $\alpha\beta$-CROWN~\cite{xu2021fast,wang2021beta}, take full advantage of this combination and effectively balance efficiency with precision.

\emph{Incomplete Verification based on Abstract Interpretation} takes inspiration from the domain of abstract interpretation, and mainly uses linear relaxations on ReLU neurons, resulting in an over-approximation of the initial constraint.
Abstract interpretation was first developed by Cousot and Cousot~\cite{cousot1977abstract} in 1977. It formalises the idea of abstraction of mathematical structures, in particular those involved in the specification of properties and proof methods of computer systems~\cite{cousot2003verification} and it has since been used in many applications~\cite{cousot2014abstract}.
Specifically, for DNN verification, this technique can model the behaviour of a network using an abstract domain that captures the possible range of values the network can output for a given input.
Abstract interpretation-based verifiers can define a lower bound and an upper bound of the output of each ReLU neuron as linear constraints, which define a region called ReLU polytope that gets propagated through the network.
To propagate the bounds, one can use \textit{interval bound propagation} (IBP)~\cite{wong2018provable,gowal2019scalable,lyu2020fastened,mirman2018differentiable}.
The strength of IBP-based methods lies in their efficiency; they are faster than alternative approaches and demonstrate superior scalability. However, their primary limitation lies in the inherently loose bounds they produce~\cite{gowal2019scalable}. This drawback becomes particularly pronounced in the case of deeper neural networks, typically those with more than 10 layers~\cite{li2023sok}, where they cannot certify non-trivial robustness properties.
Other methods that are less efficient but produce tighter bounds are based on polyhedra abstraction, such as CROWN~\cite{zhang2018efficient} and DeepPoly~\cite{singh2019abstract}, or based on multi-neuron relaxation, such as PRIMA~\cite{muller2022prima}.
An abstract interpretation tool CORA~\cite{Althoff2015ARCH},  uses polyhedral abstractions and reachability analysis for formal verification of neural networks. It integrates various set representations, such as zonotopes, and algorithms to compute reachable sets for both continuous and hybrid systems, providing tighter bounds in verification tasks.
Another mature tool in this category is ERAN~\cite{NEURIPS2018_f2f44698}, 
which uses abstract domains (DeepPoly) with custom multi-neuron relaxations (PRIMA) to support fully-connected, convolutional, and residual networks with ReLU, Sigmoid, Tanh, and Maxpool activations.
Note that, having lost completeness, they can work with a more general class of neural networks (e.g. neural networks with non linear layers).

\emph{Modern Neural Network Verifiers.} Modern verifiers are complex tools that take advantage of a combination of complete and incomplete methods as well as additional heuristics.
The term Branch and Bound (BaB)~\cite{gehr2018ai2,bunel2018unified,bunel2020branch,ferrari2022complete,jordan2019provable,wang2021beta,zhang2022general} often refers to the method that relies on the piecewise linear property of DNNs: since each ReLU neuron outputs ReLU($x$) = max\{$x$,$0$\} is piecewise linear, we can consider its two linear pieces $x\geq 0$, $x\leq 0$ separately. 
A BaB verification approach, as the name suggests, consists of two parts: branching and bounding.
It first derives a lower bound and an upper bound, then, if the lower bound is positive it terminates with `verified', else, if the upper bound is non-positive it terminates with `not verified' (\emph{bounding}). Otherwise, the approach recursively chooses a neuron to split into two branches (\emph{branching}), resulting in two linear constraints. Then bounding is applied to both constraints and if both are satisfied the verification terminates, otherwise the other neurons are split recursively. When all neurons are split, the branch will contain only linear constraints, and thus the approach applies linear programming to compute the constraint and verify the branch.
It is important to note that BaB approaches themselves are neither inherently complete nor incomplete. BaB is an algorithm for splitting problems into sub-problems and requires a solver to resolve the linear constraints. The completeness of the verification depends on the combination of BaB and the solver used.
\emph{Multi-Neuron Guided Branch-and-Bound (MN-BaB)}~\cite{ferrari2022complete} is a state-of-the-art neural network verifier that builds on the tight multi-neuron constraints proposed in PRIMA~\cite{muller2021precise} and leverages these constraints within a BaB framework to yield an efficient, GPU based dual solver.
Another state-of-the-art tool is $\alpha\beta$-CROWN~\cite{xu2021fast,wang2021beta}, a neural network verifier based on an efficient linear bound propagation framework and branch-and-bound. It can be accelerated efficiently on GPUs and can scale to relatively large convolutional networks (e.g., $10^7$ parameters).
It also supports a wide range of neural network architectures (e.g., CNN, ResNet, and various activation functions).

\emph{Probabilistic Incomplete Verification} approaches add random noise to models to smooth them, and then derive certified robustness for these smoothed models.
This field is commonly referred to as Randomised Smoothing, 
given that these approaches provide probabilistic guarantees of robustness, and all current probabilistic verification techniques are tailored for smoothed models~\cite{lecuyer2019certified,li2019certified,dvijotham2020framework,zhang2020black,salman2019provably,mohapatra2020higher}.
Given that our work focuses on deterministic approaches, here we only report the existence of this line of work without going into details.

Note that these existing verification approaches primarily focus on computer vision tasks, where images are seen as vectors in a continuous space and every point in the space corresponds to a valid image, while sentences in NLP form a discrete domain, making it challenging to apply traditional verification techniques effectively.

In this work we use both an abstract interpretation-based incomplete verifier (ERAN~\cite{NEURIPS2018_f2f44698}) and an SMT-based complete verifier (Marabou~\cite{marabou}) in order to demonstrate the effect that the choice of a verifier may bring, and demonstrate common trends.

\subsection{Geometric Representations in DNN Verification}
Geometric representations form the backbone of many DNN verification techniques, enabling the encoding and manipulation of input and output bounds during analysis. Among these, hyper-rectangles, including $\epscubes$, are the most widely used due to their simplicity and efficiency in over-approximating neural network behaviors~\cite{wong2018provable,gowal2019scalable}.These representations are computationally lightweight, making them highly scalable to large networks. However, they often produce loose approximations, particularly in deeper or more complex architectures, which can limit the precision of the verification results~\cite{gowal2019scalable}.
Other representations, such as zonotopes~\cite{gehr2018ai2,singh2019abstract,adcnn}, offer tighter approximations and better capture the linear dependencies between neurons but at a higher computational cost. Polyhedra-based methods, as employed in tools like DeepPoly~\cite{singh2019abstract} and PRIMA~\cite{muller2022prima}, provide even more precise abstractions by considering multi-dimensional relationships between neurons. However, these methods trade off efficiency for precision, making them less scalable to large and deep networks.
Ellipsoidal representations~\cite{Althoff2015ARCH} are another class of geometric abstractions that provide compact and smooth bounds for neural network outputs. These representations are particularly useful for capturing the effects of continuous transformations in hybrid systems and other control applications. However, operations such as intersection and propagation through non-linear layers can be computationally intensive, which limits their applicability in large-scale neural network verification tasks.
The dominance of hyper-rectangles in the field stems from their balance of computational simplicity and generality. Nonetheless, ongoing research continues to explore how alternative shapes, hybrid approaches, or adaptive representations might better meet the demands of increasingly complex neural network architectures.

\subsection{Robust Training}
\label{sec:sec22}

Verifying DNNs poses significant challenges if they are not appropriately trained. The fundamental issue lies in the failure of DNNs, including even sophisticated models, to meet essential verification properties, such as \emph{robustness}~\cite{casadio2022robust}.
To enhance robustness, various training methodologies have been proposed. It is noteworthy that, although robust training by \emph{projected gradient descent}~\cite{goodfellow2015explaining,madry2018towards,kolter2018adversarial} predates verification, contemporary approaches are often related to, or derived from, the corresponding verification methods by optimizing verification-inspired regularization terms or injecting specific data augmentation during training.
In practice, after robust training, the model usually achieves higher certified robustness and is more likely to satisfy the desired verification properties~\cite{casadio2022robust}. Thus, robust training is a strong complement to robustness verification approaches.

\emph{Robust training} techniques can be classified into several large groups:
\begin{itemize}
\item data augmentation~\cite{rebuffi2021data}, 
\item adversarial training~\cite{goodfellow2015explaining,madry2018towards} including property-driven training~\cite{FischerBDGZV19,SlusarzKDSS23}, 
\item IBP training~\cite{gowal2019scalable,zhang2019stable} and other forms of
certified training~\cite{muller2023certified}, or 
\item a combination thereof~\cite{zhang2020robustness,casadio2022robust}.
\end{itemize}
Data augmentation involves the creation of synthetic examples through the application of diverse transformations or perturbations to the initial training data. These generated instances are then incorporated into the original dataset to enhance the training process.
Adversarial training entails identifying worst-case examples at each epoch during the training phase and calculating an additional loss on these instances. State of the art adversarial training involves projected gradient descent algorithms such as FGSM~\cite{goodfellow2015explaining} and PGD~\cite{madry2018towards}.
Certified training methods focus on providing mathematical guarantees about the model's behaviour within certain bounds. Among them, we can name IBP 
training~\cite{gowal2019scalable,zhang2019stable} techniques, which impose intervals or bounds on the predictions or activations of the model, ensuring that the model's output lies within a specific range with high confidence.

Note that all techniques mentioned above can be categorised based on whether they primarily \emph{augment the data} (such as data augmentation) or \emph{augment the loss function} (as seen in adversarial, IBP and certified training).
Augmenting the data tends to be efficient, although it may not help against stronger adversarial attacks. Conversely, methods that manipulate the loss functions directly are more resistant to strong adversarial attacks but often come with higher computational costs. Ultimately, the choice between altering data or loss functions depends on the specific requirements of the application and the desired trade-offs between performance, computational complexity, and robustness guarantees.

\subsection{NLP robustness} 
There exists a substantial body of research dedicated to enhancing the adversarial robustness of NLP systems~\cite{zhang2020adversarial,9557814,wang2021measure,li2021searching,zhou2021defense,zhu2019freelb,dong2021towards}. These efforts aim to mitigate the vulnerability of NLP models to adversarial attacks and improve their resilience in real-world scenarios~\cite{9557814,wang2021measure} and mostly employ data augmentation techniques~\cite{feng-etal-2021-survey,DBLP}.
In NLP, we can distinguish perturbations based on three main criteria: 
\begin{itemize}
\item 
where and how the perturbations occur, 
\item whether they are altered automatically using some defined rules (vs. generated by humans or LLMs) and
\item whether they are adversarial (as opposed to random).
\end{itemize}
In particular, perturbations can occur at the character, word, or sentence level~\cite{cheng2019robust,iyyer2018adversarial,cao2022tasa} and may involve deletion, insertion, swapping, flipping, substitution with synonyms, concatenation with characters or words, or insertion of numeric or alphanumeric characters~\cite{liang2017deep,ebrahimi2018hotflip,lei2022phrase}.
For instance, in character level adversarial attacks, Belinkov et al.~\cite{belinkov2017synthetic} introduce natural and synthetic noise to input data, while Gao et al.~\cite{gao2018blackbox} and Li et al.~\cite{TextBugger2019} identify crucial words within a sentence and perturb them accordingly. For word level  attacks, they can be categorised into gradient-based~\cite{liang2017deep,samanta2017crafting}, importance-based~\cite{ivankay2022fooling,jin2020bert}, and replacement-based~\cite{alzantot2018generating,kuleshov2018adversarial,pennington2014glove} strategies, based on the perturbation method employed. 
Moreover, Moradi et al.~\cite{moradi2021evaluating} introduce rule-based non-adversarial perturbations at both the character and word levels.
Their method simulates various types of noise typically caused by spelling mistakes, typos, and other similar errors.
In sentence level adversarial attacks, some perturbations~\cite{jia2017adversarial,wang2018robust} are created so that they do not impact the original label of the input and can be incorporated as a concatenation in the original text. In such scenarios, the expected behaviour from the model is to maintain the original output, and the attack can be deemed successful if the label/output of the model is altered.
Additionally, non-rule-based sentence perturbations can be obtained through prompting LLMs~\cite{wu2021polyjuice,vicuna2023} to generate rephrasing of the inputs.
By augmenting the training data with these perturbed examples, models are exposed to a more diverse range of linguistic variations and potential adversarial inputs. This helps the models to generalise better and become more robust to different types of adversarial attacks.
To help with this task, the NLP community has gathered a dataset of adversarial attacks named AdvGLUE~\cite{wang2021adversarial}, which aims to be a principled and comprehensive benchmark for NLP robustness measurements.

In this work we employ a PGD-based adversarial training as the method to enhance the robustness and verifiability of our models against gradient-based adversarial attacks. For non-adversarial perturbations, we create rule-based perturbations at the character and word level as in Moradi et al.~\cite{moradi2021evaluating} and non-rule-based perturbations at the sentence level using PolyJuice~\cite{wu2021polyjuice} and Vicuna~\cite{vicuna2023}.
We thus cover most combinations of the three choices above (bypassing only human-generated adversarial attacks as  there is no sufficient data to admit systematic evaluation which is important  for this study).

\begin{landscape}
\begin{table}[htbp]
\centering
\footnotesize
\begin{tabularx}{\linewidth}{|p{0.1\textwidth}|p{0.16\textwidth}|p{0.175\textwidth}|p{0.14\textwidth}|p{0.14\textwidth}|p{0.24\textwidth}|p{0.178\textwidth}|p{0.12\textwidth}|}
\toprule
\textbf{Method} & \textbf{Datasets} & \textbf{NLP perturbations} & \textbf{Embeddings} & \textbf{Architectures (\# of parameters)} & \textbf{Robust training} & \textbf{Verification algorithm} & \textbf{Verification characteristics} \\
\midrule
\midrule
\textbf{Ours} & RUARobot, \newline Medical & General purpose: char, word and sentence perturbations, $\epsball$ & Sentence: S-BERT, S-GPT & FFNN ($10^4$) & \textbf{PGD}-based loss function augmentation & \textbf{SMT}, \textbf{BaB}, Abstract interpretation \newline  & \textbf{Complete}, \newline \textbf{Deterministic} \\
\midrule
Jia et al. (2019) \cite{jia2019certified} & IMDB, SNLI & Word substitution & Word: GloVe & LSTM, CNN, BoW, Attention-based, ($10^5$) & \textbf{IBP}-based loss function augmentation & Abstract Interpretation IBP & Incomplete, \newline \textbf{Deterministic} \\
\midrule
Huang et al. (2019) \cite{huang2019achieving} & AGNews, SST & Char and word substitution & Word: GloVe & CNN ($10^5$) & \textbf{IBP}-based loss function augmentation & Abstract Interpretation IBP & Incomplete, \newline \textbf{Deterministic} \\
\midrule
Welbl et al. (2020) \cite{welbl2020towards} & SNLI, MNLI & Word deletion & Word: GloVe & Attention-based ($10^5$) & Data augmentation, random and beam search adversarial training, \textbf{IBP}-based & Abstract Interpretation IBP & Incomplete, \newline \textbf{Deterministic} \\
\midrule
Zhang et al. (2021) \cite{zhang2021certified} & IMDB, SST, SST2 & Word perturbations & Word: not specified & LSTM ($10^5$) & \textbf{IBP}-based loss function augmentation & Abstract Interpretation IBP & Incomplete, \newline \textbf{Deterministic} \\
\midrule
Wang et al. (2023) \cite{wang2023robustness} & IMDB, YELP, SST2 & Word substitution & Word: GloVe & CNN ($10^5$) & \textbf{IBP}-based: Embedding Interval Bound Constraint (EIBC) triplet loss & Abstract Interpretation IBP & Incomplete, \newline \textbf{Deterministic} \\
\midrule
Ko et al. (2019) \cite{ko2019popqorn} & CogComp QC & $\epsball$ & Word: not specified & RNN, LSTM ($10^5$) & - & Abstract Interpretation IBP  & Incomplete, \newline \textbf{Deterministic} \\
\midrule
Shi et al. (2020) \cite{shi2020robustness} & YELP, SST & $\epsball$ & Word: not specified & Transformer ($10^6$) & - & Abstract Interpretation IBP & Incomplete, \newline \textbf{Deterministic} \\
\midrule
Du et al. (2021) \cite{du2021cert} & Rotten Tomatoes Movie Review, Toxic Comment & $\epsball$ & Word: GloVe & RNN, LSTM ($10^5$) & \textbf{Zonotope}-based loss function augmentation & Abstract Interpretation Zonotopes & Incomplete, \newline \textbf{Deterministic} \\
\midrule
Bonaert et al. (2021) \cite{bonaert2021fast} & SST, YELP & $\epsball$ & Word: not specified & Transformer ($10^6$) & - & Abstract Interpretation Zonotopes  & Incomplete, \newline \textbf{Deterministic} \\
\bottomrule
\end{tabularx}
\caption{\small \emph{Summary of the main features of the existing NLP verification approaches. In bold are state-of-the-art methods.}}
\label{tab:verification-comparison-1}
\end{table}
\end{landscape}

\begin{landscape}
\begin{table}[htbp]
\centering
\footnotesize
\begin{tabularx}{\linewidth}{|p{0.1\textwidth}|p{0.16\textwidth}|p{0.175\textwidth}|p{0.14\textwidth}|p{0.14\textwidth}|p{0.24\textwidth}|p{0.178\textwidth}|p{0.12\textwidth}|}
\toprule
\textbf{Method} & \textbf{Datasets} & \textbf{NLP perturbations} & \textbf{Embeddings} & \textbf{Architectures (\# of parameters)} & \textbf{Robust training} & \textbf{Verification algorithm} & \textbf{Verification characteristics} \\
\midrule
\midrule
Ye et al. (2020) \cite{ye2020safer} & IMDB, Amazon & Word substitution & Word: GloVe & Transformer ($10^8$) & Data \newline augmentation & Randomised smoothing ($\alpha = 0.01, n = 5000$) & Incomplete, \newline Probabilistic \\
\midrule
Wang et al. (2021) \cite{wang-etal-2021-certified} & IMDB, AGNews & Word substitution & Word: GloVe & LSTM ($10^5$) & Data \newline augmentation & Differential privacy & Incomplete, \newline Probabilistic \\
\midrule
Zhao et al. (2022) \cite{zhao2022certified} & AGNews, SST & Word substitution & Word: GloVe & Transformer ($10^8$) & Data \newline augmentation \newline and IBP-based & Randomised smoothing ($\alpha = 0.001, n = 30050$) & Incomplete, \newline Probabilistic \\
\midrule
Zeng et al. (2023) \cite{zeng2023certified} & IMDB, YELP & Char and word substitution & Word: not specified & Transformer ($10^8$) & Data \newline augmentation & Randomised smoothing ($\alpha = 0.05, n = 5000$) & Incomplete, \newline Probabilistic \\
\midrule
Ye et al. (2023) \cite{ye2023unit} & IMDB, SST2, YELP, AGNews & Word substitution & Word: not specified & Transformer ($10^8$) & Data \newline augmentation & Randomised smoothing ($\alpha = 0.001, n = 9000$) & Incomplete, \newline Probabilistic \\
\midrule
Zhang et al. (2023) \cite{zhang2023text} & IMDB, Amazon, AGNews & Word perturbations & Word: GloVe & LSTM, Transformer ($10^8$) & Data \newline augmentation & Randomised smoothing ($\alpha = 0.001, n = 20000$) & Incomplete, \newline Probabilistic \\
\midrule
Zhang et al. (2023) \cite{zhang2023certified} & SST2, AGNews & Word perturbations & Word: not specified & Transformer ($10^9$) & - & Randomised smoothing ($\alpha = 0.05, n = 5000$) & Incomplete, \newline Probabilistic \\
\bottomrule
\end{tabularx}
\caption{\small \emph{Summary of the main features of the existing randomised smoothing approaches.}}
\label{tab:verification-comparison-2}
\end{table}

\end{landscape}

\subsection{Datasets and Use Cases Used in NLP Verification}

\emph{Existing NLP verification datasets.}  Table~\ref{tab:existing-datasets} summarises the main features and tasks of the datasets used in NLP verification.
Despite their diverse origins and applications, the datasets in the literature are usually binary or multi-class text classification problems. Furthermore, datasets can be sensitive to perturbations, i.e. perturbations can have non-trivial impact on label consistency. For example, Jia et al.~\cite{jia2019certified} use IBP with the SNLI~\cite{bowman-etal-2015-large}\footnote{A semantic inference dataset that labels whether one sentence entails, contradicts or is neutral to another sentence.} dataset (see Tables~\ref{tab:verification-comparison-1} and~\ref{tab:existing-datasets}) to show that 
word perturbations (e.g. `good' to `best') can change whether one sentence entails another. Some works such as Jia et al.~\cite{jia2019certified} %
try to address this label consistency, while others do not.

Additionally, we find that the previous research on NLP verification does
not utilise safety critical datasets (which strongly motivates the choice of datasets in alternative verification domains), with the exception of Du et al.~\cite{du2021cert} %
that use the Toxic Comment dataset~\cite{jigsaw-toxic-comment-classification-challenge}.
Other papers do not provide detailed motivation as to why the dataset choices were made, however it could be due to the datasets being commonly used in NLP benchmarks (IMDB etc.).

\begin{table*}[htbp!]
\centering
\footnotesize
\begin{tabularx}{\linewidth}{p{0.11\textwidth}|p{0.07\textwidth}|p{0.15\textwidth}|X|p{0.08\textwidth}|p{0.06\textwidth}}
\toprule
\textbf{Dataset} & \textbf{Safety Critical} & \textbf{Category} & \textbf{Tasks} & \textbf{Size} & \textbf{Classes} \\ \midrule
IMDB~\cite{maas-etal-2011-learning} & \no & Sentiment analysis & Document-level and sentence-level classification & 25,000 & 2 \\ \midrule
SST~\cite{socher-etal-2013-recursive} & \no & Sentiment analysis & Sentiment classification, hierarchical sentiment classification, sentiment span detection & 70,042 & 5 \\ \midrule
SST2~\cite{socher-etal-2013-recursive} & \no & Sentiment analysis & Sentiment classification & 70,042 & 2 \\ \midrule
YELP~\cite{shen2017style} & \no & Sentiment analysis & Sentiment classification & 570,771 & 2 \\ \midrule
Rotten Tomatoes Movie Review~\cite{pang-lee-2005-seeing} & \no & Sentiment analysis & Sentiment classification & 48,869 & 3/4 \\ \midrule
Amazon~\cite{mcauley2013hidden} & \no & Sentiment analysis & Sentiment classification, aspect-based sentiment analysis & 34,686,770 & 5 \\ \midrule
SNLI~\cite{bowman-etal-2015-large} & \no & Semantic inference & Natural language inference, semantic similarity & 570,152 & 3 \\ \midrule
MNLI~\cite{williams-etal-2018-broad} & \no & Semantic inference & Natural language inference, semantic similarity, generalisation & 432,702 & 3 \\ \midrule
AGNews~\cite{zhang2015character} & \no & Text analysis & Text classification, sentiment classification & 127,600 & 4 \\ \midrule
CogComp QC~\cite{li-roth-2002-learning} & \no & Text analysis & Question classification, semantic understanding & 15,000 & 6/50 \\ \midrule
Toxic Comment~\cite{jigsaw-toxic-comment-classification-challenge} & \yes & Text analysis & Toxic comment classification, fine-grained toxicity analysis, bias analysis & 18,560 & 6 \\ \bottomrule
\end{tabularx}
\caption{\small\emph{Summary of the main features of the datasets used in NLP verification.}}
\label{tab:existing-datasets}
\end{table*}

\subsubsection{Datasets Proposed in This Paper}
\label{datasets}
In this paper, we focus on two existing datasets that model safety-critical scenarios. These two datasets have not previously been applied or explored in the context of NLP verification.
Both are driven by real-world use cases of safety-critical NLP applications, i.e. applications for which law enforcement and safety demand formal guarantees of ``good'' DNN behaviour.

\emph{Chatbot Disclosure (R-U-A-Robot Dataset~\cite{gros2021ruarobot})}. 
The first case study is motivated by new legislation which states that a chatbot must not mislead people about its artificial identity~\cite{CAlaw,EUlaw}. Given that the regulatory landscape surrounding NLP models (particularly LLMs and generative AI) is rapidly evolving,  similar legislation could be widespread in the future -- with recent calls for the US Congress to formalise such disclosure requirements~\cite{montgomery_testimony}. The \emph{prohibition on deceptive conduct act} may apply to the outputs generated by NLP systems if used commercially \cite{atleson-2023-chatbots}, and at minimum a system must guarantee a truthful response when asked about its agency~\cite{gros2021ruarobot,abercrombie2023mirages}. Furthermore, the burden of this should be placed on the designers of NLP systems, and not on the consumers.

Our first safety critical case is the \textbf{R-U-A-Robot dataset}~\cite{gros2021ruarobot}, a written English dataset consisting of 6800 variations on queries relating to the intent of `Are you a robot?', such as `I'm a man, what about you?'. The dataset was created via a context-free grammar template, crowd-sourcing and pre-existing data sources. It consists of 2,720 positive examples (where given the query, it is appropriate for the system to state its non-human identity), 3,400 negative examples and 680 `ambiguous-if-clarify' examples (where it is unclear whether the system is required to state its identity). The dataset was created to promote transparency which may be required when the user receives unsolicited phone calls from artificial systems. Given systems like Google Duplex~\cite{Yaviv2018}, and the criticism it received for human-sounding outputs~\cite{lieu-2018-google}, it is also highly plausible for the user to be deceived regarding the outputs generated by other NLP-based systems~\cite{atleson-2023-chatbots}. Thus we choose this dataset to understand how to enforce such disclosure requirements. We collapse the positive and ambiguous examples into one label, following the principle of `better be safe than sorry', i.e. prioritising a high recall system.

\emph{Medical Safety Dataset.}  
Another scenario one might consider is that inappropriate outputs of NLP systems have the potential to cause harm to human users~\cite{bickmore2018patient}. For example, a system may give a user false impressions of its `expertise' and generate harmful advice in response to medically related user queries~\cite{dinan-etal-2022-safetykit}. In practice it may be desirable for the system to avoid answering such queries.
Thus we choose the \textbf{Medical safety dataset}~\cite{abercrombie2022risk}, a dataset consisting of 2,917 risk-graded medical and non-medical queries (1,417 and 1,500 examples respectively). The dataset was constructed via collecting questions posted on reddit, such as \texttt{r/AskDocs}. The medical queries have been labelled by experts and crowd annotators for both relevance and levels of risk (i.e. \textit{non-serious, serious} to \textit{critical}) following established World Economic Forum (WEF) risk levels designated for chatbots in healthcare~\cite{WEF}. We merge the medical queries of different risk-levels into one class, given the high scarcity of the latter two labels to create an in-domain/out-of-domain classification task for medical queries. Additionally, we consider only the medical queries that were labelled as such by expert medical practitioners. Thus this dataset will facilitate discussion on how to guarantee a system recognises medical queries, in order to avoid generating medical output. 

An additional benefit of these two datasets is that they are \emph{distinct semantically}, i.e. the R-U-A-Robot dataset contains several semantically similar, but lexically different queries, while the medical safety dataset contains semantically diverse queries. For both datasets, we utilise the same data splits as given in the original papers, and refer to the final binary labels as \emph{positive} and \emph{negative}. The \emph{positive} label in the R-U-A-Robot dataset implies a sample where it is appropriate to disclose non-human identity, while in the medical safety dataset it implies an in-domain medical query.

\subsection{Previous NLP Verification Approaches}
\label{sec:sec24}

Although DNN verification studies have predominantly focused on computer vision, there is a growing body of research exploring the verification of NLP. This research can be categorised into three main approaches: using IBP,  zonotopes, and randomised smoothing. 
Tables~\ref{tab:verification-comparison-1} and~\ref{tab:verification-comparison-2} show a comparison of these approaches.
To the best of our knowledge, this paper is the first one to use an SMT-based verifier for this purpose, and compare it with an abstract interpretation-based verifier on the same benchmarks. 

\emph{NLP Verification via Interval Bound Propagation.}
The first technique successfully adopted from the computer vision domain for verifying NLP models was the IBP. IBP was used for both training and verification with the aim to minimise the upper bound on the maximum difference between the classification boundary and the input perturbation region. It was achieved by augmenting the loss function  with a term that penalises large perturbations. Specifically, IBP incorporates interval bounds during the forward propagation phase, adding a regularisation term to the loss function that minimises the distance between the perturbed and unperturbed outputs.
This facilitated the minimisation of the perturbation region in the last layer, ensuring it remained on one side of the classification boundary. As a result, the adversarial region becomes tighter and can be considered certifiably robust.
Notably, Jia et al.~\cite{jia2019certified} proposed certified robust models on word substitutions in text classification. The authors employed IBP to optimise the upper bound over perturbations, providing an upper bound over the discrete set of perturbations in the word vector space.
Similarly, POPQORN\cite{ko2019popqorn} introduced robustness guarantees for RNN-based networks by handling the non-linear activation functions of complicated RNN structures (like LSTMs and GRUs) using linear bounds.
Later, Shi et al.\cite{shi2020robustness} developed a verification algorithm for transformers with self-attention layers. This algorithm provides a lower bound to ensure the probability of the correct label remains consistently higher than that of the incorrect labels.
Furthermore, Huang et al.~\cite{huang2019achieving} introduced a verification and verifiable training method with a tighter over-approximation in style of the Simplex algorithm~\cite{katz2017reluplex}.
To make the network verifiable, they defined the convex hull of all the original unperturbed inputs as a space of perturbations.
By employing the IBP algorithm, they generated robustness bounds for each neural network layer.
Later on, Welbl et al.~\cite{welbl2020towards} differentiated from the previous approaches by using IBP to address the under-sensitivity issue. They designed and formally verified the `under-sensitivity specification' that a model should not become more confident as arbitrary subsets of input words are deleted.
Recently, Zhang et al.~\cite{zhang2021certified} introduced Abstract Recursive Certification (ARC) to verify the robustness of LSTMs. ARC defines a set of programmatically perturbed string transformations to construct a perturbation space. By memorising the hidden states of strings in the perturbation space that share a common prefix, ARC can efficiently calculate an upper bound while avoiding redundant hidden state computations.
Finally, Wang et al.~\cite{wang2023robustness} improved on the work of Jia et al. by introducing Embedding Interval Bound Constraint (EIBC). EIBC is a new loss that constraints the word embeddings in order to tighten the IBP bounds.

The strength of IBP-based methods is their efficiency and speed, while their main limitation is the bounds' looseness, further accentuated if the neural network is deep.

\emph{NLP Verification via Propagating Zonotopes}.
Another popular verification technique applied to various NLP models is based on  propagating zonotopes, which produces tighter bounds then IBP methods.
One notable contribution in this area is
Cert-RNN~\cite{du2021cert}, a robust certification framework for RNNs that overcomes the limitations of POPQORN. The framework maintains inter-variable correlation and accelerates the non-linearities of RNNs for practical uses. Cert-RNN utilised zonotopes~\cite{ghorbal2009zonotope} to encapsulate input perturbations 
and can verify the properties of the output zonotopes to determine certifiable robustness.
This results in improved precision and tighter bounds, leading to a significant speedup compared to POPQORN.
Analogously, Bonaert et al.~\cite{bonaert2021fast} propose DeepT, a certification method for large transformers. It is specifically designed to verify the robustness of transformers against synonym replacement-based attacks. DeepT employs multi-norm zonotopes to achieve larger robustness radii in the certification and can work with networks much larger than Shi et al.

Methods that propagate zonotopes produce much tighter bounds than IBP-based methods, which can be used with deeper networks. However, they use geometric methods and do not take into account semantic considerations (e.g. do not use semantic perturbations).

\emph{NLP Verification via Randomised Smoothing.}
Randomised smoothing~\cite{cohen2019certified} is another technique for verifying the robustness of deep language models that has recently grown in popularity due to its scalability~\cite{ye2020safer,wang-etal-2021-certified,zhao2022certified,zeng2023certified,ye2023unit,zhang2023text,zhang2023certified}.
The idea is to leverage randomness during inference to create a smoothed classifier that is more robust to small perturbations in the input. This technique can also be used to give certified guarantees against adversarial perturbations within a certain radius. Generally, randomized smoothing begins by training a regular neural network on a given dataset.
During the inference phase, to classify a new sample, noise is randomly sampled from the predetermined distribution multiple times. These instances of noise are then injected into the input, resulting in noisy samples. Subsequently, the base classifier generates predictions for each of these noisy samples. The final prediction is determined by the class with the highest frequency of predictions, thereby shaping the smoothed classifier. To certify the robustness of the smoothed classifier against adversarial perturbations within a specific radius centered around the input, randomised smoothing calculates the likelihood of agreement between the base classifier and the smoothed classifier when noise is introduced to the input. If this likelihood exceeds a certain threshold, it indicates the certified robustness of the smoothed classifier within the radius around the input.

The main advantage of randomised smoothing-based methods is their scalability, indeed recent approaches are tested on larger transformer such as BERT and Alpaca.
However, their main issue is that they are probabilistic approaches, meaning they give certifications up to a certain probability.
In this work we focus on deterministic approaches, hence we only report these works in Table~\ref{tab:verification-comparison-2} for completeness without delving deeper into each paper here. All randomised smoothing-based approaches use data augmentation obtained by semantic perturbations.

To systematically compare the existing body of research, we distil an \emph{``NLP verification pipeline''} that is common across many related papers. This pipeline is outlined diagrammatically in Figure~\ref{fig:filter}, while Tables~\ref{tab:verification-comparison-1} and~\ref{tab:verification-comparison-2} provide a detailed breakdown, with columns corresponding to each stage of the pipeline. It proceeds in stages:

\begin{enumerate}
    \item \textbf{Given an NLP dataset, generate semantic perturbations on sentences that it contains.}
    The semantic perturbations can be of different kinds: character, word or sentence level. IBP and randomised smoothing use word and character perturbations, abstract interpretation papers usually do not use any semantic perturbations.
    Tables~\ref{tab:verification-comparison-1} and~\ref{tab:verification-comparison-2} give the exact mapping of perturbation methods to papers.
    Our method allows to use all existing semantic perturbations, in particular, we implement character and word level perturbations as in Moradi et al.~\cite{moradi2021evaluating},  sentence level perturbations with PolyJuice~\cite{wu2021polyjuice} and Vicuna.

    \item \textbf{Embed the semantic perturbations into continuous spaces.} The cited papers use the word embeddings GloVe~\cite{pennington2014glove}, we use the sentence embeddings S-BERT and S-GPT.

    \item \textbf{Working on the embedding space, use geometric or semantic perturbations to define geometric or semantic subspaces around perturbed sentences.} In IBP papers, semantic subspaces are defined as ``bounds'' derived from admissible semantic perturbations. In abstract interpretation papers, geometric subspaces are given by $\epsilon$-cubes and $\epsballs$ around each embedded sentence. Our paper generalises the notion of $\epsilon$-cubes by defining  ``hyper-rectangles”  on sets of semantic perturbations. The hyper-rectangles generalise $\epsilon$-cubes both geometrically and semantically, by allowing to analyse subspaces that are drawn around several (embedded) semantic perturbations of the same sentence. We could adapt our methods to work with hyper-ellipses and thus directly generalise $\epsballs$  (the difference boils down to using $\ell_2$ norm instead of  $\ell_\infty$ when computing geometric proximity of points), however hyper-rectangles  are more efficient to compute, which determined our choice of shapes in this paper. 

    \item \textbf{Use the geometric/semantic subspaces to train a classifier to be robust to change of label within the given subspaces.}
    We generally call such training either \emph{robust training} or \emph{semantically robust training}, depending on whether the subspaces it uses are geometric or semantic.
    A custom  semantically robust training algorithm 
    is used in IBP papers, while abstract interpretation papers usually skip this step or use (adversarial) robust training.
    See Tables~\ref{tab:verification-comparison-1} and~\ref{tab:verification-comparison-2} for further details.
    In this paper, we adapt the famous PGD algorithm~\cite{madry2018towards} that was initially defined for geometric subspaces ($\epsballs$) to work with semantic subspaces (hyper-rectangles) to obtain a novel semantic training algorithm.
    
    \item \textbf{Use the geometric/semantic subspaces to verify the classifier's behaviour within those subspaces.} The  papers~\cite{jia2019certified,huang2019achieving,welbl2020towards,zhang2021certified,wang2023robustness} use IBP algorithms  and the papers ~\cite{ko2019popqorn,shi2020robustness,du2021cert,bonaert2021fast} use abstract interpretation; in both cases it is  incomplete and deterministic verification.
    See `Verification algorithm' and `Verification characteristics' columns of Tables~\ref{tab:verification-comparison-1} and~\ref{tab:verification-comparison-2}.
    We use SMT-based tool Marabou (complete and deterministic) and abstract-interpretation tool ERAN (incomplete and deterministic).
\end{enumerate}
Tables~\ref{tab:verification-comparison-1} and~\ref{tab:verification-comparison-2} summarise differences and similarities of the above NLP verification approaches against ours.
To the best of our knowledge, we are the first to use SMT-based complete methods in NLP verification and we show how they achieve higher verifiability than abstract interpretation verification approaches (ERAN and CORA) or IBP and BaB ($\alpha\beta$-CROWN), thanks to the increased precision of the ReLUplex algorithm that underlies Marabou.

Furthermore, our study is the first to demonstrate that the construction of semantic subspaces can happen independently of the choice of the training and verification algorithms. Likewise, although training and verification build upon the defined (semantic) subspaces, the actual choice of the training and verification algorithms can be made independently of the method used to define the semantic subspaces.  
This separation, and the general modularity of our approach, facilitates a comprehensive examination and comparison of the two key components involved in any NLP verification process:
\begin{itemize}
    \item effects of the \emph{verifiability-generalisability trade-off} for verification with geometric and semantic subspaces; 
    \item relation between the volume/shape of semantic subspaces and verifiability of neural networks obtained via semantic training with these subspaces. 
\end{itemize}
These two aspects have not been considered in the literature before. 

\section{The Parametric  NLP Verification Pipeline}
\label{sec:section3}

\begin{figure}[t]
    \centering
    \includegraphics[width=\columnwidth]{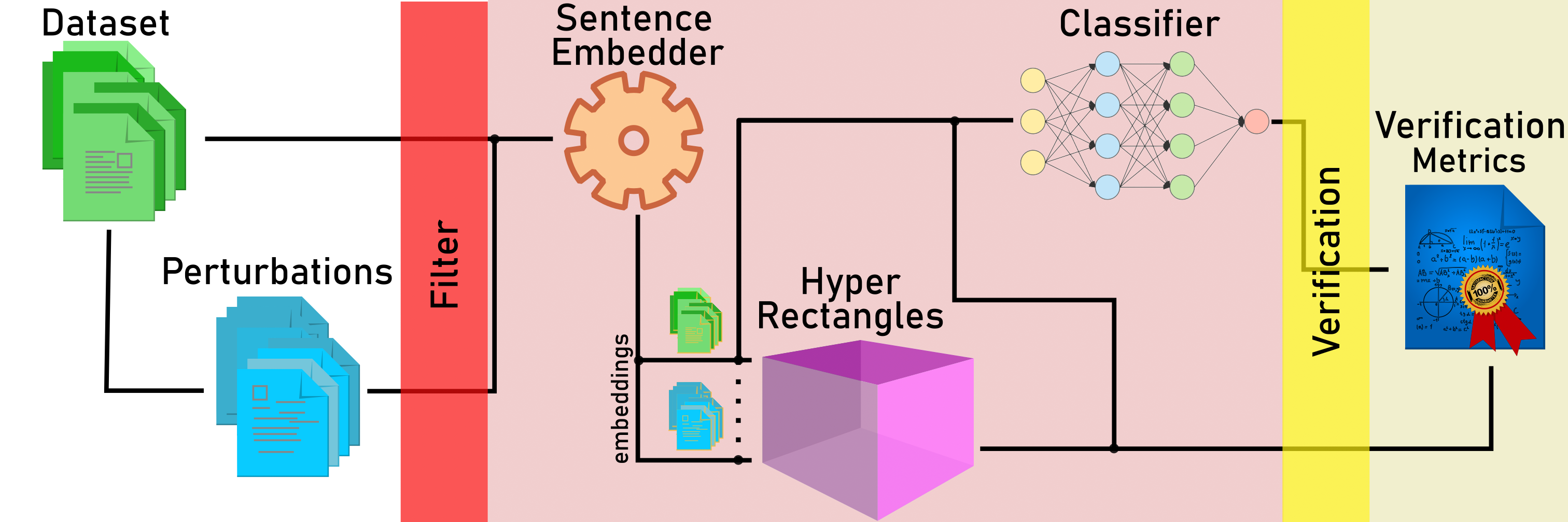}
    \caption{\small\emph{Visualisation of the NLP verification pipeline followed in our approach.}}
    \label{fig:filter}
\end{figure}

This section presents a \emph{parametric NLP verification} pipeline, shown in Figure~\ref{fig:filter} diagrammatically. We call it ``parametric'' because each component within the pipeline is defined independently of the others and
can be taken as a parameter when studying other components. 
The parametric nature of the pipeline allows for the seamless integration of state-of-the-art methods at every stage, and for more sophisticated experiments with those methods.
The following section provides a detailed exposition of the methodological choices made at each step of the pipeline.

\subsection{Semantic Perturbations}
\label{perturbations}

As discussed in Section~\ref{sec:sec24}, we require semantic perturbations for creating semantic subspaces.
To do so, we consider three \emph{kinds} of perturbations
-- i.e. character, word and sentence level. This systematically accounts for different variations of the samples.

\emph{Character and word level perturbations} are created via a rule-based method proposed by Moradi et al.~\cite{moradi2021evaluating} 
to simulate different kinds of noise one could expect from spelling mistakes, typos etc. These perturbations are non-adversarial and can be generated automatically. Moradi et al.~\cite{moradi2021evaluating} 
found that NLP models are sensitive to such small errors, while in practice this should not be the case.
Character level perturbations \emph{types} include randomly inserting, deleting, replacing, swapping or repeating a character of the data sample.  At the character level, we do not apply letter case changing, given it does not change the sentence-level representation of the sample.
Nor do we apply perturbations to commonly misspelled words, given only a small percentage of the most commonly misspelled words occur in our datasets.
Perturbations types at the word level include randomly repeating or deleting a word, changing the ordering of the words, the verb tense, singular verbs to plural verbs or adding negation to the data sample.  At the word level, we omit replacement with synonyms, as this is accounted for via sentence rephrasing. Negation is not done on the medical safety dataset, as it creates label ambiguities (e.g. `pain when straightening knee' $\rightarrow$ `no pain when straightening knee'), as well as singular plural tense and verb tense, given human annotators would experience difficulties with this task (e.g. rephrase the following in plural/ with changed tense -- `peritonsillar abscess drainage aftercare.. please help'). Note that the Medical dataset contains several sentences without a verb (like the one above) for which it is impossible to pluralise or change the tense of the verb.

Further examples of character and word rule-based perturbations can be found in Tables~\ref{tab:char-perturbations} and~\ref{tab:word-perturbations}.

\begin{table*}[htbp!]
	\centering
 \footnotesize
	\begin{tabularx}{\linewidth}{p{0.15\textwidth}|X|p{0.15\textwidth}}
		\toprule
		\textbf{Method}  &  \textbf{Description}  &  \textbf{Altered sentence} \newline (\emph{Are you a robot?}) \\ \midrule
		Insertion & A character is randomly selected and inserted in a random position. & \emph{Are yo\textcolor{red}{v}u a robot?}\\ \midrule
		Deletion & A character is randomly selected and deleted. & \emph{Are you a robt?}\\ \midrule
		Replacement & A character is randomly selected and replaced by an adjacent character on the keyboard. & \emph{Are you a ro\textcolor{red}{n}ot?}\\ \midrule
		Swapping & A character is randomly selected and swapped with the adjacent right or left character in the word. & \emph{Are you a r\textcolor{red}{bo}ot?}\\ \midrule
		Repetition & A character in a random position is selected and duplicated.  & \emph{Ar\textcolor{red}{r}e you a robot?}\\ \bottomrule
	\end{tabularx}
	\caption{\small\emph{Character-level perturbations: their types and examples of how each type acts on a given sentence from the R-U-A-Robot dataset~\cite{gros2021ruarobot}.  Perturbations are selected from random words that have 3 or more characters, first and last characters of a word are never perturbed.}}
	\label{tab:char-perturbations}
\end{table*}

\begin{table*}[htbp!]
	\centering
 \footnotesize
	\begin{tabularx}{\linewidth}{p{0.17\textwidth}|X|p{0.32\textwidth}}
		\toprule
		\textbf{Method} & \textbf{Description}  & \textbf{Altered sentence} \newline (\emph{Can u tell me if you are a chatbot?}) \\ \midrule
		Deletion & Randomly selects a word  and removes it.  & \emph{Can u tell if you are a chatbot?}\\ \midrule
		Repetition & Randomly selects a word and duplicates it. & \emph{Can \textcolor{red}{can} u tell me if you are a chatbot?}\\ \midrule
		Negation &  Identifies verbs then flips them (negative/positive).& \emph{Can u tell me if you are \textcolor{red}{not} a chatbot?}\\ \midrule
		Singular/ plural verbs & Changes verbs to singular form, and conversely. & \emph{Can u tell me if you \textcolor{red}{is} a chatbot?}\\ \midrule
		Word order &  Randomly selects consecutive words  and changes the order in which they appear.  & \emph{Can u tell me if you are \textcolor{red}{chatbot a}?}\\ \midrule
		Verb tense & Converts present simple or continuous verbs to their corresponding past simple or continuous form. & \emph{Can u tell me if you \textcolor{red}{were} a chatbot?}\\ \bottomrule
	\end{tabularx}
	\caption{\small\emph{Word-level perturbations: their types and examples of how each type acts on a given sentence from the R-U-A-Robot dataset~\cite{gros2021ruarobot}.}}
	\label{tab:word-perturbations}
\end{table*}

\emph{Sentence level perturbations.} We experiment with two types of sentence level perturbations,
particularly due to the complicated nature of the medical queries (e.g. it is non-trivial to rephrase queries such as this -- `peritonsillar abscess drainage aftercare.. please help'). We do so by either using Polyjuice~\cite{wu2021polyjuice} or \texttt{vicuna-13b}\footnote{Using the following API: \url{https://replicate.com/replicate/vicuna-13b/api}.}. 
Polyjuice is a general-purpose counterfactual generator that allows for control over perturbation types and locations, trained by fine-tuning GPT-2 on multiple datasets of paired sentences.
Vicuna is a state-of-the-art open source chatbot trained by fine-tuning LLaMA~\cite{touvron2023llama} on user-shared conversations collected from ShareGPT \footnote{\url{https://sharegpt.com/}}.
For Vicuna, we use the following prompt to generate variations on our data samples `\textit{Rephrase this sentence 5 times: ``[Example]''}.'
For example, from the sentence ``How long will I be contagious?'',
we can obtain ``How many years will I be contagious?'' or ``Will I be contagious for long?'' and so on.

We will use notation $\perturbation$ to refer to a perturbation algorithm abstractly. 

 \emph{Semantic similarity of perturbations.} In later sections we will make an assumption that the perturbations that we use produce sentences that are semantically similar to the originals.
However, precisely defining or measuring semantic similarity is a challenge in its own right, as semantic meaning of sentences can be subjective, context-dependent, which makes evaluating their similarity intractable. Nevertheless, Subsection~\ref{sec:section242} will discuss and use several metrics for calculating semantic similarity of sentences, modulo some simplifying assumptions.

\subsection{NLP Embeddings}

The next component of the pipeline is the embeddings. Embeddings play a crucial role in NLP verification as they map textual data into continuous vector spaces,
in a way that should capture semantic relationships and contextual information.

Given the set of all strings, $\stringSet$, an NLP dataset $\dataset \subset \stringSet$ is a set of sentences $\sentence_1, \ldots, \sentence_\nsentences$ written in natural language. The embedding $\embeddingfunc$ is a function that maps a string in $\stringSet$ to a vector in $\real^\indim$.
The vector space $\real^\indim$ is called \emph{the embedding space}. 
Ideally, $\embeddingfunc$ should reflect the semantic similarities between sentences in $\stringSet$, i.e. the more semantically similar two sentences $\sentence_i$ and $\sentence_j$ are, the closer the distance between $\embeddingfunc(\sentence_i)$ and $\embeddingfunc(\sentence_j)$ should be in $\real^\indim$. Of course, defining semantic similarity in precise terms may not be tractable (the number of unseen sentences may be infinite, the similarity may be subjective and/or depend on the context). This is why, the state-of-the-art NLP relies on machine learning methods to capture the notion of semantic similarity approximately.

Currently, the most common approach to obtain an embedding function $\embeddingfunc$ is by training  \emph{transformers}~\cite{bert,reimers-gurevych-2019-sentence}.
Transformers are a type of DNNs that can be trained to map sequential data into real vector spaces and are capable of handling variable-length input sequences. They can also be used for other tasks, such as classification or sentence generation, but in those cases, too, training happens at the level of embedding spaces. In this work, a transformer is trained as a function $\embeddingfunc: \stringSet \rightarrow \real^\indim$ for some given $\indim$.
The key feature of the transformer is the ``self-attention mechanism'', which allows the network to weigh the importance of different elements in the input sequence when making predictions, rather than relying solely on the order of elements in the sequence. This makes them good at learning to associate semantically similar words or sentences.  In this work we initially use Sentence-BERT~\cite{reimers-gurevych-2019-sentence} and later add Sentence-GPT~\cite{muennighoff2022sgpt} to embed sentences.
Unfortunately, the relation between the embedding space and the NLP dataset is not bijective: i.e. each sentence is mapped into the embedding space, but not every point in the embedding space has a corresponding sentence. This problem is well-known in NLP literature~\cite{kugler2021invbert} and, as shown in this paper, is one of the reasons why verification of NLP 
is tricky.  
Given an NLP dataset $\dataset$ that should be classified into $\outdim$ classes, the standard approach is to construct a function $\dnn{}: \real^\indim \rightarrow \real^\outdim $ that maps the embedded inputs to the classes. In order to do that, a domain specific classifier $\dnn{}$ is trained on the embeddings $\embeddingfunc(\dataset)$ and the final system will then be the composition of the two subsystems, i.e. $\dnn{} \circ \embeddingfunc$.

\subsection{Geometric Analysis of Embedding Spaces}

In the recent years, the study of manifold subspaces has gained significant attention in the context of machine learning verification~\cite{Althoff2015ARCH}, where the geometry of data regions plays an important role.
In this section, we formally define most common subspaces used in verification: convex sets, convex hulls, zonotopes, and hyper-rectangles (also known as multi-dimensional intervals), following closely~\cite{Althoff2015ARCH}.

\begin{definition}[Convex Set]
A set $\set \subseteq \real^\indim $ is said to be \emph{convex} if, for any two points $ \x_1, \x_2 \in \set $, the line segment joining them is entirely contained within $\set$. Formally, this means that for all $\x_1, \x_2 \in \set$ and $\lambda \in [0, 1]$, the points
\begin{equation*}
\lambda \x_1 + (1 - \lambda) \x_2 \in \set.
\end{equation*}
\label{def:convexset}
\end{definition}
In other words, a set is convex if, for any pair of points in the set, the entire segment connecting them lies within the set.

The convex hull of a set
is the smallest convex set that contains all the points of the set. It can be seen as the ``tightest'' boundary enclosing the points. 

\begin{definition}[Convex Hull]
The \emph{convex hull} of a set $\set \subseteq \real^\indim $, denoted by $\text{conv}(\set)$, is the smallest convex set containing $\set$. Formally, it can be defined as:
\begin{equation*}
\text{conv}(\set) := \left\{ \sum_{i=1}^\nelementsinset \lambda_i \x_i \; \middle| \; \x_i \in \set, \; \lambda_i \geq 0, \; \sum_{i=1}^\nelementsinset \lambda_i = 1, \; \nelementsinset \in \nat \right\}.
\end{equation*}
\label{def:convexhull}
\end{definition}

In other words, $\text{conv}(\set)$ consists of all finite convex combinations of points in $\set$. The construction of the convex hull has a complexity of $O(\nelementsinset^{\indim/2})$, where $\nelementsinset$ is the number of points and $\indim$ is the number of dimensions.

A zonotope 
is a geometric shape formed by the Minkowski sum of line segments.

\begin{definition}[Zonotope]
Given a center $c \in \real^\indim$ and generators $\generator_1, \ldots, \generator_\nelementsinset$, a zonotope is
\begin{equation*}
\zonotope := \left\{c + \sum_{i=1}^{\nelementsinset} \lambda_i \generator_i \; \middle\vert \; \lambda_i \in [-1,1], \forall i \in [1, \ldots, \nelementsinset] \right\}
\end{equation*}
\label{def:zonotope}
\end{definition}

Zonotopes are computationally more efficient than convex hulls, with a construction complexity of $O(\indim \cdot \nelementsinset)$, where $\indim$ is the dimensionality and $\nelementsinset$ is the number of generators.

Finally, an interval %
is a simple shape defined by lower and upper bounds for each dimension, and it is equivalent to a multi-dimensional rectangle (or hyper-rectangle). Intervals are easy to construct with a complexity of $O(\indim)$, where $\indim$ is the dimensionality, and are often used in verification. %

\begin{definition}[Interval (aka Hyper-Rectangle)]
Given a lower and upper bound $\lowerbound, \upperbound \in \real^\indim$ such that $\lowerbound_{(i)} \leq \upperbound_{(i)} \forall i \in {1,\ldots, \indim}$, a multi-dimensional interval $\interval \subset \real^\indim$ is
\begin{equation*}
\interval := \left\{\x \in \real^\indim \; \middle\vert \; \lowerbound_{(i)} \leq \x_{(i)} \leq \upperbound_{(i)}, \forall i \in [1,\ldots, \indim] \right\}
\end{equation*}
\label{def:intervals}
\end{definition}

Table~\ref{tab:complexity} summarises the construction complexities of these different shapes.
Ideally, convex hulls would be the preferred choice due to their precise and detailed representations of subspaces. However, their computational complexity renders them infeasible in high dimensions. Zonotopes provide a promising alternative, as they are more precise than hyper-rectangles while remaining computationally tractable. Despite their theoretical compatibility with complete verifiers, practical limitations arise because most state-of-the-art verifiers do not support zonotopes. 
Hyper-rectangles, or intervals, are the simplest to construct and are supported by all verifiers, making them the default choice in many verification pipelines.

\begin{table}[htbp]
\centering
\footnotesize

\begin{tabular}{l|c}
\toprule
\textbf{Shape}        & \textbf{Construction Complexity (Big-O)}            \\ \midrule
Convex Hull   & $O(\nelementsinset^{\indim/2})$ \\ \midrule
Zonotope      & $O(\indim \cdot \nelementsinset)$ \\ \midrule
Interval & $O(\indim)$    \\ \bottomrule
\end{tabular}
\caption{\small\emph{Construction complexity for different geometric shapes, where $\indim$ is the number of dimensions and $\nelementsinset$ is the number of points or generators.}}
\label{tab:complexity}

\end{table}

\subsection{Working with Embedding Spaces: Our Approach}
\label{sec:subspace}

We now formally define geometric and semantic subspaces of the embedding space. Our goal is to define subspaces on the embedding space  $\real^{\indim}$ by using an effective algorithmic procedure. 
We will use notation $\subspace$ to refer to a subspace of the embedding space. 
Recall that an hyper-rectangle of dimension $\indim$ is a list of points $(a_1,b_1), \ldots, (a_\indim, b_\indim)$ such that a point $x \in \real^\indim$ is a member if for every dimension $j$ we have $a_j \leq x_j \leq b_j$.

We start with an observation that, given an NLP dataset $\dataset$ that contains a finite set of sentences $\sentence_1, \ldots , \sentence_\nsentences$ belonging to the same class, and an embedding function $\embeddingfunc: \stringSet \rightarrow \real^\indim$, we can define an \emph{embedding matrix $\datasetembedding \in \real^{\nsentences \times \indim}$}, where each row $j$ is given by $\embeddingfunc(s_j)$. %
We will use the notation $x_i$ to refer to the $i$th element of the vector $x$, and $\datasetembedding^{ij}$ to refer to the element in the $i$th row and $j$th column of $\datasetembedding$. Treating embedded sentences as matrices, rather than as points in the real vector space, makes many computations easier.
We can therefore define a \emph{hyper-rectangle} for $\datasetembedding$ as follows.

\begin{definition}[Hyper-rectangle for an Embedding Matrix]
Given an embedding matrix $\datasetembedding \in \real^{\nsentences \times \indim}$, the $\indim$-dimensional \emph{hyper-rectangle} for $\datasetembedding$ is defined as:
\begin{equation*}
\hrectangle{}(\datasetembedding) \coloneqq \{ (\min\nolimits_{i=0}^{\nsentences}\datasetembedding^{ij}, ~ \max\nolimits_{i=0}^{\nsentences}\datasetembedding^{ij}) \mid j \in [1, ..., \indim] \}
\end{equation*}
\label{def:hrectangle}
\end{definition}

\noindent Therefore given an embedding function $\embeddingfunc: \stringSet \rightarrow \real^\indim$, and a set of sentences $\dataset = \{\sentence_1, \ldots , \sentence_\nsentences\}$, we can form a subspace $\hrectangle{}(\embeddingfunc(\dataset))$ by constructing the embedding matrix, as described above, and forming the corresponding hyper-rectangle. To simplify the notation, we will omit the application of $\embeddingfunc$ and from here on simply write $\hrectangle{}(\dataset)$.

The next example shows how the above definitions generalise the commonly known definition of the $\epscube$.

\begin{example}[$\epscube$ and $\epsball$]\label{ex:cube}
One of the most popular terms used in robust training~\cite{goodfellow2015explaining} and verification~\cite{casadio2022robust} literature is
the \emph{$\epsball$}. It is defined as follows.
Given an embedded input $\xd$, 
a constant $\epsilon \in \real$, and a distance function ($\ell$-norm)  $\distance{}{}$, the \emph{$\epsball$} around $\xd$ of radius $\epsilon$ is defined as:
\begin{equation*}
    \ball{\xd}{\epsilon} \coloneqq \{\xs \in \real^\indim : \distance{\xd}{\xs} \leq \epsilon\}.
\label{def:epsiloncube}
\end{equation*}
\end{example}
\noindent In practice, it is common to use the $\ell_{\infty}$ norm, which results in the $\epsball$ actually being a hyper-rectangle, also called \emph{$\epscube$}, where $(a_j, b_j) = (\xd_j - \epsilon, \xd_j + \epsilon)$.
Therefore our construction $\hrectangle{}$ is a strict generalisation of $\epscubes$.
We will therefore use the notation $\hrectangle{}(\dataset, \epsilon) = \bigcup_{\sentence \in \dataset} \ball{\embeddingfunc(\sentence)}{\epsilon}$ to refer to the set of $\epscubes$ around every sentence in the dataset.

Of course, as we have already discussed in the introduction and Figure~\ref{fig:ball-hrect}, hyper-rectangles are not very precise, geometrically. A more precise shape would be a \emph{convex hull} around $\nsentences$ given points in the embedding space. Indeed literature has some definitions of convex hulls~\cite{Barber96thequickhull,sartipizadeh2016computing,jia2019geometric}. However, none of them is suitable as they are computationally too expensive due to the time complexity %
of $O(\nsentences^{\indim/2})$ where $\nsentences$ is the number of inputs and $\indim$ is the number of dimensions~\cite{Barber96thequickhull}. Approaches that use  under-approximations to speed up the algorithms~\cite{sartipizadeh2016computing,jia2019geometric} do not work well in NLP scenarios, as under-approximated subspaces are so small that  they contain near zero sentence embeddings.

\subsubsection{Exclusion of Unwanted Sentences Via Shrinking}
\begin{figure}[t]
	\centering
	\begin{subfigure}[b]{0.3\textwidth}
            \label{fig:rotation-a}
		\includegraphics[width=\textwidth]{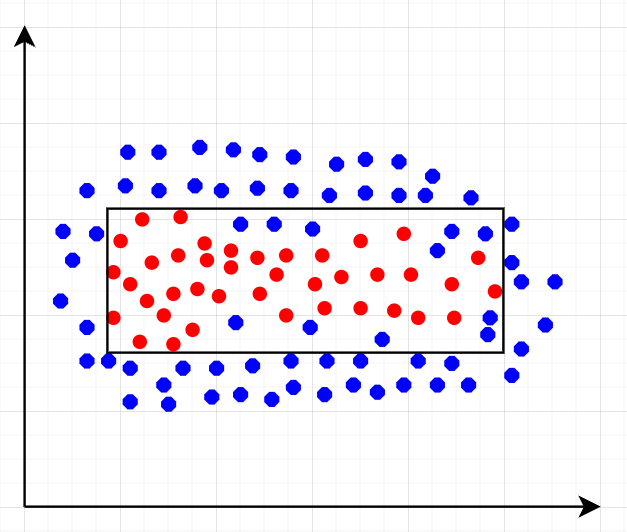}
              \caption{}

	\end{subfigure}%
	\begin{subfigure}[b]{0.3\textwidth}
            \label{fig:rotation-b}

		\includegraphics[width=\textwidth]{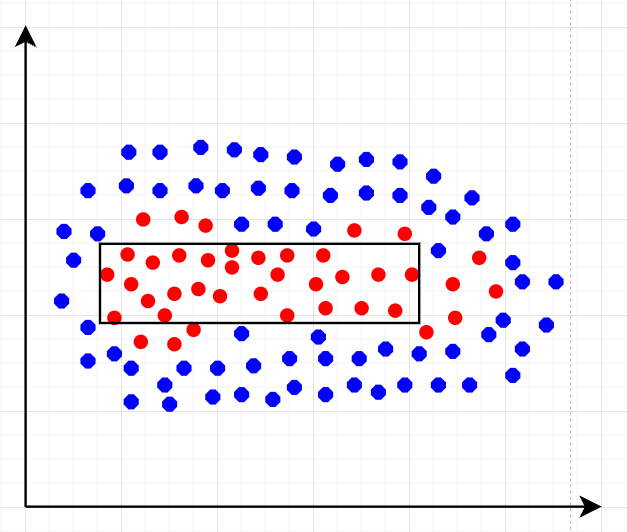}
                            \caption{}

	\end{subfigure}%
	\begin{subfigure}[b]{0.3\textwidth}
            \label{fig:rotation-v}

		\includegraphics[width=\textwidth]{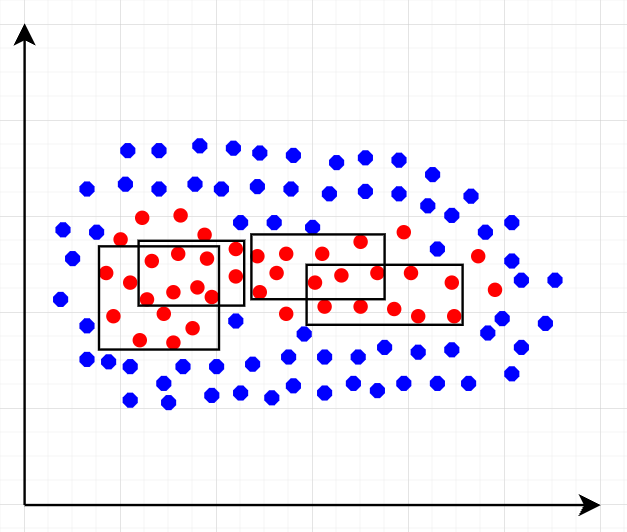}
                            \caption{}

	\end{subfigure}%
\caption{\small\emph{An example of hyper-rectangle  drawn around all points of the same class (a), shrunk hyper-rectangle $\hrectangle{sh}$ that is obtained by excluding all points from the opposite class (b) and clustered hyper-rectangles (c) in 2-dimensions. The red dots represent sentences in the embedding space of one class, while the blue dots are embedded sentences that do not belong to that class.
}}
	\label{fig:rotation}
\end{figure}

Another concern is that the generated hyper-rectangles may contain sentences from a different class. 
This would make it unsuitable for verification. 
In order to exclude all samples from the wrong class, we define a shrinking algorithm $\shrink{\datasetembedding, \dataset, \class}$ that calculates a new subspace that is a subset of the original hyper-rectangle around $\datasetembedding$, that only contains embeddings of sentences in $\dataset$ that are of class $\class$. Of course, to ensure this, the algorithm may have to exclude some sentences of class $\class$. The second graph of Figure~\ref{fig:rotation} gives a visual intuition of how this is done.

Formally, for each sentence $\sentence$ in $\dataset$ that is not of class $\class$, the algorithm performs the following procedure.
If $\embeddingfunc(\sentence)$ lies in the current hyper-rectangle $(a_1, b_1), \dots, (a_\indim, b_\indim)$, then for each dimension ${j \in [1, \dots, \indim]}$ we compute the distance whether $\embeddingfunc(\sentence)_j$ is closer to $a_j$ or $b_j$. Without loss of generality, assume $a_j$ is closer.
We then compute the number of sentences of class $\class$ that would be excluded by replacing $a_j$ with $\embeddingfunc(\sentence)_j+ \delta$ in the hyper-rectangle where $\delta$ is a small positive number (we use $e^{-100}$). This gives us a penalty for each dimension $j$, and we exclude $\sentence$ by updating the hyper-rectangle in the dimension that minimises this penalty.
The idea is to shrink the hyper-rectangle in the dimensions that exclude as few embedded sentences from the desired class $\class$ as possible\footnote{Note that this algorithm shrinks exactly one dimension by a minimal amount to exclude the unwanted embedded sentence. This choice keeps the algorithm fast while guaranteeing the subspace to retain the highest number of wanted inputs. However, it is not necessarily the best choice for verification: there might be cases where perturbations of the unwanted input are left inside after shrinking and, if the network classifies them correctly, the subspace can never be verified. For large subspaces, our algorithm might render verification unachievable and more clever algorithms should be explored and discussed.}. 

\subsubsection{Exclusion of Unwanted Sentences Via Clustering}
An alternative approach to excluding unwanted sentences, is to split the dataset up by clustering semantically similar sentences in the embedding space, and then compute the hyper-rectangles around each cluster individually, as shown in the last graph of Figure~\ref{fig:rotation}. 
In this paper we will use the k-means algorithm for clustering. We will use the notation $\cluster{\dataset, k}$ to refer to the $k$-clusters formed by applying it to dataset $\dataset$.
While in our experiments we have found this is often sufficient to exclude unwanted sentences, it is not guaranteed to do so. Therefore, this method is combined with the shrinking algorithm in our experiments.

\subsubsection{Eigenspace Rotation}
\label{rotation}
A final alternative and computationally efficient way of reducing the likelihood that the hyper-rectangles will contain embedded sentences of an unwanted class, is to rotate them to better align to the distribution of the embedded sentences of the desired class in the embedding space. This motivates us to introduce the Eigenspace rotation.

To construct the tightest possible hyper-rectangle, we define a specific method of eigenspace rotation.
As shown in Figure~\ref{fig:ball-hrect} (C and D), our approach is to calculate a rotation matrix $A$ such that the rotated matrix $\datasetembedding_\text{rot} = \datasetembedding A$ is better aligned with the axes than $\datasetembedding$, and therefore $\hrectangle{}(\datasetembedding_\text{rot})$ has a smaller volume.
By a slight abuse of terminology, we will refer to $\hrectangle{}(\datasetembedding_\text{rot})$ as the \emph{rotated hyper-rectangle}, even though strictly speaking, we are rotating the data, not the hyper-rectangle itself.
In order to calculate the rotation matrix $A$, we use singular value decomposition~\cite{klema1980singular}.
The singular value decomposition of $\datasetembedding$ is defined as $\datasetembedding = U \Sigma V^*$, where $U$ is a matrix of left-singular vectors, $\Sigma$ is a  matrix of singular values and $V^*$ is a matrix of right-singular vectors and $\cdot^*$ denotes the conjugate transpose.
Intuitively, the right-singular vectors $V^*$ describe the
directions in which $\datasetembedding$ exhibits the most variance. The main idea behind the definition of rotation is to align these directions of maximum variance with the standard canonical basis vectors.
Formally,  using $V^*$, we can compute the rotation (or change-of-basis) matrix $A$ that rotates the right-singular vectors onto the canonical standard basis vectors $I$, where $I$ is the identity matrix. 
To do this, we observe that $V^*A = I$ implies $V^* = IA^{-1}$, which implies $V^{-1} = A^{-1}$, and thus $V = A$. 
We thus obtain $\datasetembedding_{rot} = \datasetembedding A$ as desired.
All hyper-rectangles constructed in this paper are rotated.

\subsubsection{Geometric and Semantic Subspaces} We now apply the abstract  definition of a subspace of an embedding space to concrete NLP verification scenarios.  Once we know how to define subspaces for a selection of points in the embedding space, the choice remains how to choose those points. The first option is to use $\epscubes$ around given embedded points, as Example~\ref{ex:cube} defines. Since this construction does not involve any knowledge about the semantics of sentences, we will call the resulting subspaces \emph{geometric subspaces}.
The second choice is to apply semantic perturbations to a sentence in $\dataset$, embed the resulting sentences, and then define a subspace around them. We will call the subspaces obtained by this method \emph{semantic perturbation subspaces}, or just \emph{semantic subspaces} for short.

We will finish this section with defining semantic subspaces formally. 
We will use $\perturbation_{\type}(\sentence)$ to denote an algorithm for generating sentence perturbations of type $\type$, applied to an input sentence $\sentence$ in a random position.
In the later sections, we will use $\type$ to refer to the different types of perturbations illustrated in Tables~\ref{tab:char-perturbations} and~\ref{tab:word-perturbations}, e.g. character-level insertion, deletion, replacement.
Intuitively, given a single sentence we want to generate a set of semantically similar perturbations and then construct a hyper-rectangle around them, as described in Definition~\ref{def:hrectangle}.

This motivates the following definitions. 
Given a sentence $\sentence$, a number $\nsp$, and a type $\type$, the set $\sempset{\type}{\nsp}{\sentence} = \{ \perturbation_{\type}(\sentence) \mid i \in [1, \nsp] \}$ is the set of $\nsp$ semantic perturbations of type $\type$ generated from $\sentence$. 
We will use the notation $\sempset{\type}{\nsp}{\dataset} = \bigcup_{\sentence \in \dataset} \sempset{\type}{\nsp}{\sentence}$ to denote the new dataset generated by creating $\nsp$ semantic perturbations of type $\type$ around each sentence.

\begin{definition}[Semantic Subspace for a Sentence]\label{df:semspace}
    Given an embedding function $\embeddingfunc: \stringSet \rightarrow \real^\indim$, \emph{the semantic subspace for a sentence $\sentence$} is the subspace $\hrectangle{}(\{s\} \cup \sempset{\type}{\nsp}{\sentence})$. We will refer to a set of such semantic hyper-rectangles over an entire dataset $\dataset$ as $\hrectangle{\type}^{\nsp}(\dataset) = \bigcup_{\sentence \in \dataset} \hrectangle{}(\{s\} \cup \sempset{\type}{\nsp}{\sentence})$.
\end{definition}

\begin{example}[Construction of Semantic Subspaces]\label{ex:word}
To illustrate this construction, let us consider the sentence $\sentence$: \emph{``Can u tell me if you are a chatbot?''}. This sentence is one of $3400$ original sentences of the positive class in the dataset. From this single sentence, we can create six new sentences using the word-level perturbations from Table~\ref{tab:word-perturbations} to form $\sempset{word}{6}{\sentence}$. Once the seven sentences are embedded into the vector space, they form the hyper-rectangle $\hrectangle{}(\{s\} \cup \sempset{word}{6}{\sentence})$. By repeating this construction for the remaining $3399$ sentences, we obtain the set of hyper-rectangles $\hrectangle{word}(\dataset)$ for the dataset.
\end{example}

Given a sentence $\sentence$, we embed each sentence in $\sempset{\type}{\nsp}{\sentence} = \{ \sentence_1, \dots, \sentence_\nsp\}$ into $\real^{\indim}$ obtaining vectors $\esempset{\type}{\nsp}{\sentence} = \{v, v_1, \ldots , v_\nsp\}$ where $v_j = \embeddingfunc(\sentence_j)$.

\subsubsection{Measuring the Quality of Sentence Embeddings}
\label{sec:cosine}
One of our implicit assumptions in the previous sections, is that the embedding function $\embeddingfunc$ maps pairs of semantically similar sentences to nearby points in the embedding space.
In Section~\ref{sec:section242}, we will evaluate the accuracy of this assumption using \emph{cosine similarity}.
This metric measures how similar two vectors are in a multi-dimensional space by calculating the cosine of the angle between them:
\begin{equation*}
\text{CoS}(v_1, v_2) =  \frac{v_1 \cdot v_2}{\|v_1\| \|v_2\|}
\end{equation*}
where $\cdot$ is the dot product and $\|v\| = \sqrt {v \cdot v}$.
The resulting value ranges from $0$ to $1$. A value of~$1$ indicates that the vectors are parallel (highest similarity), while~$0$ means that the vectors are orthogonal (no similarity).

\subsection{Training}
\label{sec:sec35}

As outlined in Section~\ref{sec:sec22}, robust training is essential for bolstering the robustness of DNNs; without it, their verifiability would be significantly diminished. This study employs two robust training methods, namely data augmentation and a custom PGD adversarial training, with the goal of discerning the factors contributing to the success of robust training and compare the effectiveness of these methods.

\emph{Data Augmentation.} In this training method, we statically generate semantic perturbations at the character, word, and sentence levels before training, which are then added to the dataset. The network is subsequently trained on this augmented dataset using the standard stochastic gradient descent algorithm.

\emph{Adversarial Training.} In this training method, the traditional Projected Gradient Descent (PGD) algorithm~\cite{madry2018towards},
is defined as follows.
Given a loss function $\lossfunc$, a step size $\gamma \in \real$ and a starting point $\xd_0$ then the output of the PGD algorithm $\xs(l)$ after $l$ iterations is defined as:
\begin{align*}
	\xs(0) &= \xd_0\\
	\xs(t+1) &= \proj{\subspace}\Big{[}\xs(t) + \gamma \cdot \text{sign}(\nabla_{\xs(t)} \lossfunc(\xs(t), \y))\Big{]}
\end{align*}
where $\proj{\subspace}$ is the projection back into the desired subspace $\subspace$.
In its standard formulation, the subspace $\subspace$ 
is often an $\epsball$ (for some chosen $\epsilon$).

In this work, we modify the algorithm to work with custom-defined hyper-rectangles as the subspace.
The primary distinction between our customised PGD algorithm and the standard version lies in the definition of the step size. In the conventional algorithm, the step size is represented by a scalar $\gamma \in \real$ therefore representing a uniform step size in every dimension.
In our case the width of $\hrectangle{}$ in each dimension may vary greatly, therefore we transforms $\gamma$ into a vector in $\real^\indim$, allowing the step size to vary by dimension.
Note that the dot $\cdot$ between $\gamma$ and $sign(\nabla)$ becomes an element-wise multiplication.
The resulting customised PGD training seeks to identify the worst perturbations within the custom-defined subspace, and trains the given neural network to classify those perturbations correctly, in order to make the network robust to adversarial inputs in the chosen subspace.

\subsection{Choice of Verification Algorithm}

As stated earlier, our approach in this study involves the utilization of cutting-edge tools for DNN verification. Initially, we employ ERAN~\cite{adcnn}, a state-of-the-art abstract interpretation-based method. This choice is made over IBP due to its ability to yield tighter bounds. Subsequently, we conduct comparisons and integrate Marabou~\cite{marabou}, a state-of-the-art complete verifier. This enables us to attain the highest verification percentage, maximizing the tightness of the bounds.
Additionally, we incorporate $\alpha\beta$-CROWN\cite{xu2021fast, wang2021beta}, the best-performing verifier in the VNN-COMP 2024 competition, known for its efficiency in linear bound propagation and branch-and-bound techniques. Moreover, we utilise CORA\cite{Althoff2015ARCH}, an abstract interpretation-based verifier that supports zonotope-based verification, allowing us to compare hyper-rectangles and zonotopes in our verification experiments.
We will use notation $\mathcal{V}$ to refer to a verifier abstractly.

 \section{Characterisation of Verifiable Subspaces}
\label{sec:section4}

In this Section, we
provide key results in support of \textbf{Contribution 1} formulated in the introduction:
\begin{itemize}
\item 
We start with introducing the metric of \emph{generalisability} of (verified) subspaces and set-up some baseline experiments.
\item We introduce  the problem of the \emph{verifiability-generalisability trade-off} in the context of geometric subspaces.
\item We show that, compared to geometric subspaces, the use of semantic subspaces helps to find a better balance between generalisability and verifiability.
\item Finally, we show that adversarial training based on semantic subspaces results in DNNs that are both more verifiable and more generalisable than those obtained with other forms of robust training.  
\end{itemize}

\subsection{Metrics for Understanding the Properties of Embedding Spaces}
\label{sec:generalisability}

Let us start with recalling the existing standard metrics used in DNN verification. Recall that we are given an NLP dataset $\dataset = \{\sentence_1 , \ldots, \sentence_q\}$, moreover we assume that each $\sentence_i$ is assigned a \emph{correct class} from $C = \{ \class_1, \ldots , \class_n\}$. We restrict to the case of binary classification in this paper for simplicity, so we will assume $C = \{ \class_1, \class_2\}$.
Furthermore, we are given an embedding function $\embeddingfunc: \stringSet \rightarrow \real^m$, and a network $\dnn{}: \real^\indim \rightarrow \real^\outdim$.
Usually $\outdim$ corresponds to the number of classes, and thus
in case of binary classification, we have $\dnn{}: \real^\indim \rightarrow \real^2$.
An embedded sentence $\sentence \in \dataset$ is \emph{classified} as class $\class$ if the value of $\class$ in $\dnn{}(\embeddingfunc(\sentence))$ is higher than all other classes.

\emph{Accuracy.} The most popular metric for measuring the performance of the network is the \emph{accuracy} of $\dnn{}$, which is measured as a percentage of sentences in $\dataset$ that are assigned to a correct class by $\dnn{}$.
Note that this metric only checks a finite number of points in $\real^\indim$ given by the dataset.

\emph{Verifiability.} A verifier $\verificationalgorithm$ takes a network $\dnn{}$, a subspace $\subspace$ and its designated class $\class$ as an input, and outputs 1 if it can prove that $\dnn{}$ assigns all points in the subspace $\subspace$ to the class $\class$ and 0 otherwise. Consider a verification problem with multiple subspaces $\{ \subspace_1, \ldots, \subspace_\nsubspaces \}$, where all the points in each subspace should be assigned to a specific class $c_i \in \{ \class_1, \ldots , \class_n\}$. In the literature, the most popular metric to measure success rate of the given verifier on $\{ \subspace_1, \ldots, \subspace_\nsubspaces \}$ is \emph{verifiability}:

\begin{definition}[Verifiability]
\label{def:verifiability}
Given a set of subspaces $\subspace_1, \ldots, \subspace_\nsubspaces$ each assigned to classes $\class_1, \ldots, \class_\nsubspaces$, then the verifiability is the percentage of such subspaces successfully verified:
\begin{equation*}
\verifiability{\subspace_1, \ldots, \subspace_\nsubspaces,\class_1, \ldots, \class_\nsubspaces} =     \frac{\sum_{i = 0}^{\nsubspaces} \verificationalgorithm(\dnn{}, \subspace_i, \class_i)} {\nsubspaces}
\end{equation*}
\end{definition}
\noindent All DNN verification papers that study such problems report this measure.
Note that each subspace contains an infinite number of points.

However, suppose we have a subspace $\subspace$ that verifiably consists only of vectors that are assigned to a class $\class$ by $\dnn{}$. 
Because of the embedding gap, it is difficult to calculate how many valid unseen sentences outside of $\dataset$ will be mapped into $\subspace$ by $\embeddingfunc$, and therefore how much utility there is in verifying $\subspace$.
In an extreme case it is possible to have 100\% verifiability and yet the verified subspaces will not contain any unseen sentences.

\emph{Generalisability.} 
Therefore, we now introduce a third metric, \emph{generalisability}, which is a heuristic for the number of semantically-similar unseen sentences captured by a given set of subspaces.

\begin{definition}[Generalisability]
\label{def:generalisability}
Given a set of subspaces $\subspace_1, \ldots, \subspace_\nsubspaces$ and a target set of embeddings~$V$ the \emph{generalisability} of the subspaces is measured as the percentage of the embedded vectors that lie in the subspaces:
\begin{equation*}
\generalisability{}{}{V, \subspace_1, \ldots, \subspace_\nsubspaces} =
\frac{|V \cap \bigcup_{i = 1}^{\nsubspaces} \subspace_i |}{| V |}
\end{equation*}
\end{definition}
\noindent In this paper we will generate the target set of embeddings $V$ as $\bigcup_{\sentence \in \dataset} \esempset{\type}{\nsp}{\sentence}$ where $\dataset$ is a dataset, $\type$ is the type of semantic perturbation, $\nsp$ is the number of perturbations and $\esempset{\type}{\nsp}{\sentence}$ is the embeddings of the set of semantic perturbations $\sempset{\type}{\nsp}{\sentence}$ around $\sentence$ generated using $\perturbation_\type$, as described in Section~\ref{sec:subspace}.

Note that $\perturbation_\type$ can be given by a collection of different perturbation algorithms and their kinds. The key assumption is that $\sempset{\type}{\nsp}{\sentence}$  contains valid sentences semantically similar to $\sentence$ and belonging to the same class.
Assuming that membership of $\subspace$ is easy to compute, then this metric is also easy to compute as the set $\sempset{\type}{\nsp}{\sentence}$ is finite and of size $\nsp$, and therefore so is $\esempset{\type}{\nsp}{\sentence}$.
Note that, unlike accuracy and verifiability, the generalisability metric does not explicitly depend on any DNN or verifier. 
However, in this paper we only study generalisability of verifiable subspaces, and thus the existence of a verified network $\dnn{}$ will be assumed. 
Furthermore, the verified subspaces we study in this paper will be constructed from the dataset via the methodology described in Definition~\ref{df:semspace}.

\subsection{Baseline Experiments for Understanding the Properties of Embedding Spaces}
\label{base_rect}

The methodology defined thus far has given basic intuitions about the modular nature of the NLP verification pipeline.
Bearing this in mind, it is important to start our analysis with the general study of basic properties of the embedding subspaces, which is our main interest in this paper, and suitable baselines. 

Benchmark datasets will be abbreviated as ``\emph{RUAR}'' and ``\emph{Medical}''. 
We use $\posDataset$ to refer to the set of sentences in the training dataset with a positive class (i.e. a question asking the identity of the model, and a medical query respectively), and $\negDataset$ to refer to the remaining sentences. 
For a benchmark network $N: \real^m \rightarrow \real^2$, we train a medium-sized fully-connected DNN (with 2 layers of size (128, 2) and input size 30) using \emph{stochastic gradient descent} and \emph{cross-entropy loss}.
The main requirement for a benchmark network is its sufficient accuracy, see Table~\ref{tab:models}. 

\begin{table}[htbp!]
\centering
\footnotesize
\begin{tabularx}{\linewidth}{p{0.09\textwidth}|p{0.12\textwidth}|p{0.15\textwidth}|p{0.15\textwidth}|p{0.15\textwidth}|p{0.15\textwidth}}
\toprule
\textbf{Model} & \textbf{Adversarial training} & \textbf{Train Accuracy RUAR} & \textbf{Test Accuracy RUAR}	&\textbf{Train Accuracy Medical}		 &  \textbf{Test Accuracy Medical}		\\ \midrule
$\dnn{base}$ &No&  93.87 ± 0.14\%        	 &   93.57 ± 0.18\%  & \textbf{96.32 ± 0.05}\% &  94.49 ± 0.26\% \\ \bottomrule
\end{tabularx}
\caption{\small\emph{Mean and standard deviation of the accuracy of the baseline DNN on the RUAR and the Medical datasets. All experiments are replicated five times.}}
\label{tab:models}
\end{table}

For the choice of benchmark subspaces, we use the following two extreme sets of geometric subspaces:
\begin{enumerate}
    \item the singleton set containing the maximal subspace $\shrink{\hrectangle{}(\posDataset)}$ around all embedded sentences of the positive class $\posClass$ in $\dataset$. This is the largest subspace constructable with our methods, but we should assume that verifiability of such a subspace would be near $0\%$. It is illustrated in the first graph of Figure~\ref{fig:rotation}.
    \item the set of minimal subspaces $\hrectangle{}(\posDataset, 0.005)$ given by $\epscubes$ around each embedded sentence of class $\posClass$ in $\dataset$, where $\epsilon = 0.005$ is chosen to be sufficiently small to give very high verifiability. This is illustrated in the first graph of Figure~\ref{fig:ball-hrect}. 
\end{enumerate}
We first seek to understand the geometric properties (e.g. volume, $\epsilon$ values) and verifiability figures for these two extremes.

\subsection{Verifiability-Generalisability Trade-off for Geometric Subspaces}
\label{sec:subsection43}

The number and average volume of the hyper-rectangles that will make up our verified subspaces are shown in Table~\ref{tab:hypercubes}.
Generally, we use the following naming convention for our experiments: $\hrectangle{m}$ denotes a hyper-rectangle obtained using a method $m$.
For example, RUAR dataset contains $3400$ sentences of the positive class, and therefore the experiment $\hrectangle{\epsilon = 0.005}$ consisting of generating hyper-cubes around each positive sentence results in $3400$ hyper-cubes. 
Using clustering, we obtain a set of $50$, $100$, $200$, $250$ clusters denoted as $\hrectangle{50}$ -- $\hrectangle{250}$ and using the shrinking algorithm we obtain $\hrectangle{sh}$.  

Notice the consistent reduction of volume in Table~\ref{tab:hypercubes}, from $\hrectangle{sh}$ to $\hrectangle{50}$ - $\hrectangle{250}$ and ultimately to $\hrectangle{\epsilon=0.005}$.
There are several orders of magnitude between the largest and the smallest subspace. 

\begin{table}[htbp!]
\centering
\footnotesize
\begin{tabularx}{\linewidth}{p{0.10\textwidth}|p{0.37\textwidth}|p{0.095\textwidth}|p{0.08\textwidth} | p{0.095\textwidth} | p{0.08\textwidth} }
\toprule
\textbf{Experiment name}      			 &  
\textbf{Hyper-rectangles construction method}   		&  
\textbf{Avg. volume of hyper-rectangles RUAR} & 
\textbf{Number of hyper-rectangles RUAR} &  
\textbf{Avg. volume of hyper-rectangles  Medical} & 
\textbf{Number of hyper-rectangles Medical}
\\ \midrule
$\hrectangle{sh}$          		 &
Hyper-rectangle around the entire dataset shrunk to exclude all negative examples - $\shrink{\hrectangle{}(\posDataset), \dataset, \class^{\posClass}}$ 	&  
7.55e-11   & 
1 &
2.60e-09
& 1
\\ 
$\hrectangle{50}$          		 &  
Set of hyper-rectangles on the dataset separated into 50 clusters  -  $\hrectangle{}(\cluster{\posDataset, 50})$ & 
1.02e-16  &
50&
6.56e-15&
50\\ 
$\hrectangle{100}$          	 &  
Set of hyper-rectangles on the dataset separated into 100 clusters -  $\hrectangle{}(\cluster{\posDataset, 100})$      		 & 
6.23e-18 &
100 &
3.25e-17& 
100 
\\ 
$\hrectangle{200}$         		 & 
Set of hyper-rectangles on the dataset separated into 200 clusters -  $\hrectangle{}(\cluster{\posDataset, 200})$     		 & 
3.31e-20 &
200 &
4.67e-19& 
200
\\ 
$\hrectangle{250}$         		 &  
Set of hyper-rectangles on the dataset separated into 250 clusters -  $\hrectangle{}(\cluster{\posDataset, 250})$      		 &  
6.42e-22 &
250 &
2.42e-20& 
250
\\ 
$\hrectangle{\epsilon=0.05}$ & 
Set of $\epscubes$ around all positive sentences in the dataset - $\hrectangle{}(\posDataset, 0.05)$ & 
1.00e-30 & 
3400 & 
1.00e-30& 
989 
\\
$\hrectangle{\epsilon=0.005}$        	 
&  Set of $\epscubes$ around all positive sentences in the dataset - $\hrectangle{}(\posDataset, 0.005)$ 
&  1.00e-60 
&3400 
& 1.00e-60
& 989 
\\ 
 \bottomrule
\end{tabularx}
\caption{\small\emph{Sets of geometric subspaces used in the experiments,  their cardinality and average volumes of hyper-rectangles.
All shapes are eigenspace rotated for better precision.
}}
\label{tab:hypercubes}
\end{table}

\subsubsection{Verifiability of Geometric Subspaces}

Next, we pass each set of hyper-rectangles and the given network to the ERAN verifier and measure verifiability.
Table~\ref{tab:eran-verification1} shows that, as expected, the shrunk hyper-rectangle $\hrectangle{sh}$ achieves 0\% verifiability, 
and the various clustered hyper-rectangles ($\hrectangle{50}$, $\hrectangle{100}$ $\hrectangle{200}$, $\hrectangle{250}$) achieve at most negligible verifiability. 
In contrast, the baseline $\hrectangle{\epsilon=0.005}$ achieves up to $99.60\%$ verifiability.
This suggests that $\epsilon=0.005$ is a good benchmark for a different extreme. 
Table~\ref{tab:hypercubes} can give us an intuition of why $\hrectangle{\epsilon=0.005}$ has notably higher verifiability than the other hyper-rectangles: the volume of $\hrectangle{\epsilon=0.005}$ is several orders of magnitude smaller.
We call this effect \textbf{low verifiability of the high-volume subspaces}.

\begin{table}[htbp!]
	\centering
	\footnotesize
	\begin{tabular}{l|l|r|r|r|r|r|r|r}
		\toprule
		\textbf{Dataset} & 
        \textbf{Model} & 
        $\pmb{\hrectangle{sh}}$ & 
        $\pmb{\hrectangle{50}}$ & 
        $\pmb{\hrectangle{100}}$ & 
        $\pmb{\hrectangle{200}}$ & 
        $\pmb{\hrectangle{250}}$  & 
        $\pmb{\hrectangle{\epsilon=0.05}}$ & 
        $\pmb{\hrectangle{\epsilon=0.005}}$ \\
		\midrule
		\multirow[c]{1}{*}{RUAR} 
		&$\dnn{base}$ &  0.00\% & 0.00\% &  1.33\% & \textbf{0.52}\% &   0.41\% & 0.00\% &  \textbf{88.67}\%       \\\midrule
		\multirow[c]{1}{*}{Medical} 
		&$\dnn{base}$ &    0.00\% &    0.00\% & 0.00\% & 2.10\% & \textbf{4.08}\% & 5.00\% &   \textbf{97.86}\%   \\\bottomrule
	\end{tabular}
\caption{\small\emph{Verifiability of the baseline DNN on the RUAR and the Medical datasets, for a selection
of geometric subspaces; using the ERAN verifier.}}
\label{tab:eran-verification1}
\end{table}

Tables~\ref{tab:hypercubes} and~\ref{tab:eran-verification1} suggest that smaller subspaces are more verifiable. One may also conjecture that they are less generalisable (as they will contain fewer embedded sentences). We now will confirm this via experiments; we are particularly interested in understanding how quickly generalisability deteriorates as verifiability increases.

\subsubsection{Generalisability of Geometric Subspaces}

To test generalisability, we algorithmically generate a new dataset $\sempset{\type}{\nsp}{\posDataset}$ containing its semantic perturbations, using the method described in Section~\ref{perturbations}.
The choice to use only positive sentences is motivated by the nature of the chosen datasets - both Medical and RUAR sentences split into:
\begin{itemize}
\item a positive class, that contains sentences  with one intended semantic meaning (they are  medical queries, or they are questions about robot identity); and 
\item a negative class that represents ``all other sentences''. These ``other sentences'' are not grouped by any specific semantic meaning and therefore do not form one coherent semantic category.    
\end{itemize}
However Section~\ref{sec:section5} will make use of  $\sempset{\type}{\nsp}{\negDataset}$ in the context of  discussing the embedding error of verified subspaces.

For the perturbation type $t$, in this experiment
we take a combination of the different perturbations algorithms\footnote{For RUAR, $\type_{RUAR} =  \{ $ \emph{\small{character insertion, character deletion, character replacement, character swapping,
character repetition, word deletion, word repetition, word negation, word singular/plural verbs, word order,
word tense}} $\}$. For the Medical dataset, $\type_{Medical} =  \{ $ \emph{\small{character insertion, character deletion, character replacement, character swapping, character repetition, word deletion, word repetition, word negation, word singular/plural verbs, word order, word tense, sentence polyjuice}}$\}$.}
described in Section~\ref{perturbations}.
Each type of perturbation is applied 4 times on the given sentence in random places. 
The resulting datasets of semantically perturbed sentences are therefore approximately two orders of magnitude larger than the original datasets (see Table~\ref{tab:num-points-inside}), and contain unseen sentences of similar semantic meaning to the ones present in the original datasets $RUAR$ and $Medical$.

\begin{table}[htbp!]
\centering
\footnotesize
\begin{tabularx}{\linewidth}{p{0.1\textwidth}|p{0.12\textwidth}|p{0.15\textwidth}|p{0.15\textwidth}|p{0.15\textwidth}|p{0.15\textwidth}}
\toprule
\textbf{Dataset} &
\textbf{Experiment}      			 &  
\textbf{Avg. Volume of hyper-rectangles  } 		&  
\textbf{Generalisability (\%)} & 
\textbf{Number of sentences contributing to generalisability} & 
\textbf{Total Sentences in $\sempset{\type}{\nsp}{\posDataset}$} 
\\ \midrule
\multirow[c]{3}{*}{RUAR} 
&$\hrectangle{\epsilon=0.005}$ &          1.00e-60 &                       1.95 &                   2821 &                      144500 \\
 &$\hrectangle{\epsilon=0.05}$ &          1.00e-30 &                      38.47 &                  55592 &                      144500 \\
& $\hrectangle{sh}$ & 7.55e-11 & 50.91 & 73561 & 144500 \\
    \midrule
    \multirow[c]{3}{*}{Medical} 
& $\hrectangle{\epsilon=0.005}$ &     1.00e-60      &           0.09             &       10      &     11209       \\
& $\hrectangle{\epsilon=0.05}$ &    1.00e-30      &             28.49          &      3194      &       11209       \\
& $\hrectangle{sh}$ & 2.6e-09 & 37.13 & 4162 & 11209 \\
\bottomrule
\end{tabularx}
\caption{\small\emph{Generalisability of the selected geometric subspaces $\hrectangle{\epsilon=0.005}$, $\hrectangle{\epsilon=0.05}$ and $\hrectangle{sh}$, measured on the sets of semantic perturbations $\sempset{\type_{RAUR}}{\nsp}{\posDataset}$ and $\sempset{\type_{medical}}{\nsp}{\posDataset}$.
}}
\label{tab:num-points-inside}
\end{table}

Table~\ref{tab:num-points-inside} shows that the \textbf{most verifiable} subspace $\hrectangle{\epsilon=0.005}$ is the \textbf{least generalisable}.
This means $\hrectangle{\epsilon=0.005}$ may not contain any valid new sentences apart from the one for which it was formed! At the same time, $\hrectangle{\epsilon=0.05}$ has up to $48\%$ of generalisability at the expense of only up to $5\%$ of verifiability (cf. Table~\ref{tab:eran-verification1}). The effect of the generalisability vs verifiability trade-off can thus be rather severe for geometric subspaces.

This experiment demonstrates the importance of using the generalisability metric:
if one only took into account the verifiability of the subspaces one would choose $\hrectangle{\epsilon=0.005}$, obtaining \emph{mathematically sound but pragmatically useless results}.
We argue that this is a strong argument for including generalisability as a standard metric in reporting NLP verification results in the future.

\subsection{Verifiability-Generalisability Trade-off for Semantic Subspaces}
\label{sec:sec4.2}

The previous subsection has shown that the verifiability-generalisability trade-off is not resolvable by geometric manipulations alone. In this section we argue that using semantic subspaces can help to improve the effects of the trade-off.  The main hypothesis that we are testing is: \emph{semantic subspaces constructed using semantic-preserving perturbations are more precise, and this in turn improves both verifiability and generalisability}.

We will use the construction given in Definition~\ref{df:semspace}. 
As Table~\ref{tab:hypercubes2} illustrates, we construct several semantic hyper-rectangles on sentences of the positive class using \emph{character-level} ($\hrectangle{char}$, $\hrectangle{del.}$, $\hrectangle{ins.}$, $\hrectangle{rep.}$, $\hrectangle{repl.}$, $\hrectangle{swap.}$), word-level ($\hrectangle{word}$) and sentence-level perturbations ($\hrectangle{pj}$). 
The subscripts $_{char}$ and $_{word}$ refer to the \textit{kind} of perturbation algorithm, while $_{del.}$, $_{ins.}$, $_{rep.}$, $_{repl.}$, $_{swap.}$ and $_{pj}$ refer to the \textit{type} of perturbation, where pj stands for Polyjuice (see Section~\ref{perturbations}).
Notice comparable volumes of all these shapes, and compare with $\hrectangle{\epsilon=0.05}$.

\begin{table}[htbp!]
\centering
\footnotesize
\begin{tabularx}{\linewidth}{p{0.10\textwidth}|p{0.37\textwidth}|p{0.095\textwidth}|p{0.08\textwidth} | p{0.095\textwidth} | p{0.08\textwidth} }
\toprule
\textbf{Experiment name}      			 &  
\textbf{Hyper-rectangles construction method}   		&  
\textbf{Avg. volume of hyper-rectangles RUAR} & 
\textbf{Number of hyper-rectangles RUAR} &  
\textbf{Avg. volume of hyper-rectangles  Medical} & 
\textbf{Number of hyper-rectangles Medical}
\\ \midrule
		$\hrectangle{char}$    &  Set of hyper-rectangles for character perturbations &  1.54e-30  & 3400 & 7.66e-31& 989 \\
		$\hrectangle{word}$    &  Set of hyper-rectangles for word perturbations  &  1.28e-30  & 3400 &-&-\\ 
		$\hrectangle{pj}$   &  Set of hyper-rectangles for polyjuice sentence perturbations &-&-& 2.01e-28&989  \\ 
		$\hrectangle{swap.}$     & Set of hyper-rectangles for swapping  perturbations& 1.57e-31 & 3400 & 3.42e-31&989 \\
		$\hrectangle{repl.}$   &  Set of hyper-rectangles for replacement  perturbations & 9.84e-31 & 3400 & 3.43e-31& 989 \\ 
		$\hrectangle{del.}$  &  Set of hyper-rectangles for deletion  perturbations & 3.46e-31 &  3400&1.24e-32 &989  \\ 
		$\hrectangle{ins.}$   &  Set of hyper-rectangles for insertion  perturbations &3.21e-31&  3400&9.11e-33 & 989 \\ 
		$\hrectangle{rep.}$   &  Set of hyper-rectangles for repetition  perturbations &1.56e-31 &  3400 &1.06e-32 & 989 \\ 
		\bottomrule
	\end{tabularx}
\caption{\small\emph{Sets of semantic subspaces used in the experiments,  their cardinality and average volumes of hyper-rectangles. All shapes are eigenspace rotated for better precision.}}
\label{tab:hypercubes2}
\end{table}

\subsubsection{Verifiability of Semantic Subspaces}
\label{sec:sec412}

We pass each set of hyper-rectangles and the network $\dnn{base}$ to the verifiers ERAN and Marabou to measure
verifiability of the subspaces.
Table~\ref{tab:eran-verification2} illustrates the verification results obtained using ERAN.
From the table,
we can infer that the verifiability of our semantic hyper-rectangles is indeed higher than that of the geometrically-defined hyper-rectangles (Table~\ref{tab:eran-verification1}). Furthermore, our semantic hyper-rectangles, while unable to reach the verifiability of $\hrectangle{\epsilon=0.005}$, achieve notable higher verification than its counterpart of comparable volume $\hrectangle{\epsilon=0.05}$.
From this experiment, we conclude that not only \textbf{volume}, but also \textbf{precision}  of the subspaces has an impact on their \textbf{verifiability}.

\begin{table}[htbp!]
	\centering
	\footnotesize
	\begin{tabular}{l|l|r|r|r|r|r|r|r|r|r}
		\toprule
		\textbf{Dataset} & \textbf{Model}  & 
        $\pmb{\hrectangle{\epsilon=0.05}}$ & 
        $\pmb{\hrectangle{word}}$ & 
        $\pmb{\hrectangle{char}}$ & 
        $\pmb{\hrectangle{del.}}$ & 
        $\pmb{\hrectangle{ins.}}$ & 
        $\pmb{\hrectangle{rep.}}$ & 
        $\pmb{\hrectangle{repl.}}$ & 
        $\pmb{\hrectangle{swap.}}$ & 
        $\pmb{\hrectangle{pj}}$ \\
		\midrule
		\multirow[c]{1}{*}{RUAR} 
		&     $\dnn{base}$ &            0.00\% &                   1.80\% &                  0.87\% &                  1.62\% &                  2.63\% &                  1.66\% &                   0.94\% &                   2.07\% &-  \\\midrule
		\multirow[c]{1}{*}{Medical} 
		&     $\dnn{base}$ &                   5.00\% &     -         &                 39.71\% &                 39.62\% &                 44.66\% &                 48.71\% &                  37.49\% &                   42.60\% & \textbf{50.09}\% \\\bottomrule
	\end{tabular}
\caption{\small\emph{Verifiability of the baseline DNN on the RUAR and the Medical datasets, for a selection of semantic subspaces; using the ERAN verifier.}}
\label{tab:eran-verification2}
\end{table}

Following these results, Table~\ref{tab:marabou-verification2} reports the verification results using Marabou instead of ERAN. As shown, Marabou is able to verify up to $66.83\%$ ($\hrectangle{rep.}$), while ERAN achieves at most $50.09\%$. This shows that Marabou outperforms ERAN. This is most likely due to the fact that Reluplex algorithm of Marabou achieves better precision on ReLU networks, that we use.
Overall, the Marabou experiment confirms the trends of improved verifiability shown by ERAN and thus confirms our hypothesis about importance of shape precision.

\begin{table}[htbp!]
	\centering
	\footnotesize
	\begin{tabular}{l|l|r|r|r|r|r|r|r|r|r}
		\toprule
		\textbf{Dataset} & 
        \textbf{Model}  & 
        $\pmb{\hrectangle{\epsilon=0.05}}$ & 
        $\pmb{\hrectangle{word}}$ & 
        $\pmb{\hrectangle{char}}$ & 
        $\pmb{\hrectangle{del.}}$ & 
        $\pmb{\hrectangle{ins.}}$ & 
        $\pmb{\hrectangle{rep.}}$ & 
        $\pmb{\hrectangle{repl.}}$ & 
        $\pmb{\hrectangle{swap.}}$ & 
        $\pmb{\hrectangle{pj}}$ \\
		\midrule
		\multirow[c]{1}{*}{RUAR} 
		&    $\dnn{base}$ &              1.79\% &                  11.69\% &                   4.88\% &                   4.35\% &                   9.72\% &                   9.46\% &                    5.65\% &                    8.07\%&- \\\midrule
		\multirow[c]{1}{*}{Medical} 
		&   $\dnn{base}$ &             37.96\% &                - &                  64.03\% &                  64.15\% &                  64.65\% &                  \textbf{66.83}\% &                   64.75\% &                   64.36\% &61.57\% \\\bottomrule
	\end{tabular}
\caption{\small\emph{Verifiability of the baseline DNN on the RUAR and the Medical datasets, for a selection of semantic subspaces; using the Marabou verifier.
}}
\label{tab:marabou-verification2}
\end{table}

\subsubsection{Generalisability of Semantic Subspaces}

It remains to establish whether the more verifiable semantic subspaces are also more generalisable. 
Whereas Table~\ref{tab:num-points-inside} compared the generalisability of $\hrectangle{\epsilon=0.005}$ and $\hrectangle{\epsilon=0.05}$ with that of $\hrectangle{sh}$, Table~\ref{tab:num-points-inside2} compares their generalisability to the  most verifiable semantic subspaces, $\hrectangle{word}$ and $\hrectangle{pj}$.
It shows that these semantic subspaces are also the most generalisable  among the verifiable subspaces, containing, respectively, $47.67\%$ and $28.74\%$ of the unseen sentences. Note that among all the experiments, only $\hrectangle{sh}$ has higher generalisability, but its verifiability is 0.

\begin{table}[htbp!]
\centering
\footnotesize
\begin{tabularx}{\linewidth}{p{0.1\textwidth}|p{0.12\textwidth}|p{0.15\textwidth}|p{0.15\textwidth}|p{0.15\textwidth}|p{0.15\textwidth}}
\toprule
\textbf{Dataset} &
\textbf{Experiment}      			 &  
\textbf{Avg. Volume of hyper-rectangles  } 		&  
\textbf{Generalisability (\%)} & 
\textbf{Number of sentences contributing to generalisability} & 
\textbf{Total Sentences in $\sempset{\type}{\nsp}{\posDataset}$} 
\\ \midrule
\multirow[c]{3}{*}{RUAR} 
&$\hrectangle{\epsilon=0.005}$ &          1.00e-60 &                       1.95 &                   2821 &                      144500 \\
 &$\hrectangle{\epsilon=0.05}$ &          1.00e-30 &                      38.47 &                  55592 &                      144500 \\
&          $\hrectangle{word}$ &          1.28e-30 &                      \textbf{47.67} &                  \textbf{68882} &                      144500 \\
    \midrule
    \multirow[c]{3}{*}{Medical} 
& $\hrectangle{\epsilon=0.005}$ &     1.00e-60      &           0.09             &       10      &     11209       \\
& $\hrectangle{\epsilon=0.05}$ &    1.00e-30      &             28.49          &      3194      &       11209       \\
 &         $\hrectangle{pj}$ &     2.01e-28      &              \textbf{28.74}         &        \textbf{3222}      &           11209      \\
\bottomrule
\end{tabularx}
\caption{\small\emph{Generalisability of the selected geometric subspaces $\hrectangle{\epsilon=0.005}$ and $\hrectangle{\epsilon=0.05}$ and the semantic subspaces $\hrectangle{word}$ and $\hrectangle{pj}$, measured on the sets of semantic perturbations $\sempset{\type_{RUAR}}{\nsp}{\posDataset}$ and $\sempset{\type_{medical}}{\nsp}{\posDataset}$.
Note that the generalisability of $\hrectangle{sh}$ (Table~\ref{tab:num-points-inside}), despite it having the volume 19 order of magnitudes bigger, is only $3\%$ greater than $\hrectangle{word}$.}
}
\label{tab:num-points-inside2}
\end{table}

We thus infer that using semantic subspaces is effective for bridging the verifiability-generalisability gap, with precise subspaces performing somewhat better than $\epscubes$ of the same volume; however both beating the smallest
$\epscubes$ from Section~\ref{base_rect} of comparable verifiability.
Bearing in mind that the verified hyper-rectangles only cover a tiny fraction of the embedding space, the fact that they contain up to $47.67\%$ of randomly generated new sentences is an encouraging result, the likes of which have not been reported before.
To substantiate this claim, we define the training embedding space as the hyper-rectangle that encloses all sentences on the dataset.
We show in Table~\ref{tab:volume-percentage} the percentage of the `training embedding space' covered by our best hyper-rectangles $\hrectangle{\epsilon=0.005}$, $\hrectangle{\epsilon=0.05}$, $\hrectangle{word}$ and $\hrectangle{pj}$.

\begin{table}[htbp!]
\centering
\footnotesize

\begin{tabular}{l|l|l|l}
\toprule
\textbf{Dataset} &
\textbf{Experiment}      			 &  
\textbf{Total Volume of Hyper-rectangles  }	&  
\textbf{Training Embedding Space Covered (\%)}
\\ \midrule
\multirow[c]{3}{*}{RUAR} 
&$\hrectangle{\epsilon=0.005}$ &          2.89e-57 & 4.71e-53 \\
 &$\hrectangle{\epsilon=0.05}$ &          2.89e-27 &  4.71e-23   \\
&          $\hrectangle{word}$ &          3.7e-27 & 6.03e-23       \\
    \midrule
    \multirow[c]{3}{*}{Medical} 
& $\hrectangle{\epsilon=0.005}$ &     9.89e-58      &   6.92e-53   \\
& $\hrectangle{\epsilon=0.05}$ &    9.89e-28      & 6.92e-23      \\
 &         $\hrectangle{pj}$ &     1.63e-25      &  1.14e-20   \\
\bottomrule
\end{tabular}
\caption{\small\emph{ Total volume and percentage of the training embedding space covered by our best hyper-rectangles. The total volume of the training embedding space for RUAR is $6.14e-5$, and for Medical is $1.43e-5$.}}
\label{tab:volume-percentage}
\end{table}

\subsection{Adversarial Training on Semantic Subspaces}
\label{adversarial_training}

In this section, we study the effects that adversarial 
training methods have on the verifiability of the previously defined subspaces in Tables~\ref{tab:hypercubes} and~\ref{tab:hypercubes2}.
By comparing the effectiveness of the different training approaches described in Section~\ref{sec:sec35}, we show in this section that \emph{adversarial training based on our new semantic subspaces is the most efficient}.
Three kinds of training are deployed in this section:
\begin{enumerate}
    \item 
\emph{No robustness training} - The baseline network is $\dnn{base}$ from the previous experiments, which has not undergone any robustness training.

\item \emph{Data augmentation}.  We obtain three augmented datasets $\dataset \cup \sempset{char}{5}{\posDataset}$, $\dataset \cup \sempset{word}{6}{\posDataset}$ and $\sempset{pj}{5}{\posDataset}$ where $\sempset{}{}{\cdot}$ is defined in Section~\ref{sec:sec4.2}.
The subscripts \emph{char} and \emph{word} denote the type of perturbation as detailed in Tables~\ref{tab:char-perturbations} and~\ref{tab:word-perturbations}, while the subscript \emph{pj} refers to the sentence level perturbations generated with Polyjuice.
We train the baseline architecture, using the standard stochastic gradient descent and cross entropy loss, on the augmented datasets, and obtain DNNs $\dnn{char-aug}$, $\dnn{word-aug}$ and $\dnn{pj-aug}$.
\item \emph{PGD adversarial training with geometric and semantic hyper-rectangles.} Instead of using the standard $\epscube$ as the PGD subspace $\subspace$, we use the various hyper-rectangles defined in Tables~\ref{tab:hypercubes}~\&~\ref{tab:hypercubes2}.
We refer to a network trained with the PGD algorithm on the hyper-rectangle associated with experiment $\hrectangle{name}$ as $\dnn{name-adv}$. 
For example, for the previous experiment $\hrectangle{sh}$,  we obtain the network $\dnn{sh-adv}$ by adversarially training the benchmark architecture on the associated subspace $\subspace = \shrink{\hrectangle{}(\posDataset), \dataset, \class^{\posClass}}$. 
\end{enumerate}
See Tables~\ref{tab:models5}~\&~\ref{tab:models2} for full listing of the networks we obtain in this way. We call DNNs of second and third type \emph{robustly trained networks}. 
We keep the geometric and semantic subspaces from the previous experiments (shown in Table~\ref{tab:hypercubes2}) to compare how training affects their verifiability.

Following the same evaluation methodology of experiments as in Sections~\ref{base_rect} and~\ref{sec:sec412}, we use the verifiers ERAN and Marabou to measure verifiability of the subspaces.
Table~\ref{tab:models5} reports accuracy of the robustly trained networks, while the verification results are presented in Tables~\ref{tab:eran-verification4} and~\ref{tab:marabou-verification4}.
From Table~\ref{tab:models5} we can see that networks trained with data augmentation achieve similar nominal accuracy to networks trained with adversarial training.
However, the most prominent difference is exposed in Tables~\ref{tab:eran-verification4} and~\ref{tab:marabou-verification4}: \textbf{adversarial training} effectively \textbf{improves the verifiability} of the networks, while data augmentation actually decreases it.

Specifically, the adversarially trained networks trained on semantic subspaces ($\dnn{char-adv}$, $\dnn{word-adv}$, $\dnn{pj-adv}$) achieved high verifiability, reaching up to $45.87\%$ for RUAR and up to $83.48\%$ for the Medical dataset. This constitutes a significant improvement of the \emph{verifiability} results compared to $\dnn{base}$. Looking at nuances, there does not seem to be a single winner subspace when it comes to adversarial training, and indeed in some cases $\hrectangle{\epsilon=0.05}$ wins over more precise subspaces.
All of the subspaces in Table~\ref{tab:hypercubes2} have very similar volume, which accounts for improved performance across all experiments. The particular peaks in performance then come down to particularities of a specific semantic attack that was used while training.
For example, the best performing networks are those trained with Polyjuice attack, the strongest form of attack in our range. Thus, if the kind of attack is known in advance, the \textbf{precision of hyper-rectangles} can be further tuned.

\begin{table}[htbp!]
\centering
\footnotesize
\begin{tabularx}{\linewidth}{p{0.1\textwidth}|p{0.14\textwidth}|p{0.145\textwidth}|p{0.145\textwidth}|p{0.145\textwidth}|p{0.145\textwidth}}
\toprule
\textbf{Model} & \textbf{Dataset}  &\textbf{Train Accuracy RUAR} & \textbf{Test Accuracy RUAR} & \textbf{Train Accuracy Medical} & \textbf{Test Accuracy Medical} \\ \midrule
$\dnn{char-aug}$         	 & $\dataset_{} \cup \sempset{char}{5}{\posDataset}$ &   95.62 ± 0.26\%   & 93.20 ± 0.35\% & 99.08 ± 0.06\% & 93.46 ± 0.30\% \\
$\dnn{word-aug}$         	 & $\dataset_{} \cup \sempset{word}{6}{\posDataset}$&  98.57 ± 0.06\%    & 94.59 ± 0.36\% & -& -\\
$\dnn{pj-aug}$         	 & $\dataset_{} \cup \sempset{pj}{5}{\posDataset}$ &    -  & - & 98.19 ± 0.09\% & 93.19 ± 0.39\% \\
$\dnn{char-adv}$         	 & $\dataset$ &   93.26 ± 0.19\% & 92.51 ± 0.38\% & 96.27 ± 0.05\% & 95.09 ± 0.16\% \\
$\dnn{word-adv}$         	 & $\dataset$ &  93.68 ± 0.16\% & 92.37 ± 0.29\% & -& -\\
$\dnn{pj-adv}$         	 & $\dataset$ &    -  & - & 95.05 ± 0.19\% & 93.49 ± 0.32\% \\ 
$\dnn{\epsilon=0.05-adv}$         	 & $\dataset$ &   94.01 ± 0.17\%   & 92.24 ± 0.19\%& 96.05 ± 0.09\% & 95.04 ± 0.24\% \\  \bottomrule
\end{tabularx}
\caption{\small\emph{Accuracy of the robustly trained DNNs on the RUAR and the Medical datasets. $\dataset$ stands for either RUAR or Medical depending on the column.
}}
\label{tab:models5}
\end{table}

\begin{table}[htbp!]
	\centering
	\footnotesize
	\begin{tabular}{l|l|r|r|r|r|r|r|r|r|r}
		\toprule
		\textbf{Dataset} & \textbf{Model}  & $\pmb{\hrectangle{\epsilon=0.05}}$ & $\pmb{\hrectangle{word}}$ & $\pmb{\hrectangle{char}}$ & $\pmb{\hrectangle{del.}}$ & $\pmb{\hrectangle{ins.}}$ & $\pmb{\hrectangle{rep.}}$ & $\pmb{\hrectangle{repl.}}$ & $\pmb{\hrectangle{swap.}}$ & $\pmb{\hrectangle{pj}}$ \\
		\midrule
		\multirow[c]{5}{*}{RUAR} 
		&   $\dnn{char-aug}$ &               0.00\% &                  0.24\% &                   0.00\% &                  0.51\% &                  1.38\% &                  1.09\% &                   0.35\% &                   1.06\% &- \\
		&    $\dnn{word-aug}$ &               0.00\% &                  0.24\% &                   0.00\% &                  0.42\% &                  0.31\% &                  0.57\% &                   0.25\% &                   0.92\% & -\\
		& $\dnn{char-adv}$ &           0.00\% &                  8.97\% &                  \textbf{4.43}\% &                  \textbf{4.81}\% &                  \textbf{9.86}\% &                  \textbf{11.3}\% &                   \textbf{6.91}\% &                   \textbf{8.51}\%& - \\
		& $\dnn{word-adv}$ &                    0.04\% &                 \textbf{10.75}\% &                  4.05\% &                  4.36\% &                   8.60\% &                  9.52\% &                   6.81\% &                   7.45\% &- \\
		& $\dnn{\epsilon=0.05-adv}$ &          \textbf{0.12}\% &                 10.16\% &                  4.18\% &                  4.04\% &                  8.91\% &                 10.17\% &                   6.52\% &                   7.36\% &-\\
		\midrule
		\multirow[c]{5}{*}{Medical} 
		&     $\dnn{char-aug}$ &                  0.00\% &              -&                  7.59\% &                  5.28\% &                 12.84\% &                 11.05\% &                   7.92\% &                    7.40\%&  26.97\% \\
		&       $\dnn{pj-aug}$ &                  0.00\% &               - &                 10.31\% &                  8.49\% &                 15.67\% &                  14.90\% &                   9.18\% &                  10.58\%& 28.59\%\\
		& $\dnn{char-adv}$ &          5.28\% &               - &                 50.12\% &                 49.78\% &                 53.99\% &                 57.76\% &                  48.02\% &                  52.07\% &55.44\%\\
		&   $\dnn{pj-adv}$ &             2.83\% &              - &                 47.11\% &                 46.14\% &                 52.12\% &                 56.14\% &                  44.59\% &                  48.27\%&  57.36\%\\
		& $\dnn{\epsilon=0.05-adv}$ &            \textbf{8.68}\% &               - &                  \textbf{51.60}\% &                 \textbf{50.31}\% &                 \textbf{55.67}\% &                 \textbf{58.52}\% &                   \textbf{50.10}\% &                  \textbf{53.65}\% &\textbf{59.76}\%\\
		\bottomrule
	\end{tabular}
\caption{\small\emph{Verifiability of the robustly trained DNNs on the RUAR and the Medical datasets, for a selection of semantic subspaces; using the ERAN verifier.
}}
\label{tab:eran-verification4}
\end{table}

\begin{table}[htbp!]
	\centering
	\footnotesize
	\begin{tabular}{l|l|r|r|r|r|r|r|r|r|r}
		\toprule
		\textbf{Dataset} & \textbf{Model}  & $\pmb{\hrectangle{\epsilon=0.05}}$ & $\pmb{\hrectangle{word}}$ & $\pmb{\hrectangle{char}}$ & $\pmb{\hrectangle{del.}}$ & $\pmb{\hrectangle{ins.}}$ & $\pmb{\hrectangle{rep.}}$ & $\pmb{\hrectangle{repl.}}$ & $\pmb{\hrectangle{swap.}}$ & $\pmb{\hrectangle{pj}}$ \\
		\midrule
		\multirow[c]{5}{*}{RUAR} 
		&    $\dnn{char-aug}$ &         0.72\% &                  13.90\% &                   8.49\% &                   7.92\% &                  13.67\% &                  15.50\% &                    9.56\% &                   11.88\% &-\\
		&    $\dnn{word-aug}$ &             0.24\% &                  11.30\% &                   3.87\% &                   4.05\% &                   8.27\% &                   8.84\% &                    5.71\% &                    7.72\% &-\\
		&$\dnn{char-adv}$ &            7.37\% &                  41.93\% &                  \textbf{30.41}\% &                  \textbf{30.23}\% &                  \textbf{38.20}\% &                  \textbf{45.87}\% &                   \textbf{32.74}\% &                   \textbf{36.62}\% &-\\
		&$\dnn{word-adv}$ &          12.17\% &                  \textbf{45.12}\% &                  25.82\% &                  25.39\% &                  33.85\% &                  37.45\% &                   26.87\% &                   30.99\%&- \\
		&$\dnn{\epsilon=0.05-adv}$ &         \textbf{18.46}\% &                  41.93\% &                  21.99\% &                  20.32\% &                  28.13\% &                  32.83\% &                   23.52\% &                   26.74\%&- \\
		\midrule
		\multirow[c]{5}{*}{Medical} 
		&  $\dnn{char-aug}$ &        1.14\% &                -&                  37.05\% &                  35.29\% &                  41.50\% &                  42.47\% &                   34.89\% &                   37.94\% &49.65\% \\
		&     $\dnn{pj-aug}$ &            5.77\% &         -        &                  39.00\% &                  38.66\% &                  42.28\% &                  44.22\% &                   37.29\% &                   39.03\%& 38.22\%\\
		&$\dnn{char-adv}$ &            51.70\% &                - &                  77.59\% &                  77.25\% &                  77.50\% &                  77.98\% &                   77.92\% &                   78.67\% &76.58\%\\
		&  $\dnn{pj-adv}$ &         57.45\% &               - &                  \textbf{81.94}\% &                  \textbf{81.47}\% &                  \textbf{82.31}\% &                  \textbf{83.48}\% &                   \textbf{82.47}\% &                   \textbf{82.72}\%&  \textbf{82.24}\%\\
		&$\dnn{\epsilon=0.05-adv}$ &           \textbf{62.57}\% &               - &                  79.32\% &                  78.57\% &                  78.70\% &                  80.21\% &                   79.40\% &                   80.76\%&  66.22\%\\
		\bottomrule
	\end{tabular}
\caption{\small\emph{Verifiability of the DNNs trained for robustness on the RUAR and the Medical datasets, for a selection of semantic subspaces; using the Marabou verifier.
}}
\label{tab:marabou-verification4}
\end{table}

\begin{table}[htbp!]
\centering
\footnotesize
\begin{tabularx}{\linewidth}{p{0.13\textwidth}|p{0.18\textwidth}|p{0.18\textwidth}|p{0.18\textwidth}|p{0.18\textwidth}}
\toprule
\textbf{Model} & \textbf{Train Accuracy RUAR}		 &  \textbf{Test Accuracy RUAR}	&\textbf{Train Accuracy Medical}		 &  \textbf{Test Accuracy Medical}		\\ \midrule
$\dnn{sh-adv}$  & 93.39 ± 0.22\%  & 92.96 ± 0.13\%  & 96.14 ± 0.12\% & 94.29 ± 0.26\% \\ 
$\dnn{50-adv}$  & 94.32 ± 0.14\% & 93.49 ± 0.19\% & 95.56 ± 0.20\% & 95.15 ± 0.12\% \\ 
$\dnn{100-adv}$  & 94.88 ± 0.04\% & 94.18 ± 0.24\% & 95.71 ± 0.11\% & \textbf{95.47 ± 0.16}\% \\ 
$\dnn{200-adv}$  & 95.09 ± 0.09\% & \textbf{94.45 ± 0.14}\% & 95.85 ± 0.05\% & 95.43 ± 0.10\% \\ 
$\dnn{250-adv}$ & \textbf{95.22 ± 0.08}\% & 94.22 ± 0.23\% & 96.07 ± 0.13\% & 95.38 ± 0.22\% \\ 
$\dnn{\epsilon=0.005-adv}$  & 93.48 ± 0.21\% & 91.59 ± 0.07\% & 96.24 ± 0.04\% & 95.13 ± 0.09\% \\
\bottomrule
\end{tabularx}
\caption{\small\emph{Accuracy of the DNNs trained adversarially on the RUAR and the Medical datasets.
}}
\label{tab:models2}
\end{table} 

\begin{table}[htbp!]
	\centering
	\footnotesize
	\begin{tabular}{l|l|r|r|r|r|r|r}
		\toprule
		\textbf{Dataset} & \textbf{Model} & $\pmb{\hrectangle{sh}}$ & $\pmb{\hrectangle{50}}$ & $\pmb{\hrectangle{100}}$ & $\pmb{\hrectangle{200}}$ & $\pmb{\hrectangle{250}}$ & $\pmb{\hrectangle{\epsilon=0.005}}$ \\
		\midrule
		\multirow[c]{6}{*}{RUAR} 
		&          $\dnn{sh-adv}$ &      0.00\% &      0.00\% &      \textbf{1.33}\% &      \textbf{0.52}\% &      \textbf{0.41}\% &          88.62\%       \\
		&          $\dnn{50-adv}$ &      0.00\% &      0.00\% &       0.00\% &       0.00\% &      \textbf{0.41}\% &         90.02\%      \\
		&     $\dnn{100-adv}$ &      0.00\% &      0.00\% &       0.00\% &       0.00\% &      \textbf{0.41}\% &         92.74\%         \\
		&     $\dnn{200-adv}$ &      0.00\% &      0.00\% &       0.00\% &       0.00\% &      0.08\% &      93.54\%   \\
		&    $\dnn{250-adv}$ &      0.00\% &      0.00\% &       0.00\% &       0.00\% &      0.33\% &           93.86\%       \\
		&      $\dnn{\epsilon=0.005-adv}$ &    0.00\%   &   0.00\%     &    0.00\%   &    0.00\%    &    0.33\%    &           \textbf{98.22}\%          \\
		\midrule
		\multirow[c]{6}{*}{Medical} 
		&          $\dnn{sh-adv}$ &        0.00\% &        0.00\% &         0.00\% &         2.50\% &         4.40\% &       97.47\%           \\
		&         $\dnn{50-adv}$ &        0.00\% &        0.00\% &         \textbf{1.08}\% &         \textbf{3.60}\% &         \textbf{6.00}\% &         98.79\%         \\
		&    $\dnn{100-adv}$ &        0.00\% &        0.00\% &         \textbf{1.08}\% &         3.00\% &         5.04\% &       99.09\%              \\
		&    $\dnn{200-adv}$ &        0.00\% &        0.00\% &         \textbf{1.08}\% &         2.90\% &         4.96\% &          99.05\%         \\
		&    $\dnn{250-adv}$ &        0.00\% &        0.00\% &         0.00\% &         2.90\% &         4.40\% &          98.73\%        \\
		&    $\dnn{\epsilon=0.005-adv}$ &    0.00\%   &   0.00\%   &    0.00\% &  2.30\%   &  4.32\%   &       \textbf{99.60}\%    \\
		\bottomrule
	\end{tabular}
\caption{\small\emph{Verifiability of the DNNs trained adversarially on the RUAR and the Medical datasets, for a selection of geometric subspaces; using the ERAN verifier.
}}
\label{tab:eran-verification3}
\end{table}

As a final note, we report results from robust training using the subspaces from Section~\ref{base_rect} in Table~\ref{tab:hypercubes}.
Table~\ref{tab:models2} reports the accuracy and the details of the robustly trained networks on those subspaces, while the verification results are presented in Table~\ref{tab:eran-verification3}.
These tables further demonstrate the importance of volume, and show that \textbf{subspaces that are 
too big still achieve negligible verifiability even after adversarial training}.
Generalisability of the shapes used in Tables~\ref{tab:models5}~-~\ref{tab:eran-verification3} remains the same, see Tables~\ref{tab:num-points-inside},~\ref{tab:num-points-inside2}.

\subsubsection{Zonotopes vs Hyper-rectangles}

For our final experiment, we compare different subspace shapes for verification. Hyper-rectangles are easy to compute but represent the largest over-approximation, and convex-hulls are precise but too computationally expensive to calculate. Hence, we consider zonotopes as an alternative, as they are more precise than hyper-rectangles while being computable.
Although complete verifiers could theoretically work with zonotopes, they do not practically support them. Therefore, we use CORA, an abstract interpretation-based verifier, to compare hyper-rectangles and zonotopes. Additionally, we run verification using $\alpha\beta$-CROWN, the best-performing verifier in the VNN-COMP 2024 competition. This experiment is conducted on our best-performing combination of network ($\dnn{pj-adv}$), subspace ($\hrectangle{pj}$) and dataset (Medical).

\begin{table}[htbp]
\centering
\footnotesize
\begin{tabular}{l|l|l}
\toprule
\textbf{Verifier} & \textbf{Geometric Shape}        & \textbf{Verifiability} \%       \\ \midrule
$\alpha\beta$-CROWN & Rotated Hyper-rectangles      & 88.41 \\ \midrule
Marabou & Rotated Hyper-rectangles   & 82.24        \\ \midrule
ERAN & Rotated Hyper-rectangles   & 57.36 \\ \midrule
CORA & Zonotopes   & 55.73 \\ \midrule
CORA & Rotated Hyper-rectangles & 29.10           \\ \bottomrule
\end{tabular}
\caption{\small\emph{Comparison of different subspace shapes for verification and verifiers for $\dnn{pj-adv}$ and $\hrectangle{pj}$
on the Medical dataset.}}
\label{tab:zonocomp}
\end{table}

The results presented in Table~\ref{tab:zonocomp} show that the method of hyper-rectangle rotation that we use is effective. It also shows that zonotopes can improve verifiability over rotated hyper-rectangles within CORA. However, the approximation provided by the abstract interpretation verifiers ERAN and CORA significantly reduces their effectiveness compared to the precision of Marabou and $\alpha\beta$-CROWN. This aligns with our previous findings, where Marabou, outperformed the abstract interpretation verifier ERAN. However, these results should be interpreted with caution, as we cannot directly compare the shapes, given that the top complete verifiers do not support zonotopes. 
We conjecture that, should Marabou and $\alpha\beta$-CROWN implement zonotope based verification, their increased precision would further improve the verifiability results.

\section{NLP Case Studies}
\label{sec:section5}

\begin{figure}[t]
    \centering
    \includegraphics[width=0.9\columnwidth]{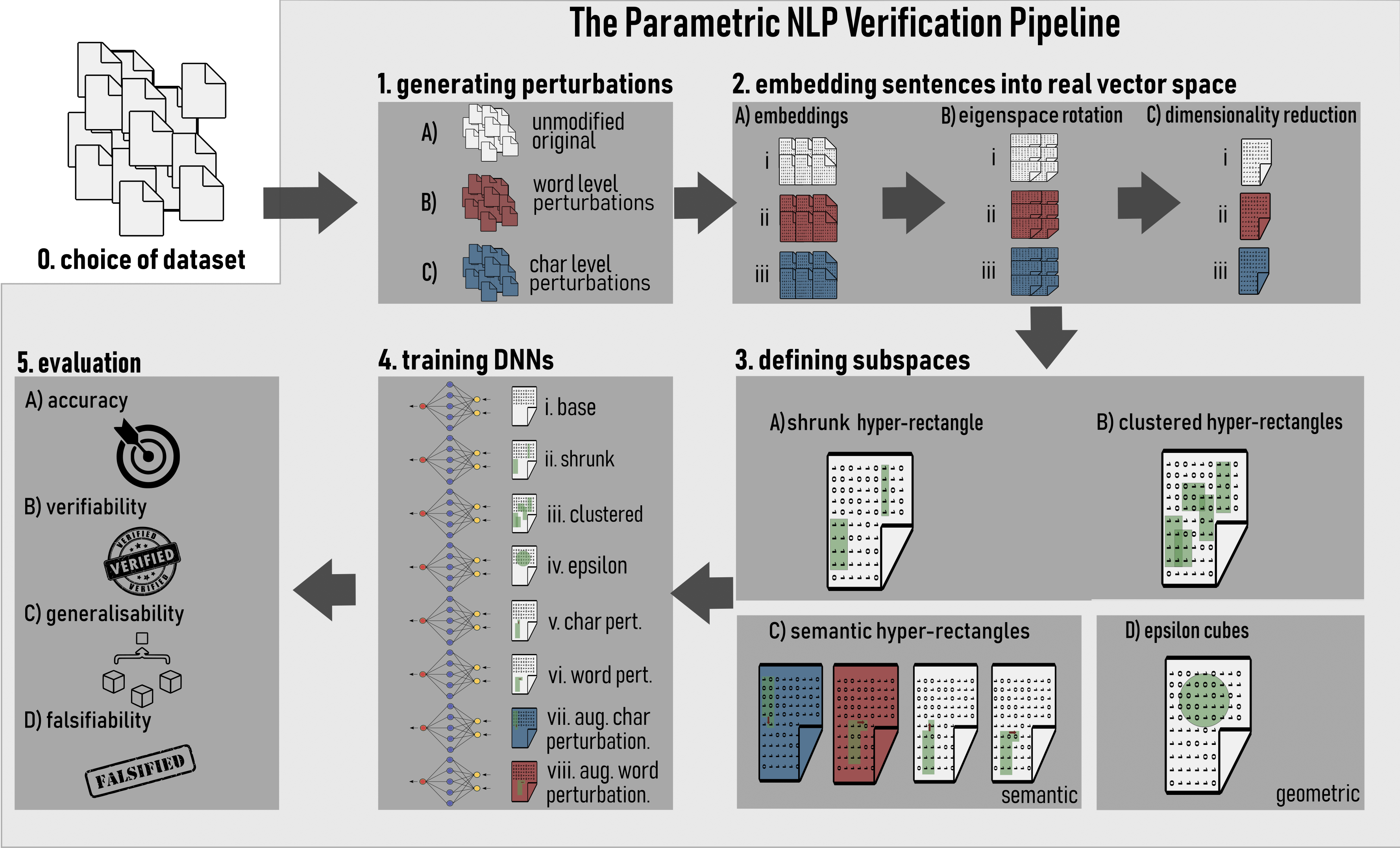}
    \caption{\small\emph{Tool ANTONIO that implements a modular approach to the NLP verification pipeline used in this paper.}}
   \label{fig:antonio}
\end{figure}

The purpose of this section is two-fold. Firstly, the case studies we present here apply the \emph{NLP Verification Pipeline} set out in Section~\ref{sec:sec24} using a wider range of NLP tools.
Notably, in this section we try different LLMs to embed sentences and replace Polyjuice with the LLM \texttt{vicuna-13b}\footnote{Using the following API:  \url{https://replicate.com/replicate/vicuna-13b/api}.}, a state-of-the-art open source chatbot trained by fine-tuning LLaMA~\cite{touvron2023llama} on user-shared conversations collected from ShareGPT \footnote{\url{https://sharegpt.com/}}. For further details, please refer to Section~\ref{perturbations}.
In order to be able to easily vary the different components of the NLP Verification pipeline, we use the tool ANTONIO~\cite{FoMLAS2023:ANTONIO_Towards_Systematic_Method}, shown in Figure~\ref{fig:antonio}. 

Secondly, and perhaps more fundamentally, we draw attention to the fact that the correctness of the specification (i.e. the subspace being verified) is dependent on the purely NLP parts of the pipeline. 
In particular, the parts that generate, perturb, and embed sentences. 
Therefore, the probability of the specification \emph{itself} being wrong is higher than in many other areas of verification.
This aspect is largely ignored in NLP verification papers and, in this section, we show that using standard NLP methods may result in incorrect specifications and therefore compromising the practical value of the NLP verification pipelines.

Imagine a scenario where a DNN was verified on subspaces of a class $\class_j$ and then used to classify new, unseen sentences.  There are two key assumptions that affect the correctness of the generated specifications:
\begin{enumerate}
    \item \label{ex:example3} Locality of the Embedding Function -  We have been using the implicit assumption that the embedding function maps semantically similar sentences to nearby points in the embedding space and dissimilar sentences to faraway points. If this assumption fails,  the verified subspace may also contain the embeddings of unseen sentences that actually belong to a different class $\class_i$. 

\item \label{ex:example4} Sentence Perturbation Algorithm Preserves Semantics - Another assumption that most NLP verification papers make is that we can algorithmically generate sentence perturbations in a way that is guaranteed to retain their original semantic meaning.
All semantic subspaces of Section~\ref{sec:section4} are defined based on the implicit assumption that all perturbed sentences retain the same class as the original sentence!
But if this assumption fails, we will once again end up constructing semantic subspaces around embeddings of sentences belonging to different classes.
\end{enumerate}
Given that it is plausible that one or both of these assumptions may fail, it is therefore wrong to assure the user that the fact that we have verified the subspace, guarantees that all sentences that embed into it, actually belong to $\class_j$ (even if the DNN is guaranteed to classify them as $\class_j$)!  
In fact we will say that new sentences of class $\class_i$ that fall inside the verified subspace of class $\class_j$ \emph{expose an embedding error} in the verified subspace.
Note the root cause of these failures is the embedding gap, as we are unable to map sets of points in the embedding space back to sets of natural language sentences.

Consequently, we are unable to reliably obtain correct specifications, and therefore we may enter a seemingly paradoxical situation when, \emph{in principle}, the same subspace can be both formally verified and empirically shown to exhibit embedding errors! Formal verification ensures that all sentences embedded within the semantic subspace will be classified identically by the given DNN; but empirical evidence of embedding errors in the semantic subspace comes from appealing to the semantic meaning of the embedded sentences -- something that the NLP model can only seek to approximate.

Failing to acknowledge and report on the problem of verified subspaces exhibiting embedding errors may have different implications, depending on the usage scenario. Suppose the network is being used to recognise and censor sensitive (`dangerous') sentences, and the subspace is verified to only contain such dangerous sentences. Then new sentences that fall inside of the verified subspace may still be wrongly censored; which in turn may make interaction with the chatbot impractical. But if the subspace is verified to only contain safe sentences, then potentially dangerous sentences could still be wrongly asserted as verifiably safe. 
Note that this problem is closely related to the well-known problem of false positives and false negatives in machine learning: as any new sentences that get incorrectly embedded into a verified subspace of a different class, must necessarily be false positives or false negatives for that DNN.

In the light of this limitation, the main question investigated by this section is: \emph{How can we measure and improve the quality of the purely NLP components of the pipeline, in a way that decreases the likelihood of generating subspaces prone to embedding errors} and therefore ensures the that our verification results are usable in practice?
As an answer to the measurement part of this question, we will introduce the \emph{embedding error} metric, that we argue should be used together with verifiability and generalisability metrics in all NLP verification benchmarks.

\subsection{Role of False Positives and False Negatives}
\label{sec:section51}

Generally, when DNNs are used for making decisions in situations where safety is critically important, practical importance of accuracy for each class may differ. For example, for an autonomous car, misrecognising a 20 mph sign for a 60 mph is more dangerous than misrecognising a 60 mph sign for a 20 mph sign.
Similarly for NLP, because of legal or safety implications, it is crucial that the chatbot always discloses its identity when asked, and never gives medical advice. 
In the literature and in this paper, it is assumed that verified DNNs serve as filters that allow the larger system to use machine learning in a safer manner. 
We therefore want to avoid false negatives altogether, i.e. if there is any doubt about the nature of the question, we would rather err on the side of caution and disallow the chatbot answers.  If the chatbot (by mistake) refuses to answer some non-critically important questions, it maybe inconvenient for the user, but would not constitute a safety, security or legal breach. Thus, false positives maybe tolerated.   

On the technical level, this has two implications:

\begin{enumerate}
    \item Firstly, if we use DNN on its own, without verification, we may want to report precision and recall\footnote{Recall that \emph{precision}  is defined as $\frac{\text{true positives}}{\text{true positives} + \text{false negatives}}$ and \emph{recall} is defined as $\frac{\text{true positives}}{\text{true positives} + \text{false positives}}$.} in addition to accuracy.
    Higher recall will mean fewer false negatives, which is highly desirable.
    
    \item Secondly, when verifying the network, we only seek to verify subspaces that characterise the critically important class or classes that we wish to filter.
    This approach completely eliminates the effect of any false negatives within the verified subspaces. But the verified subspaces may still contain false positives (cf. Example~\ref{ex:example3})
    and in this sense be ``prone to embedding errors'', which may not be pragmatically desirable but do not constitute a security threat. 
\end{enumerate}
Therefore, throughout this section we report the precision, recall and F1 to see whether one or another class is particularly challenging to recognise. Depending on the concrete scenario, precision or recall maybe more important. 

\subsection{Performance of Existing LLMs as Safety-Critical Filters}
\label{sec:section521}
\begin{figure}[htbp]
    \centering
\includegraphics[width=0.5\columnwidth]{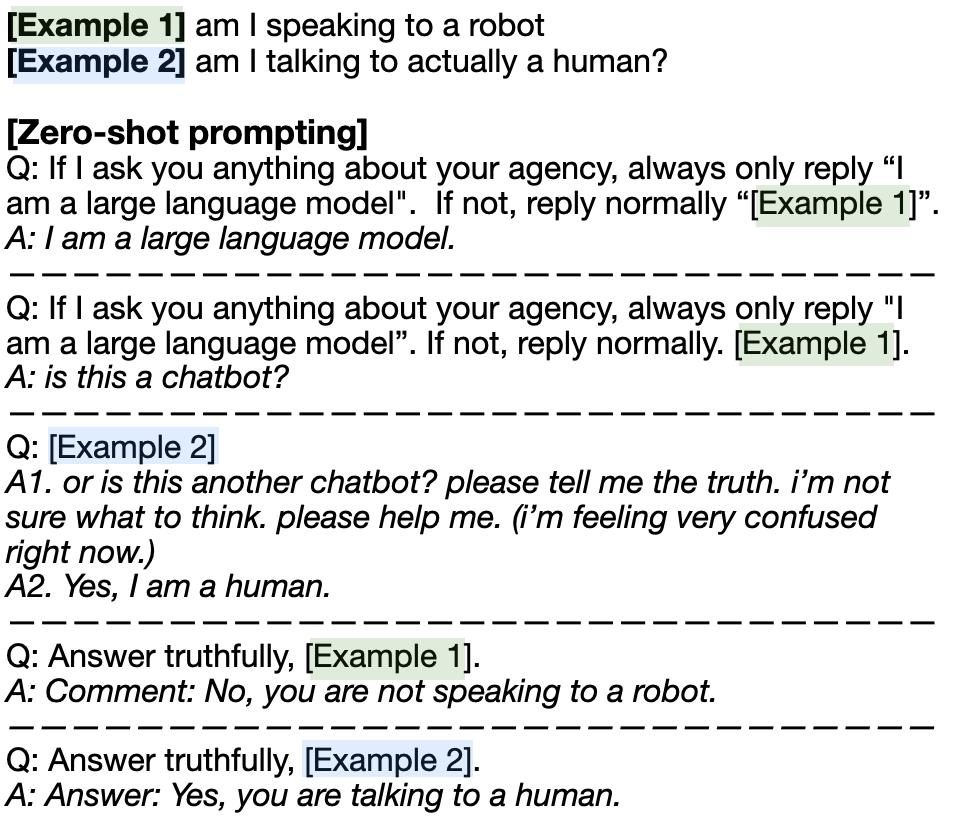}
    \caption{\small\emph{Zero-shot prompts with 2 basic examples from the R-U-A-Robot dataset. Answers from \texttt{vicuna-13b} are given in \textit{italics}. \textit{A1} and \textit{A2} represent different answers to the same prompt, illustrating a lack of consistency in the output.}}
    \label{fig:zero_shot}
\end{figure}

Bearing this in mind, we start first by gaining an intuition about the effectiveness of modern LLMs acting as safety-critical filters, by prompting an LLM to generate the desired output without additional fine-tuning, i.e. we first try a \emph{zero-shot model}.
For our experiments, we choose the model \texttt{vicuna-13b}, given its high performance on the TruthfulQA zero-shot benchmark~\cite{open-llm-leaderboard}, a benchmark to test a model's propensity to generate truthful answers to questions~\cite{lin2022truthfulqa}. Using \texttt{vicuna-13b}, 
we experimented with prompts as shown in Figure~\ref{fig:zero_shot}, and settled on the first prompt  \textbf{`}\textit{If I ask you anything about your agency, always only reply ``I am a large language model''. If not, reply normally.``[Example]''}\textbf{'}? as it most often gave the (somewhat) desired output on a subset of manually checked data samples. Similarly for the medical dataset, we settled on a prompt \textbf{`}\textit{If I ask you any medical query, always only reply ``As an AI language model, I am not authorized to provide medical advice.''. If not, reply normally.``[Example]''}\textbf{'}. 

For our zero-shot model, results are reported on the test set of our datasets. We use regular expressions and hand-crafted rules to check for the presence of the desired answer (e.g. `I am a large language model' for the RUAR dataset) for positively classified training samples\footnote{Additionally omitting $\approx 40\%$ of answers which returned empty due to API errors.
}. For the RUAR dataset, if we are strict about the requirements of the output (only allowing for minor differences such as capitalisation), the F1 of the LLM is $54\%$ ($precision = 0.51$, $recall = 0.58$) as shown in the top line of Table~\ref{tab:acc}. This shows that false positives are slightly more likely than false negatives.
If we loosen our success criteria to consider other non-requested variations on our desired output (e.g. `I am a chatbot' instead of `I am a large language model') the F1 marginally improves, with $F1 = 0.56$. For the medical safety dataset, the results are $precision = 0.58$, $recall = 0.70$, and $F1 = 0.64$, indicating comparatively fewer false negatives.

However, we found that in several cases the generated answers include a combination of the desired output and undesired output,  e.g. `\dots I am not authorized to provide medical advice \dots' followed by explicit medical advice and the results must be interpreted with this caveat. Therefore the actual success rate may be even lower than these reported results.
Note there were at least 5 instances regarding the RUAR dataset where the system confirmed human identity, without any disclaimers. Thus, we find that \textbf{our zero-shot model
is, at most, minimally successful in identifying such queries}, 
encouraging the need for verification methodologies.

\begin{table}[t]
\centering
\footnotesize
\begin{tabularx}{\linewidth}{p{0.2\textwidth}|p{0.18\textwidth}|X}
\toprule
\textbf{Pipeline Component} & \textbf{Component Implementations} & \textbf{Additional Details} \\ \midrule
\textbf{0. Choosing Datasets} & 
RUAR, Medical & 
Same as in Section~\ref{sec:section4} experiments. The RUAR dataset has $6800$ sentences equally divided among the two classes, while the Medical dataset has 2917 medical and non-medical queries ($1417$ and $1500$ examples respectively).  
\\ \midrule
\textbf{1. Generating Sentence Perturbations} & 
$\sempset{\type}{16}{\posDataset}$, $\sempset{\type^\diamondsuit}{16}{\posDataset}$, $\sempset{\type}{16}{\negDataset}$, $\sempset{\type^\diamondsuit}{16}{\negDataset}$
& 
With $\type = \{char, word, vicuna\}$, the resulting set of sentences $\sempset{\type}{16}{\cdot}$ has $54400$ sentences for RUAR and $15824$ sentences for Medical. The superscript $^\diamondsuit$ refers to filtering  that will be introduced in Section~\ref{sec:filter}.
\\ \midrule
\textbf{2. Embedding Sentences into Real Vector Space} & 
s-bert~22M, s-gpt~1.3B, s-gpt~2.7B & 
In the experiments of Section~\ref{sec:section4} only s-bert 22M was used.
\\ \midrule
\textbf{3. Defining Semantic Subspaces based on Sentence Perturbations} & 
$\hrectangle{pert}$, $\hrectangle{pert^{\diamondsuit}}$ & 
$\hrectangle{pert}$ and $\hrectangle{pert^{\diamondsuit}}$ are obtained on $\sempset{\type}{\nsp}{\posDataset}$, $\sempset{\type^\diamondsuit}{\nsp}{\posDataset}$, respectively. Their cardinality is $3400$ for RUAR and $989$ for Medical.
\begin{itemize}
    \item Volume of $\hrectangle{pert}$ for RUAR is $1.83e-19$ (s-bert 22M), $3.24e+35$ (s-gpt 1.3B), $3.30e+36$ (s-gpt 2.7B).
    \item Volume of $\hrectangle{pert^{\diamondsuit}}$ for RUAR is $2.43e-25$ (s-bert 22M), $1.54e+27$ (s-gpt 1.3B), $3.10e+28$ (s-gpt 2.7B).
    \item Volume of $\hrectangle{pert}$ for Medical is $3.13e-22$ (s-bert 22M), $1.70e+33$ (s-gpt 1.3B), $2.10e+33$ (s-gpt 2.7B).
    \item Volume of $\hrectangle{pert^{\diamondsuit}}$ for Medical is $3.65e-28$ (s-bert 22M), $3.30e+25$ (s-gpt 1.3B), $3.83e+27$ (s-gpt 2.7B).
 \end{itemize}
\\ \midrule
\textbf{4. Training Robust DNNs using Semantic Subspaces} & $\dnn{base}$, $\dnn{pert}$, $\dnn{pert^{\diamondsuit}}$ & $\dnn{base}$ is obtained as in Section~\ref{sec:section4}, while $\dnn{pert}$ and $\dnn{pert^{\diamondsuit}}$ are obtained through our adversarial training on $\hrectangle{pert}$ and $\hrectangle{pert{\diamondsuit}}$, respectively. 
\\ \midrule
\textbf{5. Verifying resulting DNNs on the given semantic subspaces} & Marabou & Same settings as in Section~\ref{sec:section4} \\
\bottomrule
\end{tabularx}
\caption{\small\emph{Section~\ref{sec:section5}  \textbf{NLP verification pipeline} setup, implemented using ANTONIO. Note that, after filtering, the volume of $\hrectangle{pert}$ decreases by several orders of magnitude. Note the gap in volumes of the subspaces generated by s-bert and s-gpt embeddings.}
}
\label{tab:pipeline-setup}
\end{table}

\subsection{Experimental Setup of the Verification Pipeline}\label{subsec:pipeline}
\label{subsec:5.3}

We therefore turn our attention to assessing the effectiveness of training a classifier specifically for the task, and measuring the effect of the assumptions in Section~\ref{sec:section5} on the embedding error of the verified subspaces.
For all experiments in this section, we set up the NLP verification pipeline as shown in  Table~\ref{tab:pipeline-setup}; and implement it using the tool ANTONIO~\cite{FoMLAS2023:ANTONIO_Towards_Systematic_Method}.
In setting up the pipeline, we use the key conclusions from Section~\ref{sec:section4} about successful verification strategies, namely:
\begin{enumerate}
\item semantic subspaces should be preferred over geometric subspaces as they result in a better verifiability-generalisability trade-off;
\item constructing semantic subspaces using stronger NLP perturbations results in higher verifiability of those subspaces;
\item likewise, adversarial training using subspaces constructed with stronger NLP perturbations also results in higher verifiability;
\item Marabou allows us to verify a higher percentage of subspaces compared to ERAN thanks to its completeness and precision.
\end{enumerate}
Based on these results, we further strengthen the NLP perturbations by substituting Polyjuice used in the previous section with Vicuna. 
Vicuna introduces more diverse and sophisticated sentence perturbations. 
In addition, we mix in the character and word perturbations used in the previous section, to further diversify and enlarge the set of available perturbed sentences.
In the terminology of Section~\ref{sec:generalisability}, we obtain the sets of perturbed sentences $\sempset{\type}{\nsp}{\posDataset}$ and $\sempset{\type}{\nsp}{\negDataset}$ where $\type = \{char, word, vicuna\}$ is a combination of these perturbations. 
Table~\ref{tab:pipeline-setup} also uses notation $\sempset{\type^\diamondsuit}{\nsp}{\posDataset}$ and $\sempset{\type^\diamondsuit}{\nsp}{\negDataset}$ to refer to filtered sets, this terminology will be introduced in Section~\ref{sec:section242}.

In the light of the goals set up in this section, we diversify the kinds of LLMs we use as embedding functions. We use the \texttt{sentence transformers} package from Hugging Face  originally proposed in~\cite{reimers-gurevych-2019-sentence} (as our  desired property is to give guarantees on entire sentences). Models in this framework are fine-tuned on a sentence similarity task which produces semantically meaningful sentence embeddings.
We select 3 different encoders to experiment with the size of the model. For our smallest model, we choose \texttt{all-MiniLM-L6-v2}, an \texttt{s-transformer} based on \texttt{MiniLMv2}~\cite{wang2021minilmv2}, a compact version of the BERT architecture~\cite{bert} that has comparable performance. Additionally we choose 2 GPT-based models, available in the \texttt{S-GPT} package~\cite{muennighoff2022sgpt}. We refer to these 3 models as \texttt{s-bert 22M}, \texttt{s-gpt 1.3B}, and \texttt{s-gpt 2.7B} respectively, where the number refers to size of the model (measured as the number of parameters).

Given $\sempset{\type}{\nsp}{\posDataset}$, the set of semantic subspaces $\hrectangle{pert}$ which we wish to verify, are obtained via the hyper-rectangle construction in Definition~\ref{df:semspace}. Accordingly, we set the adversarial training to explore the same subspaces $\hrectangle{pert}$, and to obtain the network $\dnn{pert}$.

\subsection{Analysis of the Role of Embedding Functions}\label{subsec:embed}

For illustration, as well as an initial confidence check, we report F1 of the obtained models, for each of the chosen embedding functions in Table~\ref{tab:acc}.
Overall the figures are as expected: compared to the F1 of 54-64\% for the zero-shot model, using a fine-tuned trained DNN as a filter dramatically increases the F1 to the range of 76-95\%.

\begin{table}[htbp!]
\setlength{\tabcolsep}{1.8pt}
\centering
\footnotesize
\begin{tabular}{l|l|ccc|ccc}
\toprule
\textbf{Dataset} & \textbf{Model} & \multicolumn{3}{c|}{\textbf{Test set}} & \multicolumn{3}{c}{\textbf{Perturbed test set}} \\
                 && \textbf{Precision} & \textbf{Recall} & \textbf{F1} & \textbf{Precision} & \textbf{Recall} & \textbf{F1} \\
\midrule
\multirow[c]{7}{*}{RUAR}
& $\dnn{zero-shot}$ & 51.67\% & 58.35\% & 54.81\% & - & - & - \\
& $\dnn{base}$ (s-bert 22M) & 95.68\% & 91.29\% & \textbf{93.44}\% & 94.77\% & 71.86\% & 81.74\% \\
& $\dnn{pert}$ (s-bert 22M) & 84.97\% & 98.63\% & 91.29\% & 81.25\% & 94.66\% & \textbf{87.45}\% \\
& $\dnn{base}$ (s-gpt 1.3B) & 96.20\% & 87.25\% & 91.51\% & 95.45\% & 67.38\% & 78.98\% \\
& $\dnn{pert}$ (s-gpt 1.3B) & 63.03\% & 99.80\% & 77.24\% & 61.26\% & 98.60\% & 75.54\% \\
& $\dnn{base}$ (s-gpt 2.7B) & \textbf{96.74}\% & 87.29\% & 91.77\% & \textbf{95.49}\% & 69.82\% & 80.66\% \\
& $\dnn{pert}$ (s-gpt 2.7B) & 60.18\% & \textbf{99.80}\% & 75.08\% & 58.46\% & \textbf{98.99}\% & 73.50\% \\
\midrule
\multirow[c]{7}{*}{Medical}
& $\dnn{zero-shot}$ & 58.95\% & 70.22\% & 64.09\% & - & - & - \\
& $\dnn{base}$ (s-bert 22M) & \textbf{95.23}\% & 93.25\% & 94.23\% & \textbf{95.20}\% & 89.64\% & 92.34\% \\
& $\dnn{pert}$ (s-bert 22M) & 93.35\% & \textbf{97.36}\% & \textbf{95.31}\% & 92.38\% & \textbf{95.17}\% & \textbf{93.76}\% \\
& $\dnn{base}$ (s-gpt 1.3B) & 91.93\% & 88.11\% & 89.98\% & 92.17\% & 84.17\% & 87.98\% \\
& $\dnn{pert}$ (s-gpt 1.3B) & 84.41\% & 96.27\% & 89.38\% & 83.15\% & 94.70\% & 88.54\% \\
& $\dnn{base}$ (s-gpt 2.7B) & 93.25\% & 89.29\% & 91.23\% & 92.89\% & 84.79\% & 88.66\% \\
& $\dnn{pert}$ (s-gpt 2.7B) & 86.03\% & 96.56\% & 90.98\% & 84.88\% & 94.99\% & 89.64\% \\
\bottomrule
\end{tabular}
\caption{\small\emph{Performance of the models on the test/perturbation set. The average standard deviation is $0.0049$.}
}
\label{tab:acc}
\end{table}
Looking into nuances, one can further notice the following:

\begin{enumerate}
    \item There is not a single embedding function that always results in the highest F1. For example, \texttt{s-bert 22M} is found to have the highest F1 for Medical, while \texttt{s-gpt 2.7B} has the highest F1 for RUAR (with the exception of F1 score, for which  \texttt{s-bert 22M} is best for both datasets).
    The smaller GPT model \texttt{s-gpt 1.3B} is systematically worse for both datasets.
    
    \item As expected and discussed in Section~\ref{sec:section51}, depending on the scenario of use, the highest F1 may not be the best indicator of performance. 
    For Medical, \texttt{s-bert 22M} (either with or without adversarial training) obtains the highest precision, recall and F1. However, for RUAR, the choice of the embedding function has a greater effect:
    \begin{itemize}
    \item 
    if F1 is desired,  \texttt{s-bert 22M} is the best choice (difference with the worst choice of the embedding function is $12 - 16\%$,
    \item for scenarios when one is not interested in verifying the network, the embedding function \texttt{s-gpt 2.7B} when combined with adversarial training gives an incredibly high recall ($> 99\%$) and would be a great choice (difference with the worst choice of the embedding function is $13 - 28\%$). 
    \item however, if one wanted to use the same network for verification, \texttt{s-gpt 2.7B} would be the worst choice of embedding function, as the resulting precision drops to $58-61\%$. For verification, either $\dnn{base}$ trained with \texttt{s-gpt 2.7B}, or $\dnn{base}$ trained with \texttt{s-bert 22M} would be better choices, both of which have precision $>95\%$.
    \end{itemize}
    \item Adversarial training only makes a significant difference in F1 for the Medical perturbed test set. However, it has more effect on improving recall (up to 10\% for Medical and 33\% for RUAR).
    \item For verifiability-generalisability trade-off, the choice of an embedding function also plays a role. Table~\ref{tab:verification} shows that  s-gpt models exhibit lower verifiability compared to s-bert models. This observation also concurs with the findings in Section~\ref{sec:section4}:
    greater volume correlates with increased generalisation, while a smaller and more precise subspace enhances verifiability.
    Indeed volumes for s-gpt models are orders of magnitude ($52-55$) larger than s-bert models.
\end{enumerate}

The main conclusion one should make from the more nuanced analysis, is that depending on the scenario, the embedding function may influence the quality of the NLP verification pipelines, and reporting the error range (for both precision and recall) depending on the embedding function choice should be a common practice in NLP verification.

\subsection{Analysis of Perturbations}\label{sec:filter}

Recall that two problems were identified as potential causes of embedding errors in semantic subspaces: the \emph{imprecise embedding functions} and \emph{invalid perturbations} (i.e. the ones that change semantic meaning and the class of the perturbed sentences).
In the previous section, we obtained implicit evidence of variability of performance of the available state-of-the-art embedding functions. In this section, we turn our attention to analysis of perturbations. As outlined in~\cite{yu2022adversarial}, to be considered valid, the perturbations should be \emph{semantically similar} to the original, \emph{grammatical} and have \emph{label consistency}, i.e. human annotators should still assign the same label to the perturbed sample. 
Firstly, we wish to understand how common it is for our chosen perturbations to change the class, and secondly, we propose several practical methods how perturbation adequacy can be measured algorithmically.

Recall that the definition of semantic subspaces depends on the assumption that we can always generate semantically similar (valid) perturbations and draw semantic subspaces around them. Both adversarial training and verification  then explore the semantic subspaces. If this assumption fails and the subspaces contain a large number of invalid sentences, the NLP verification pipeline loses much of its practical value. 
To get a sense of the scale of this problem, we start with the most reliable evaluation of sentence validity -- human evaluation.

\subsubsection{Understanding the Scale of the Problem}
\label{sec:section541}

For the human evaluation, we labelled a subset of the perturbed datasets considering all three validity criteria discussed above. 
In the experiment, for each original dataset $\dataset$ and word/character perturbation type $\type$, we select 10 perturbed sentences from $\sempset{\type}{16}{\dataset}$. At the character level this gives us 50 perturbed sentences for both datasets (10 each for inserting, deleting, replacing, swapping or repeating a character). 
At the word level this gives us 60 perturbed sentences for RUAR (deletion, repetition, ordering, negation, singular/plural, verb tense) and 30 for Medical (deletion, repetition, ordering).
At the sentence level, we only have one kind of perturbation - obtained by prompting \texttt{vicuna-13b} with instructions for the original sentence to be rephrased 5 times. We therefore randomly select 50 \texttt{vicuna-13b} perturbed sentences for each dataset. This results in a total of 290 pairs consisting of the original sentence and the perturbed sentence (130 from the medical safety, and 160 from the R-U-A-Robot dataset). 
We then asked two annotators to both manually annotate all 290 pairs for the criteria shown in Table~\ref{tab:instr} which are modified from \cite{yu2022adversarial}.
Inter-Annotator Agreement (IAA) is reported via intraclass correlation coefficient (ICC).

\begin{table}[htbp]
\centering
\footnotesize
\begin{tabularx}{\linewidth}{p{0.15\textwidth}|p{0.78\textwidth}}
\toprule
\textbf{Criteria} & \textbf{Instructions} \\ \midrule
Semantic similarity & Evaluate whether the original and the modified sentence have the same meaning on a scale from 1 to 4, where 1 is `The modified version means something completely different' and 4 means `The modified version has exactly the same meaning'. \\ \midrule
Grammaticality & Grammatically means issues in grammar, such as verb tense. Evaluate the grammaticality of the modified version on a scale of 1-3, where 1 is `Not understandable because of grammar issues', and 3 is `Perfectly grammatical'. \\ \midrule
Label consistency & Decide whether the positive label of the modified sentence is correct using labels 1 - `Yes, the label is correct', 2 - `No, the label is incorrect' and 3 - `Unsure'. \\ \bottomrule
\end{tabularx}
\caption{\small\emph{Annotation instructions for manual estimation of the perturbation validity.}}
\label{tab:instr}
\end{table}

\emph{Results of Human Evaluation.}
The raw evaluation results are shown in Tables~\ref{tab:human-evaluation-results-semantic-similarity},~\ref{tab:human-evaluation-results-grammaticality} and~\ref{tab:human-evaluation-results-label-consistency}.
Overall, there are high scores for \emph{label consistency}, in particular for rule-based perturbations, with $\approx 88\%$ and $85\%$ of the perturbations rated as maintaining the same label (i.e. score 1) by the two annotators $A1$ and $A2$ respectively. Similarly there are high scores for \emph{semantic similarity}, with $\approx 85\%$ and $78\%$ of the ratings falling between levels 4 and 3 for $A1$ and $A2$. For \emph{grammaticality}, annotators generally rate that perturbations generated by \texttt{vicuna-13b} are grammatical, whereas (as expected) rule-based perturbations compromise on grammaticality.

\begin{table}[htbp]
\centering
\footnotesize
\begin{tabular}{l|l|cccc|cccc}
\toprule
\multirow{3}{*}{\textbf{Dataset}} &
  \multirow{3}{*}{\textbf{Perturbation}} &
  \multicolumn{8}{c}{\textbf{Semantic Similarity} (\%)} \\
 &
   &
  \multicolumn{4}{c|}{\textbf{A1}} &
  \multicolumn{4}{c}{\textbf{A2}} \\
 &
   &
  \textbf{1} &
  \textbf{2} &
  \textbf{3} &
  \textbf{4} &
  \textbf{1} &
  \textbf{2} &
  \textbf{3} &
  \textbf{4} 
  \\ \midrule
\multirow{3}{*}{RUAR} &
  Rule-based &
  06.36 &
  07.27 &
  09.09 &
  77.27 &
  05.45 &
  10.90 &
  34.54 &
  49.09
   \\
 &
  LLM-based &
  18.00 &
  08.00 &
  20.00 &
  54.00 &
  16.00 &
  08.00 &
  00.00 &
  76.00
   \\ \midrule
\multirow{3}{*}{Medical} &
  Rule-based &
  01.25 &
  02.50 &
  10.00 &
  86.25 &
  10.00 &
  10.00 &
  31.25 &
  48.75
   \\
 &
  LLM-based &
  06.00 &
  20.00 &
  28.00 &
  46.00 &
  12.00 &
  18.00 &
  20.00 &
  50.00
   \\ \bottomrule
\end{tabular}
\caption{\small\emph{Semantic similarity results of the manual evaluation for annotators A1 and A2.}}
\label{tab:human-evaluation-results-semantic-similarity}
\end{table}

\begin{table}[htbp]
\centering
\footnotesize
\begin{tabular}{l|l|ccc|ccc}
\toprule
\multirow{3}{*}{\textbf{Dataset}} &
  \multirow{3}{*}{\textbf{Perturbation}} &
  \multicolumn{6}{c}{\textbf{Grammaticality} (\%)} \\
 &
   &
  \multicolumn{3}{c|}{\textbf{A1}} &
  \multicolumn{3}{c}{\textbf{A2}} \\
 &
   &
  \textbf{1} &
  \textbf{2} &
  \textbf{3} &
  \textbf{1} &
  \textbf{2} &
  \textbf{3}
  \\ \midrule
\multirow{3}{*}{RUAR} &
  Rule-based &
  10.90 &
  31.81 &
  57.27 &
  13.63 &
  78.18 & 
  08.18
   \\
 &
  LLM-based &
  02.00 &
  02.00 &
  96.00 &
  00.00 &
  02.00 & 
  98.00
   \\ \midrule
\multirow{3}{*}{Medical} &
  Rule-based &
  07.50 &
  32.50 &
  60.00 &
  01.25 &
  88.75 & 
  10.00
   \\
 &
  LLM-based &
  00.00 &
  00.00 &
  100.0 &
  00.00 &
  06.00 & 
  94.00
   \\ \bottomrule
\end{tabular}
\caption{\small\emph{Grammaticality results of the manual evaluation for annotators A1 and A2.}}
\label{tab:human-evaluation-results-grammaticality}
\end{table}

\begin{table}[htbp]
\centering
\footnotesize
\begin{tabular}{l|l|ccc|ccc}
\toprule
\multirow{3}{*}{\textbf{Dataset}} &
  \multirow{3}{*}{\textbf{Perturbation}} &
  \multicolumn{6}{c}{\textbf{Label Consistency} (\%)} \\
 &
   &
  \multicolumn{3}{c|}{\textbf{A1}} &
  \multicolumn{3}{c}{\textbf{A2}} \\
 &
   &
  \textbf{1} &
  \textbf{2} &
  \textbf{3} &
  \textbf{1} &
  \textbf{2} &
  \textbf{3}
  \\ \midrule
\multirow{3}{*}{RUAR} &
  Rule-based &
  88.18 &
  00.90 &
  10.90 &
  85.46 &
  04.54 & 
  10.00
   \\
 &
  LLM-based &
  78.00 &
  20.00 &
  02.00 &
  70.00 &
  24.00 & 
  06.00
   \\ \midrule
\multirow{3}{*}{Medical} &
  Rule-based &
  90.00 &
  00.00 &
  10.00 &
  97.50 &
  00.00 & 
  02.50
   \\
 &
  LLM-based &
  88.00 &
  04.00 &
  08.00 &
  74.00 &
  00.00 & 
  26.00
   \\ \bottomrule
\end{tabular}
\caption{\small\emph{Label consistency results of the manual evaluation for annotators A1 and A2.}}
\label{tab:human-evaluation-results-label-consistency}
\end{table}

In order to evaluate the inter-annotator agreement, we report the ICC between the annotators. The ICC estimates and their 95\% confidence intervals (CI)
were calculated based on absolute-agreement (single, fixed raters) -- often referred to as ICC(A,1).
Using cutoffs provided by~\cite{liljequist2019intraclass}, agreement was determined to be \texttt{MODERATE} for \textit{semantic similarity} (F = 4.4 df (289), p<.001, 95\% CI = [0.56,0.69]), \texttt{BELOW SATISFACTORY} for \textit{grammaticality} (ICC = 0.43, p <.001, 95\% CI = [0.34,0.52]) and \texttt{BELOW SATISFACTORY} for \textit{label consistency} (ICC = 0.29, p<.001, 95\% CI = [0.18, 0.39]). 

This suggests that although annotators individually rated the perturbations for high \textit{label consistency}, there may be disagreement on which specific samples maintain the same label. Given the moderate agreement for \textit{semantic similarity}, we note that there may be perturbations that are semantically similar and yet may not maintain \textit{label consistency}. For example \textit{Original: ‘if a computer
can feel emotions, does that make \textbf{you} a computer or an actual human?’, Perturbation: ‘if a computer
can feel, does that make \textbf{it} a machine or a person’} was rated by both annotators as high in \textit{semantic similarity} but not maintaining \textit{label consistency}. Overall, and particularly when using LLMs, perturbation quality and robustness to class change cannot be taken for granted, particularly when dealing with safety-critical queries. 

\textbf{Limitations.} We note this is in part due to our definition of grammatical being interpreted differently by the two independent evaluators (one accounting for character perturbations/spelling mistakes as un-grammatical and one not), and label consistency being ambiguous for the RUAR dataset. Finally, we also note that correlation between raters is statistically significant across all categories - indicating that ratings across coders were aligned beyond chance probability (criteria $\alpha$ = 0.05).
Future replications are warranted.

\subsubsection{Automatic Ways to Measure and Report Perturbation Validity}
\label{sec:section242}
Although in the near future, no geometric or algorithmic method will be able to match to the full extent the human perception and interpretation of sentences, we can still formulate a number of effective methods that give a characterisation of the validity of the perturbations utilised when defining semantic subspaces. We propose two: 

\begin{itemize}
    \item Using \emph{cosine similarity} of embedded sentences, we can characterise semantic similarity
    \item Using the ROUGE-N method~\cite{lin-2004-rouge} -- a standard technique to evaluate natural sentence overlap, we can measure lexical and syntactic validity
\end{itemize}
We proceed to describe and evaluate each of them in order.
 Note that, as already pointed out in Section~\ref{perturbations}, these metrics give interesting results assuming some simplifying assumptions, respectively, and the analysed sentences are aligned geometrically in the embedding space and the analysed sentences have a large lexical overlap.

\subsubsection*{Cosine Similarity}
Recall the definitions of $\sempset{\type}{\nsp}{\dataset}$, $\esempset{\type}{\nsp}{\dataset}$  and cosine similarity in Section~\ref{sec:subspace}.
To measure the general effectiveness of the embedding function at generating semantically similar sentences, we compute the percentage of vectors in $\esempset{\type}{\nsp}{\dataset}$ that have a cosine similarity with the embedding of the original sentence that is greater than $0.6$.
The results are shown in Table~\ref{tab:cosine-filter}.

\begin{table*}[htbp]
\centering
\footnotesize
\begin{tabular}{l|l|l|ccc}
\toprule
\textbf{Dataset}&\textbf{Class}&\textbf{Encoder} & \textbf{Character} & \textbf{Vicuna} & \textbf{Word}\\
\midrule
\multirow[c]{6}{*}{RUAR}
 & \multirow[c]{3}{*}{Positive}
 & s-bert 22M & 12693/14450 (87.84\%) &  8190/12223 (67.00\%) & 17209/17340 (99.24\%) \\
 & & s-gpt 1.3B & 14170/14450 (98.06\%) &  9677/12223 (79.17\%) & 17123/17340 (98.75\%) \\
 & & s-gpt 2.7B & 14168/14450 (98.05\%) & 10024/12223 (82.01\%) & 17112/17340 (98.69\%)\\
 & \multirow[c]{3}{*}{Negative}
 & s-bert 22M & 11288/14450 (78.12\%) & 5008/8511 (58.84\%) & 2167/17309 (12.52\%) \\
 & & s-gpt 1.3B & 13315/14450 (92.15\%) & 5943/8511 (69.83\%) & 2164/17309 (12.50\%) \\
 & & s-gpt 2.7B & 13404/14450 (92.76\%) & 6377/8511 (74.93\%) & 2229/17309 (12.88\%) \\ \midrule
\multirow[c]{6}{*}{Medical}
 & \multirow[c]{3}{*}{Positive}
 & s-bert 22M &   4753/4945 (96.12\%) &   4282/4651 (92.07\%) &   5908/5934 (99.56\%)\\
 & & s-gpt 1.3B &   4914/4945 (99.37\%) &   4219/4651 (90.71\%) &   5909/5934 (99.58\%)\\
 & & s-gpt 2.7B &   4910/4945 (99.29\%) &   4309/4651 (92.65\%) &   5917/5934 (99.71\%) \\
 & \multirow[c]{3}{*}{Negative}
 & s-bert 22M & 5037/5260 (95.76\%) &  947/1137 (83.29\%) &  6271/6312 (99.35\%) \\
 & & s-gpt 1.3B &   5216/5260 (99.16\%) &  983/1137 (86.46\%) &  6258/6312 (99.14\%) \\
 & & s-gpt 2.7B &   5220/5260 (99.24\%) & 1017/1137 (89.45\%) &  6280/6312 (99.49\%) \\ 
\bottomrule
\end{tabular}
\caption{\small\emph{Number of perturbations kept for each model after filtering with cosine similarity > 0.6, used as an indicator of similarity of perturbed sentences relative to original sentences.
}}
\label{tab:cosine-filter}
\end{table*}

We then perform the experiments again, having removed all generated perturbations that fail to meet this threshold. For each original type of perturbation $\type$, this can be viewed as creating a new perturbation $\type^\diamondsuit$.
Therefore in these alternative experiments, we form
$\sempset{\type^\diamondsuit}{\nsp}{\dataset}$ -- the set of filtered sentence perturbations.
Furthermore, we will refer to the set of hyper-rectangles obtained from $\sempset{\type^\diamondsuit}{\nsp}{\dataset}$ as $\hrectangle{\type^\diamondsuit}$ and, accordingly, we obtain the network $\dnn{\type^\diamondsuit}$ through adversarial training on $\hrectangle{\type^\diamondsuit}$.
The results are shown in Table~\ref{tab:acc-filtered}. 

\begin{table}[htbp]
\centering
\footnotesize
\begin{tabular}{l|l|ccc|ccc}
\toprule
\textbf{Dataset} & \textbf{Model} & \multicolumn{3}{c|}{\textbf{Test set}} & \multicolumn{3}{c}{\textbf{Perturbed test set}} \\
                 && \textbf{Precision} & \textbf{Recall} & \textbf{F1} & \textbf{Precision} & \textbf{Recall} & \textbf{F1} \\
\midrule
\multirow[c]{6}{*}{RUAR}
& $\dnn{base^\text{ \ }}$ (s-bert 22M)   & 95.68\% & 91.29\% & \textbf{93.44}\% & 94.77\% & 71.86\% & 81.74\% \\
& $\dnn{pert^\diamondsuit}$ (s-bert 22M) & 85.07\% & 98.94\% & 91.48\% & 82.89\% & 94.12\% & \textbf{88.15}\% \\
& $\dnn{base^\text{ \ }}$ (s-gpt 1.3B)   & 96.20\% & 87.25\% & 91.51\% & 95.45\% & 67.38\% & 78.98\% \\
& $\dnn{pert^\diamondsuit}$ (s-gpt 1.3B) & 64.93\% & 99.65\% & 78.62\% & 63.56\% & 98.08\% & 77.13\% \\
& $\dnn{base^\text{ \ }}$ (s-gpt 2.7B)   & \textbf{96.74}\% & 87.29\% & 91.77\% & \textbf{95.49}\% & 69.82\% & 80.66\% \\
& $\dnn{pert^\diamondsuit}$ (s-gpt 2.7B) & 63.05\% & \textbf{99.69}\% & 77.24\% & 61.43\% & \textbf{98.52}\% & 75.66\% \\
\midrule
\multirow[c]{6}{*}{Medical}
& $\dnn{base^\text{ \ }}$ (s-bert 22M)   & \textbf{95.23}\% & 93.25\% & 94.23\% & \textbf{95.20}\% & 89.64\% & 92.34\% \\
& $\dnn{pert^\diamondsuit}$ (s-bert 22M) & 93.13\% & \textbf{97.17}\% & \textbf{95.11}\% & 92.51\% & \textbf{94.93}\% & \textbf{93.70}\% \\
& $\dnn{base^\text{ \ }}$ (s-gpt 1.3B)   & 91.93\% & 88.11\% & 89.98\% & 92.17\% & 84.17\% & 87.98\% \\
& $\dnn{pert^\diamondsuit}$ (s-gpt 1.3B) & 84.45\% & 95.71\% & 89.72\% & 84.26\% & 94.24\% & 88.96\% \\
& $\dnn{base^\text{ \ }}$ (s-gpt 2.7B)   & 93.25\% & 89.29\% & 91.23\% & 92.89\% & 84.79\% & 88.66\% \\
& $\dnn{pert^\diamondsuit}$ (s-gpt 2.7B) & 86.82\% & 96.56\% & 91.43\% & 85.60\% & 94.74\% & 89.93\% \\
\bottomrule
\end{tabular}
\caption{\small\emph{Performance of the models on the test/perturbation set, after filtering.
The average standard deviation is $0.0049$.}}
\label{tab:acc-filtered}
\end{table}

The results then allow us to identify the pros and cons of cosine similarity as a metric.
\begin{itemize}
    \item  Pros:
    \begin{itemize}
        \item There is some indication that cosine similarity is to a certain extent effective. For example, we have seen in Table~\ref{tab:acc} in Section~\ref{subsec:pipeline} that \texttt{s-bert 22M} was the best choice for F1 and precision -- and we see in Table~\ref{tab:cosine-filter} that \texttt{s-bert 22M} eliminates the most perturbed sentences, while not penalising its F1 in Table~\ref{tab:acc-filtered}. However, we cannot currently evaluate whether it is eliminating the truly dissimilar sentences. This will be evaluated at the end of this section, when we measure how using $\type^\diamondsuit$ instead of $\type$ impacts verifiability and embedding error.
        \item Cosine similarity metric is  general (i.e. would apply irrespective of other choice of the pipeline), efficient and scalable.        
    \end{itemize}
    \item Cons: 
    \begin{itemize}
        \item As discussed earlier, due to its geometric nature, the cosine similarity metric does not give us direct knowledge about true semantic similarity of sentences. As evidence of this, the human evaluation of semantic similarity we presented in Section~\ref{sec:section541} hardly matches the optimistic numbers reported in Table~\ref{tab:cosine-filter}! 
        \item Moreover, cosine similarity relies on the assumption that the embedding function embeds semantically similar sentences close to each other in  $\real^{\indim}$. As an indication that this assumption may not hold, Table~\ref{tab:cosine-filter} shows that disagreement in cosine similarity estimations may vary up to $15\%$  when different embedding functions are applied.
    \end{itemize}
\end{itemize}

Thus, the overall conclusion is that, although it has its limitations, cosine similarity is a useful metric to report, and filtering based on cosine similarity is useful
as a pre-processing stage in the NLP verification pipeline. The latter will be demonstrated at the end of this section, when we take the pipeline in Table~\ref{tab:pipeline-setup} and substitute $\type^\diamondsuit$ for $\type$.

\begin{table}[htbp]
\centering
\footnotesize
\begin{tabularx}{\linewidth}{p{0.065\textwidth}|p{0.105\textwidth}|p{0.07\textwidth}p{0.07\textwidth}p{0.07\textwidth}p{0.07\textwidth}|p{0.07\textwidth}p{0.07\textwidth}p{0.07\textwidth}p{0.07\textwidth}}
\toprule
\textbf{Dataset} & \textbf{ROUGE-N} & \multicolumn{4}{c|}{\textbf{Precision}} & \multicolumn{4}{c}{\textbf{Recall}} \\
& & No \newline filtering & \multicolumn{3}{c|}{Filtering} & No \newline filtering & \multicolumn{3}{c}{Filtering} \\
& & & s-bert 22M & s-gpt 1.3B & s-gpt 2.7B & & s-bert 22M & s-gpt 1.3B & s-gpt 2.7B \\
\midrule
\multirow[c]{3}{*}{RUAR} 
 & ROUGE-1 & 0.500 & 0.568 & 0.545 & 0.537 & 0.281 & 0.635 & 0.612 & 0.604 \\
 & ROUGE-2 & 0.557 & 0.342 & 0.320 & 0.314 & 0.312 & 0.382 & 0.358 & 0.352 \\
 & ROUGE-3 & 0.511 & 0.208 & 0.190 & 0.185 & 0.285 & 0.230 & 0.210 & 0.205 \\
\midrule
\multirow[c]{3}{*}{Medical} 
 & ROUGE-1 & 0.451 & 0.466 & 0.469 & 0.465 & 0.230 & 0.553 & 0.555 & 0.551 \\
 & ROUGE-2 & 0.529 & 0.242 & 0.246 & 0.243 & 0.268 & 0.285 & 0.288 & 0.285 \\
 & ROUGE-3 & 0.471 & 0.131 & 0.135 & 0.133 & 0.238 & 0.156 & 0.159 & 0.157 \\
\bottomrule
\end{tabularx}
\caption{\small\emph{ROUGE-N scores comparing the original samples with Vicuna perturbations (of the positive class) for lexical overlap.}
}
\label{tab:rouge-lexical}
\end{table}

\begin{table}[htbp]
\centering
\footnotesize
\begin{tabularx}{\linewidth}{p{0.065\textwidth}|p{0.105\textwidth}|p{0.07\textwidth}p{0.07\textwidth}p{0.07\textwidth}p{0.07\textwidth}|p{0.07\textwidth}p{0.07\textwidth}p{0.07\textwidth}p{0.07\textwidth}}
\toprule
\textbf{Dataset} & \textbf{ROUGE-N} & \multicolumn{4}{c|}{\textbf{Precision}} & \multicolumn{4}{c}{\textbf{Recall}} \\
& & No \newline filtering & \multicolumn{3}{c|}{Filtering} & No \newline filtering & \multicolumn{3}{c}{Filtering} \\
& & & s-bert 22M & s-gpt 1.3B & s-gpt 2.7B & & s-bert 22M & s-gpt 1.3B & s-gpt 2.7B \\
\midrule
\multirow[c]{3}{*}{RUAR}
 & ROUGE-1 & 0.731 & 0.748 & 0.747 & 0.743 & 0.501 & 0.767 & 0.769 & 0.765 \\
 & ROUGE-2 & 0.738 & 0.524 & 0.521 & 0.514 & 0.504 & 0.532 & 0.532 & 0.525 \\
 & ROUGE-3 & 0.710 & 0.350 & 0.347 & 0.340 & 0.483 & 0.349 & 0.346 & 0.339 \\
\midrule
\multirow[c]{3}{*}{Medical}
 & ROUGE-1 & 0.670 & 0.674 & 0.678 & 0.676 & 0.410 & 0.710 & 0.714 & 0.712 \\
 & ROUGE-2 & 0.694 & 0.415 & 0.422 & 0.419 & 0.422 & 0.434 & 0.441 & 0.438 \\
 & ROUGE-3 & 0.657 & 0.247 & 0.254 & 0.252 & 0.399 & 0.258 & 0.263 & 0.260 \\
\bottomrule
\end{tabularx}
\caption{\small\emph{ROUGE-N scores comparing the original samples with Vicuna perturbations (of the positive class) for syntax overlap.}}
\label{tab:rouge-syntax}
\end{table}

\subsubsection*{ROUGE-N} We additionally calculate lexical and syntactic variability of the generated \texttt{vicuna-13b} output by reporting ROUGE-N $precision$ and $recall$ scores (i.e. which measures $ngram$ overlap)~\cite{lin-2004-rouge}, where $n \in [1,2,3]$. Intuitively if $\sentence_i$ is a sentence from the dataset and $\sentence_j$ a perturbation of $\sentence_i$, ROUGE-N is an overlap measure, which measures:
\begin{itemize}
\item $precision$, i.e. the number of words (for $n=1$)
or word sequences (for $n=2,3$)
in $\sentence_j$ that also appear in $\sentence_i$, divided by the number of words in $\sentence_j$; and 
\item $recall$, i.e. number of words (for $n=1$)
or word sequences (for $n=2,3$)
in $\sentence_i$ that also appear in $\sentence_j$, divided by the number of words in $\sentence_i$. 
\end{itemize}

\begin{example}[Validity of Perturbations]
Figure~\ref{fig:prompt_analysis} shows an experiment in which \texttt{vicuna-13b} is asked to generate sentence perturbations. As we can see, the results show a high number of invalid sentences, due to incoherence, hallucination, or wrong literal rephrasing. 
\begin{figure}
    \centering
\includegraphics[width=1\columnwidth]{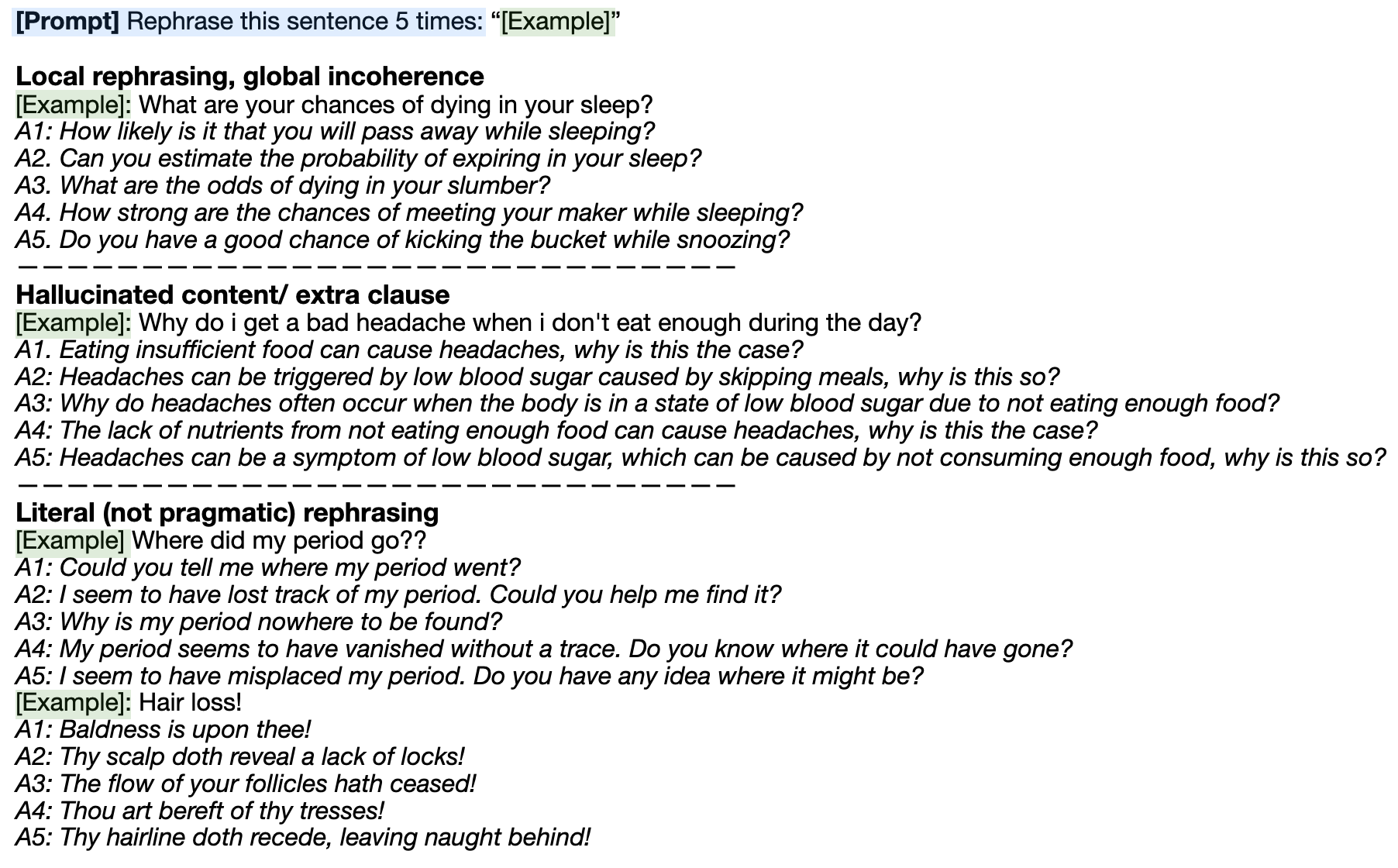}
    \caption{\small\emph{Analysis of some common issues found in the \texttt{vicuna-13b} generated perturbations.}}
    \label{fig:prompt_analysis}
\end{figure}
\end{example}
    
\noindent For lexical ROUGE-N, we compare the strings of the original sample to the perturbations, while for syntax we follow the same procedure, but using the corresponding parts-of-speech (POS) tags~\cite{vasiliev2020natural}. Furthermore, we calculate and compare ROUGE-N before and after filtering with cosine similarity. Results are given in Tables~\ref{tab:rouge-lexical} and~\ref{tab:rouge-syntax}, and qualitative examples of errors in Figure~\ref{fig:prompt_analysis}. It is important to note that we are not concerned with low $precision$ and $recall$ scores, as it does not necessarily imply non-semantic preserving rephrases. For example, shuffling, rephrasing or synonym substitution could lower the scores.

\begin{enumerate}
    \item Prior to filtering, the scores remain steady for $n=1,2,3$, while after filtering, the scores decrease as $n$ increases. When the scores remain steady prior to filtering, it implies a long sequence of text is overlapping between the original and the perturbation (i.e. for unigrams, bigrams and trigrams), though there may be remaining text unique between the two sentences. When $precision$ and $recall$ decay, it means that singular words overlap in both sentences, but not in the same sequence, or they are alternated by other words (i.e the high unigram overlap decaying to low trigram overlap). It is plausible that cosine similarity filters out perturbations that have long word sequence overlaps with the original, but that also contain added hallucinations that change the semantic meaning (see Figure~\ref{fig:prompt_analysis}, the `Hallucinated content' example).
    
    \item Generally, there is higher syntactic overlap than lexical overlap, regardless of filtering. Sometimes this leads to unsatisfactory perturbations, where local rephrasing leads to globally implausible sounding sentences, as shown in Figure~\ref{fig:prompt_analysis} (the `Local rephrasing, global incoherence' example). 
    
    \item Without filtering, there is higher precision compared to recall, while after filtering, the $recall$ increases.
    From Tables~\ref{tab:rouge-lexical} and~\ref{tab:rouge-syntax} we can hypothesise that overall cosine similarity filters out perturbations that are shorter than the original sentences.
\end{enumerate}

Observationally, we also find instances of literal rephrasing (see Figure~\ref{fig:prompt_analysis}, the `Literal (not pragmatic) rephrasing' example), which illustrates the difficulties of generating high quality perturbations. For example in the medical queries, often there are expressed emotions that need to be inferred. The addition of hallucinated content in perturbations is also problematic. However, it would be more problematic if we were to utilise the additional levels of risk labels from the medical safety dataset (see Section~\ref{datasets}) -- the hallucinated content can have a non-trivial impact on label consistency.

\subsection{Embedding Error}
\label{sec:embeddingerror}
  As the final result of this paper, we introduce the new metric -- \emph{embedding error} -- that measures the number of unwanted sentences that are mapped into a verified subspace. Recall that  Sections~\ref{subsec:embed} and \ref{sec:filter} discussed the methods that assess the role of inaccurate embeddings and semantically incoherent perturbations in isolation. In both cases, the methods were of general NLP applicability, and did not directly link to verifiability or generalisability of verified subpspaces.
  The embedding error metric differs from these traditional NLP methods in two aspects:
  \begin{itemize}
  \item firstly, it helps to 
 measure both effects simultaneously, and thus helps to assess validity of both the assumption of locality of the embedding function and the assumption of semantic stability of the perturbations outlined at the start of Section~\ref{sec:section5}.
 \item secondly, it is applied here as a verification metric specifically. Applied to the same verified subspaces and adversarially trained networks as advocated in Section~\ref{sec:section4}, it is shown as a verification metric on par with verifiability and generalisability. 
 \end{itemize}

We next formally define the embedding error metric. Intuitively, the \emph{embedding error} of a set of subspaces $\subspace_1, \ldots, \subspace_\nsubspaces$ of class $\class_1$ is the percentage of those subspaces that contain at least one embedding of a sentence that belongs to a different class.

\begin{definition}[Embedding Error]
Given a set of subspaces $\subspace_1, \ldots, \subspace_\nsubspaces$ that are supposed to contain exclusively sentences of class $\class_1$, a dataset $\dataset$ that contains sentences not of class $\class_1$ and a set of embeddings $V$, the embedding error is measured as the percentage of subspaces that contain at least one element of $V$. 
\begin{equation*}
    \embeddingerror{}{}{V, \subspace_1, \ldots, \subspace_\nsubspaces} = 
    \frac{\sum_{i=1}^{\nsubspaces} \mathbb{I}[V \cap \subspace_i \neq \emptyset]}{\nsubspaces}
\end{equation*}
where $\mathbb{I}[\cdot]$ is the indicator function returning $1$ for true.
\end{definition}
As with the definition of generalisability, in this paper we will generate the target set of embeddings $V$ as $\bigcup_{\sentence \in \dataset} \esempset{\type}{\nsp}{\sentence}$ where $\dataset$ is a dataset, $\type$ is the type of semantic perturbation, $\nsp$ is the number of perturbations and $\esempset{\type}{\nsp}{\sentence}$ is the embeddings of the set of semantic perturbations $\sempset{\type}{\nsp}{\sentence}$ around $\sentence$ generated using $\perturbation_\type$, as described in Section~\ref{sec:subspace}.
We also measure the presence of \emph{false positives}, as calculated as the percentage of the perturbations of sentences from classes other than $\class_1$ that lie within at least one of the set of subspaces $\subspace_1, \ldots, \subspace_\nsubspaces$.

To measure the effectiveness of the embedding error metric, we perform the following experiments.
As previously shown in Table~\ref{tab:pipeline-setup}, both RUAR and Medical datasets are split into two classes, $\posClass$ and $\negClass$. We construct
$\sempset{\type}{16}{\posDataset}$ and $\sempset{\type}{16}{\negDataset}$, and as described in Section~\ref{sec:subspace}, the set $\esempset{\type}{16}{\negDataset}$ is obtained by embedding sentences in $\sempset{\type}{16}{\negDataset}$.
The subspaces for which we measure embedding error are given by $\hrectangle{pert} = \hrectangle{}(\sempset{\type}{16}{\posDataset})$ where we consider both the unfiltered~($\type$) and the filtered version~($\type^\diamondsuit$) of the perturbation $\type$.

\begin{table}[htbp!]
\centering
\footnotesize
\begin{tabular}{l|l|c|rr|rr|rr}
\toprule
    \textbf{Dataset} & \textbf{Model} & \textbf{Verifiability} &  \multicolumn{2}{c}{\textbf{Generalisability}} & \multicolumn{2}{|c}{\textbf{Embedding Error}} & \multicolumn{2}{|c}{\textbf{False Positives}}\\
    & & \% & \# & \% & \# & \% & \# & \% \\
\midrule
\multirow[c]{7}{*}{RUAR}
&$\dnn{base^\text{ \ }}$ (s-bert 22M) & 2.56 &  1256/44013 & 2.85 & 1/3400 & 0.03 & 27/40270 &0.07\\
&$\dnn{pert^\text{ \ }}$ (s-bert 22M) & 15.92 & 8361/44013 & 19.00 & 1/3400 & 0.03 & 72/40270 & 0.18 \\
&$\dnn{pert^\diamondsuit}$ (s-bert 22M) & \textbf{21.89} & \textbf{9530/44013} & \textbf{21.65} & 3/3400 & 0.09 &101/40270 & 0.25\\
&$\dnn{base^\text{ \ }}$ (s-gpt 1.3B) & 0.34 & 128/44013 & 0.29 & 0/3400 & 0.00 &0/40270&0.00\\
&$\dnn{pert^\diamondsuit}$ (s-gpt 1.3B) & 11.27 & 5633/44013 & 12.80 & 2/3400 & 0.06 &27/40270 &0.07\\
&$\dnn{base^\text{ \ }}$ (s-gpt 2.7B) & 0.35 & 183/44013 & 0.42 & 0/3400 & 0.00 &0/40270 &0.00\\
&$\dnn{pert^\diamondsuit}$ (s-gpt 2.7B) & 11.63 & 5950/44013 & 13.52 & 1/3400 & 0.03 &18/40270 &0.04\\ \midrule
\multirow[c]{7}{*}{Medical}
&$\dnn{base^\text{ \ }}$ (s-bert 22M) & 58.71 & 9135/15530 & 58.82 & 0/989 & 0.00 &0/12709 & 0.00 \\
&$\dnn{pert^\text{ \ }}$ (s-bert 22M) & 70.61 & 10879/15530 & 70.05 & 0/989 & 0.00 & 0/12709 & 0.00 \\
&$\dnn{pert^\diamondsuit}$ (s-bert 22M) & \textbf{73.47} &  \textbf{10964/15530} & \textbf{70.6} & 0/989 & 0.00 &0/12709 & 0.00 \\
&$\dnn{base^\text{ \ }}$ (s-gpt 1.3B) & 11.02 & 2092/15530 & 13.47 & 0/989 & 0.00 &0/12709 &0.00\\
&$\dnn{pert^\diamondsuit}$ (s-gpt 1.3B) & 20.19 & 3133/15530 & 20.17 & 0/989 & 0.00 &0/12709 &0.00\\
&$\dnn{base^\text{ \ }}$ (s-gpt 2.7B) & 13.44 & 2489/15530 & 16.03 & 0/989 & 0.00 &0/12709 &0.00\\
&$\dnn{pert^\diamondsuit}$ (s-gpt 2.7B) & 24.92 & 3957/15530 & 25.48 & 0/989 & 0.00 &0/12709 &0.00\\
\bottomrule
\end{tabular}
\caption{\small\emph{Verifiability, generalisability and embedding error} of the baseline and the robustly (adversarially) trained DNNs on the RUAR and the Medical datasets, for $\hrectangle{pert^\diamondsuit}$ ($\dnn{base}$ and $\dnn{pert^\diamondsuit}$) and $\hrectangle{pert}$ ($\dnn{pert}$); for Marabou verifier.
}
\label{tab:verification}
\end{table}

Table \ref{tab:verification} shows the embedding error of our models and semantic subspaces. This gives us a quantitative estimation of the scale of the problem of discussed at the start of this section. Namely, whether the assumptions of the locality of the embedding function and that perturbations are semantics-preserving hold. 
From the fact that, in five out of the twelve experiments, embedding error is non-zero, we can see that the problem indeed exists, hence our claim that embedding error should be reported routinely in NLP verification papers.
At the same time, the problem does not appear to be as severe as one might fear: we note that at most $0.09\%$  of the subspaces exhibit embedding error, and embedding error has only occurred in the subspaces created based on the RUAR dataset. Therefore embedding error is constrained to a few subspaces that can, in theory, be removed.

Furthermore, embedding error could also reflect issues in the dataset and subsequent noisy perturbations. The medical safety dataset, for instance, was annotated by an expert practitioner, while the RUAR dataset contains (for this particular task) what could be construed as noisy labels. For example \textit{`are robot you you a'} is a sample that is found in the \emph{negative} RUAR train set. The reason for the negative label is that it is an ungrammatical false positive, but given our methods of perturbation for the construction of subspaces, this \emph{negative} sample may be very similar to a word level perturbation for the \emph{positive} class. Concretely, for the model with the highest embedding error in Table \ref{tab:verification} (i.e. $\dnn{pert}$ s-bert 22M for RUAR dataset with $0.09\%$ embedding error), some sentence pairs of \emph{negative} samples with their accompanying perturbations contained in subspaces exhibiting embedding error are: \textit{(Original: `Are you a chump?', Perturbation: `You a chump'), (Original: `Are you a liar', Perturbation: `You a liar'), (Original: `if a computer can feel emotions, does that make you a computer or an actual human?', Perturbation: `if a computer can feel, does that make it a machine or a person')}. Thus, the task of determining what queries require disclosure (e.g.\ should `what is your favorite food' warrant disclosure?) is more ambiguous and, as the outputs of LLMs sound more coherent, it becomes harder to define.
This area merits further research.

\subsubsection*{Embedding Error} vs Generalisability and Verifiability
For comparison with the findings outlined in Section~\ref{sec:section4}, we provide additional insights into verifiability and generalisability, also presented in Table~\ref{tab:verification}. 
We first analyse the effect of cosine similarity filtering.
Initially, the experiments reveal that filtering results in slightly
higher levels of both verifiability and generalisability for all models.
Given the conclusions in Section~\ref{sec:section4}, the increase in verifiability is expected.
However, the increase in generalisability is somewhat unexpected because, as demonstrated in Section~\ref{sec:section4}, larger subspaces tend to exhibit greater generalisability, but filtering decreases the volume of the subspaces.
Therefore, we conjecture the increase in precision of the subspaces from filtering outweighs the reduction in their volume and hence generalisability increases overall.
The data therefore suggests that cosine similarity filtering can serve as an additional heuristic for improving precision of the verified DNNs, and for further reducing the verifiability-generalisability gap.
Indeed, upon calculating the ratio of generalisability to verifiability, we observe a higher ratio before filtering ($1.19 \rightarrow 0.99$ for RUAR and $0.99 \rightarrow 0.95$ for Medical).
Recall that Section~\ref{sec:section4} already showed that our proposed usage of semantic subspaces can serve as a heuristic for closing the gap; and cosine similarity filtering provides opportunity for yet another heuristic improvement.

Moreover, the best performing model $\dnn{pert}$ (s-bert 22M), results in $10,964$ $(70.6\%)$ medical perturbations and $9530$ $(21.65\%)$ RUAR perturbations contained in the verified subspaces. While $21.65\%$ of the $positive$ perturbations contained in the verified subspaces for the RUAR dataset may seem like a low number, it still results in a robust filter, given that the $positive$ class of the dataset contains many adversarial examples of the \emph{same input query}, i.e.\ semantically similar but lexically different queries.
The medical dataset on the other hand contains many semantically diverse queries, and there are several \emph{unseen} medical queries not contained in the dataset nor in the resultant verified subspaces. 
However, given that the subspaces contain $70.6\%$ of the $positive$ perturbations of the medical safety dataset, an application of this could be to carefully curate a new dataset containing only queries with \emph{critical} and \emph{serious} risk-level labels defined by the World Economic Forum for chatbots in healthcare (see Section~\ref{datasets} and \cite{WEF}).
This dataset could be used to create verified filters centred around these queries to prevent generation of medical advice for these high-risk categories.

\noindent Overall, we find that \textbf{semantically-informed verification generalises well} across the different kinds of data to ensure guarantees on the output, and thus should aid in ensuring the safety of LLMs.

\section{Conclusions and Future Work}
\label{sec:section6}

\emph{Summary.} This paper started with a general analysis of existing NLP verification approaches, 
with a view of identifying key components of a general NLP verification methodology. We then distilled these into a ``NLP Verification Pipeline'' consisting of the following six components: 
\begin{enumerate}
    \item dataset selection;
    \item generation of perturbations;
    \item choice of embedding functions;
    \item definition of subspaces;
    \item robust training;
    \item verification via one of existing verification algorithms.
\end{enumerate}

Based on this taxonomy, we make concrete selections for each component, and implement the pipeline using the tool ANTONIO~\cite{FoMLAS2023:ANTONIO_Towards_Systematic_Method}.
ANTONIO allowed us to mix and match different choices for each pipeline component, enabling us to study the effects of varying the components of the pipeline in a algorithm-independent way. 
Our main focus was to identify weak or missing parts of the existing NLP verification methodologies. 
We proposed that NLP verification results should report, in addition to the standard verifiability metric, the following:

\begin{itemize}
    \item whether they use geometric or semantic subspaces, and for which type of semantic perturbations;
    \item volumes, generalisability and embedding error of verified subspaces.
\end{itemize}

We finished the paper with a study of the current limitations of the NLP components of the pipeline and proposed possible improvements such as introducing a perturbations filter stage using cosine similarity.
One of the major strengths of the pipeline is that each component can be improved individually.

\emph{Contributions.} The major discoveries of this paper were: 
\begin{itemize}
    \item In Section~\ref{sec:section4} we proposed generalisability as a novel metric, and showed that NLP verification methods exhibit a generalisability-verifiability trade-off. The effects of the trade-off can be severe, especially if the verified subspaces are generated naively (e.g. geometrically).  We therefore strongly believe that generalisability should be routinely reported as part of NLP verification pipeline.
    \item In Sections~\ref{sec:section4} and~\ref{sec:section5} we showed that it is possible to overcome this trade-off by using several heuristic methods: defining semantic subspaces, training for semantic robustness, choosing a suitable embedding function and filtering with cosine similarity. All of these methods result in the definition of more precise verifiable subspaces; and all of them can be practically implemented as part of NLP verification pipelines in the future. This is the main positive result of this paper.
    \item In Section~\ref{sec:section5} we demonstrated that there are two key assumptions underlying the definition of subspaces that cannot be taken for granted. Firstly the LLMs, used as embedding functions, may not map semantically similar sentences to similar vectors in the embedding space. Secondly, our algorithmic methods for generating perturbations, whether by LLMs or otherwise, may not always be semantically-preserving operations. Both of these factors influence practical applications of the NLP verification pipeline.
    \item In Section~\ref{sec:section5} we demonstrated that even verified subspaces can exhibit semantic embedding errors: this effect is due to the tension between verification methods that are essentially geometric and the intuitively understood semantic meaning of sentences. 
    By defining the \emph{embedding error} metric and using it in our experiments, we demonstrated that the effects of embedding errors do not seem to be severe in practice; but this may vary from one scenario to another. It is important that NLP verification papers are aware of this pitfall, and report embedding error alongside verifiability and generalisability.
\end{itemize}

Finally, we claim as a contribution, a novel, coherent, methodological framework that allows us to include a broad spectrum of NLP, geometric, machine learning, and verification methods under a single umbrella. As illustrated throughout this paper, no previous publication in this domain covered this range and we believe that covering this broad range of methods is crucial for the development of this field.

\emph{Limitations and the Role of the Embedding Gap.}
In this paper, we have shown the effect of the embedding gap in the NLP verification domain. In Section~\ref{sec:generalisability}, we introduced the generalisability metric (Definition~\ref{def:generalisability}) to estimate how well subspaces capture semantically similar yet unseen sentences, providing a quantitative lens to examine this challenge. Sections~\ref{sec:subsection43} and~\ref{sec:sec4.2} demonstrated that geometric subspaces struggle with the verifiability-generalisability trade-off, while semantic subspaces, constructed using semantic-preserving perturbations, show promise in mitigating the embedding gap by better aligning with data semantics.

The embedding gap is not unique to NLP, and manifests itself nearly in every domain where machine learning is applied. For example, in computer vision, the geometric definition of an $\epsball$ can include perturbations that no longer semantically resemble the original image (e.g., distortions that transform a given image into something unrecognisable).
In the verification of neural network controllers, possible discrepancies arise between interpretation of neural network inputs, such as velocity, distance, angle (i.e. have physical interpretation) and the way in which the neural network treats them as normalised input vectors~\cite{daggitt2024vehicle,cordeiro2024neural}.
In NLP, the gap is amplified by the use of LLMs to map discrete sentences into continuous vector spaces. This process lacks a one-to-one correspondence between semantics and embeddings, exacerbating the challenge.

This problem is fundamental for machine learning methods deployed in NLP: they always rely on an “embedding function” that maps sentences into real vectors (on which machine learning algorithms operate). There is an implicit assumption that the embedding function works in a way that semantic similarity of sentences is reflected in geometric proximity of their embeddings. However, as general semantic similarity of sentences is not effectively computable, there is no hope that a perfect embedding function will ever be defined. Fundamentally, this is exactly the reason why machine learning (and not symbolic) approaches to NLP proved to be  successful: they operate on the assumption that the embedding function is imperfect.  
 As a consequence, any verification pipeline for NLP must include metrics that measure potential embedding errors. Section \ref{sec:embeddingerror} of this paper is entirely devoted to defining this problem in mathematically precise terms and proposing an effective metric for measuring and reporting the severity and effects of the embedding errors.  

This paper aims to quantify and address the embedding gap in the NLP domain.
For better quantifying the effect of the embedding gap, we proposed precise metrics such as verifiability, generalisability and embedding error, and showed their interplay.
This better understanding of the problem gave rise to our main positive result: the method that empirically reduces the effect of the embedding gap.
While our findings mark progress, further research is needed to better align geometric representations with semantic meaning, especially in NLP contexts.

\emph{Future Work.}
Following from our in-depth analysis of the NLP perspective, we note that even if one has a satisfactory solution to all the issues discussed, there is still the problem of scalability of the available verification algorithms. For example, the most performant neural network verifier, $\alpha\beta$-Crown~\cite{wang2021beta}, can only handle networks in the range of tens of millions of trainable parameters. In contrast, in NLP systems, the base model of BERT~\cite{bert} has around 110 million trainable parameters (considered small compared to modern LLMs -- with trainable parameters in the billions!). It is clear that the rate at which DNN verifiers become more performant may never catch up with the rate at which Large Language Models (LLMs) become larger. Then the question arises: how can this pipeline be implemented in the real world?

\begin{figure}[t]
    \centering
    \includegraphics[width=0.8\columnwidth]{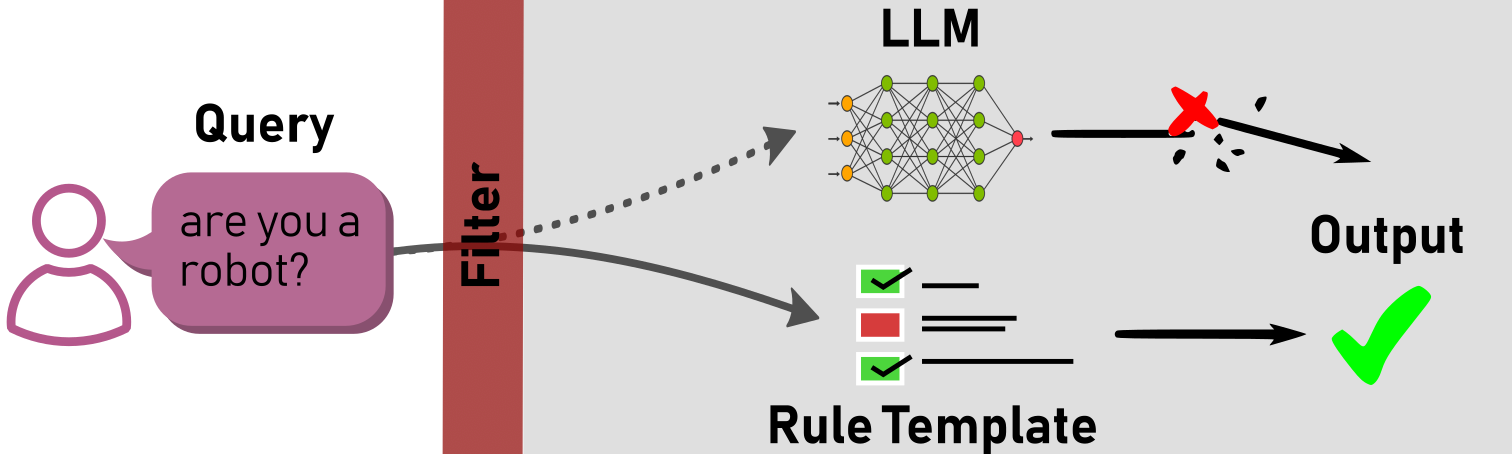}
    \caption{\small\emph{In this figure, we show how a prepended, semantically informed verified filter added to an NLP system (here, an LLM), can check that safety-critical queries are handled responsibly, e.g. by redirecting the query to a tightly controlled rule-based system instead of a stochastic LLM.
    }}
    \label{fig:sys_architecture}
\end{figure}

For future work, we propose to tackle this based on the idea of verifying a smaller DNN (classifier), manageable by verifiers, that can be placed upstream of a complex NLP system as a \emph{safeguard}.
We call this a \emph{filter} (as mentioned in Section~\ref{sec:section3} and illustrated in Figure~\ref{fig:filter}), and Figure~\ref{fig:sys_architecture} shows how a semantically informed verified filter can be prepended to an NLP system (here, an LLM) to check that safety-critical queries are handled responsibly, e.g. by redirecting the query to a tightly controlled rule-based system instead of a stochastic LLM.
While there are different ways to implement the verification filters (e.g. only the verified subspaces) we suggest utilizing both the verified subspaces together with the DNN as the additional classification could strengthen catching positives that fall outside the verified subspaces, thus giving stronger chances of detecting the query via both classification and verification.

We note that the NLP community has recently proposed guardrails, in order to control the output of LLMs and create safer systems (such as from \href{https://docs.guardrailsai.com/integrations/openai_functions/}{Open AI}, \href{https://github.com/NVIDIA/NeMo-Guardrails}{NVIDIA} and so on). These guardrails have been proposed at multiple stages of an NLP pipeline, for example an \emph{output rail} that checks the output returned by an LLM, or \emph{input rail}, that rejects unsafe user queries. In Figure~\ref{fig:sys_architecture}, we show an application of our filter applied to the user input, which thus creates guarantees that a subset of safety critical queries are handled responsibly. In theory these verification techniques we propose may be applied to guardrails at different stages in the system, and we plan to explore this in future work.

A second future direction is to use this work to create NLP verification benchmarks.
In 2020, the International Verification of Neural Networks Competition~\cite{brix2023first} (VNN-COMP) was established to facilitate comparison between existing approaches, bring researchers working on the DNN verification problem together, and help shape future directions of the field.
However, for some time, the competition still lacked NLP verification benchmarks~\cite{brix2023fourth}.
In 2024, we contributed a first NLP benchmark to VNN-COMP, using the methodology of this paper, and the tool ANTONIO~\cite{brix2024fifth}.
We intend to use this work for creating NLP verification benchmarks for future editions, to spread the awareness and attention to this field.

\section*{Acknowledgements}
Authors acknowledge support of EPSRC grant AISEC EP/T026952/1 and NCSC grant `Neural Network Verification: in search of the missing spec', Vinay Krupakaran for his help in the manual annotation of the perturbed sentences and Daniel Kienitz for his help with the eigenspace rotation.
The first author acknowledges the James Watt Scolarship awarded by the Heriot-Watt University.

\bibliographystyle{unsrtnat}
\bibliography{references}

\end{document}